\pdfoutput=1

\documentclass{article}

\usepackage[nonatbib,preprint]{neurips_2021}
\usepackage[square,sort,comma,numbers]{natbib}

\usepackage{nicefrac}       %

\usepackage{scalefnt,letltxmacro}
\LetLtxMacro{\oldtextsc}{\textsc}
\renewcommand{\textsc}[1]{\oldtextsc{\scalefont{1.10}#1}}
\usepackage[acronym,smallcaps,nowarn]{glossaries}
\glsdisablehyper
\makeglossaries
\usepackage{xspace}
\usepackage{float}
\usepackage{physics}

\usepackage{enumitem}

\usepackage{amssymb}
\usepackage{mathtools}
\usepackage{amsfonts}
\usepackage{amsmath}
\usepackage{amsthm}
\usepackage{booktabs}
\usepackage[]{microtype}

\usepackage[linesnumbered,ruled,vlined]{algorithm2e}

\SetCommentSty{mycommfont}
\SetKwInput{KwInput}{Input} 
\SetKwInput{KwOutput}{Output}

\usepackage[colorlinks,linktoc=all]{hyperref}
\usepackage[all]{hypcap}
\hypersetup{citecolor=MidnightBlue}
\hypersetup{linkcolor=MidnightBlue}
\hypersetup{urlcolor =MidnightBlue}

\usepackage[capitalize,nameinlink]{cleveref}
\crefname{section}{\S}{\S\S}
\Crefname{section}{\S}{\S\S}
\creflabelformat{equation}{#2\textup{#1}#3}

\usepackage[usenames,dvipsnames,svgnames,table]{xcolor}
\definecolor{shadecolor}{gray}{0.9}

\definecolor{mylightgray}{gray}{0.94}

\makeatletter
\@ifundefined{c@rownum}{%
  \let\c@rownum\rownum
}{}
\@ifundefined{therownum}{%
  \def\therownum{\@arabic\rownum}%
}{}
\makeatother

\makeatletter
\newcommand*{\addFileDependency}[1]{%
	\typeout{(#1)}
	\@addtofilelist{#1}
	\IfFileExists{#1}{}{\typeout{No file #1.}}
}
\makeatother

\usepackage[font=small,labelfont=bf,tableposition=top]{caption}
\usepackage{tikz}
\usepackage{graphicx}
\usepackage[]{subcaption}   %

\usepackage{pgfplots}
\pgfplotsset{compat=1.6}
\usepgfplotslibrary{groupplots}
\tikzstyle{every picture}+=[font=\sffamily]
\tikzstyle{optimized} = [circle,fill=white,draw=black, dashed,inner sep=1pt, minimum size=20pt, font=\fontsize{10}{10}\selectfont, node distance=1]
\pgfkeys{/pgf/number format/.cd,1000 sep={}}
\pgfplotsset{
	tick label style = {font=\sffamily},
	every axis label/.append style={font=\sffamily},
	typeset ticklabels with strut,
}
\pgfplotsset{every axis/.append style={
			every x tick label/.append style={font=\fontsize{6pt}{6pt}\sffamily, yshift=.5ex,},
			every y tick label/.append style={font=\fontsize{6pt}{6pt}\sffamily, xshift=.5ex},
			every y label/.append style={xshift=10ex, font=\sffamily},
			every x label/.append style={yshift=3ex, font=\sffamily},
			every title/.append style={font=\sffamily}
		},
}
\pgfplotsset{
  xticklabel={$\mathsf{\pgfmathprintnumber{\tick}}$},
  yticklabel={$\mathsf{\pgfmathprintnumber{\tick}}$},
}
\pgfplotsset{every axis title/.append style={yshift=-1ex}}
\newlength\figureheight
\newlength\figurewidth
\usepgfplotslibrary{external}

\usepackage[colorinlistoftodos,
	textsize=scriptsize,
	linecolor=red!30,
	bordercolor=red!30,
	backgroundcolor=red!10]{todonotes}
\renewcommand{\todo}[2][]{\tikzexternaldisable\@todo[#1]{#2}\tikzexternalenable}

\usepackage[most]{tcolorbox}
{\begin{tcolorbox}[toprule=2mm,left=4pt,right=4pt,top=0pt,bottom=1pt,boxsep=2pt, before skip = 2ex, after skip =2ex]}{\end{tcolorbox}}
\tcbset{boxsep=4pt,left=2pt,right=2pt,top=-0pt,bottom=0pt}

\usepackage{url}

\usepackage{xr}
\usepackage{subcaption}

\usepackage{tikz}
\usepackage{graphicx}
\usepackage{xcolor}
\usepackage{amsmath}
\usepackage{amssymb}
\usepackage{bm}

\usepackage{adjustbox}
\usepackage{multirow}
\usepackage{mathtools}

\usepackage{xspace}

\newacronym{MAP}{map}{maximum-a-posteriori}
\newacronym{MLE}{mle}{maximum likelihood estimation}
\newacronym{MNLL}{mnll}{mean negative loglikelihood}
\newacronym{NLL}{nll}{negative loglikelihood}
\newacronym{LL}{ll}{log-likelihood}
\newacronym{RMSE}{rmse}{root mean square error}
\newacronym{ECE}{ece}{expected calibration error}
\newacronym{FID}{fid}{Fr\'echet Inception Distance}

\newacronym{AE}{ae}{autoencoder}
\newacronym{WAE}{wae}{Wasserstein Autoencoder}
\newacronym{VAE}{vae}{Variational Autoencoder}
\newacronym{BAE}{bae}{Bayesian autoencoder}
\newacronym{CDF}{cdf}{cumulative density function}
\newacronym{GAN}{gan}{Generative Adversarial Network}

\newacronym{MC}{mc}{Monte Carlo}
\newacronym{MCMC}{mcmc}{Markov chain Monte Carlo}
\newacronym{HMC}{hmc}{Hamiltonian Monte Carlo}
\newacronym{MH}{mh}{Metropolis-Hastings}
\newacronym{NUTS}{nuts}{no-u-turn sampler}
\newacronym{SGHMC}{sghmc}{stochastic gradient Hamiltonian Monte Carlo}

\newacronym{DGP}{dgp}{deep Gaussian process} %
\newacronym{GPLVM}{gplvm}{Gaussian process latent variable model}
\newacronym{DPMM}{dpmm}{Dirichlet Process Mixture Model}

\newacronym{VFE}{vfe}{variational free energy}

\newacronym[firstplural=Gaussian Processes]{GP}{gp}{Gaussian Process}

\newacronym{VI}{vi}{variational inference}

\newacronym{ELBO}{elbo}{evidence lower bound}
\newacronym{NELBO}{nelbo}{negative evidence lower bound}
\newacronym{ELL}{ell}{expected log likelihood}
\newacronym{KL}{kl}{Kullback-Leibler divergence}
\newacronym{AUC}{auc}{area under the curve}

\newacronym[firstplural=Bayesian neural networks]{BNN}{bnn}{Bayesian neural network}
\newacronym[firstplural=deep neural networks]{DNN}{dnn}{deep neural network}
\newacronym[]{CNN}{cnn}{convolutional neural network}
\newacronym{MLP}{mlp}{multilayer perceptron}
\newacronym{NN}{nn}{neural network}
\newacronym{RELU}{ReLU}{rectified linear unit}

\newacronym{NF}{nf}{normalizing flow}

\newacronym{RBF}{rbf}{radial basis function}
\newacronym{ARD}{ard}{automatic relevance determination}

\newacronym{RKHS}{rkhs}{reproducing kernel Hilbert space}

\newacronym{OT}{ot}{optimal transport}
\newacronym{WD}{wd}{Wasserstein distance}
\newacronym{SWD}{swd}{sliced-Wasserstein distance}
\newacronym{DSWD}{dswd}{distributional sliced-Wasserstein distance}

\newcommand{\name}[1]{{\textsc{#1}}\xspace}

\newcommand{\mnist}{\name{mnist}}
\newcommand{\celeba}{\name{celeba}}
\newcommand{\yale}{\name{yale}}
\newcommand{\frey}{\name{frey}}
\newcommand{\freyyale}{\textsc{frey}-\textsc{yale}\xspace}

\newcommand{\wae}{\name{wae}}

\newcommand{\mathbold}[1]{{\boldsymbol{\mathbf{#1}}}}

\newcommand{\g}{\,|\,}

\newcommand{\nestedmathbold}[1]{{\mathbold{#1}}}

\newcommand{\mba}{\nestedmathbold{a}}

\newcommand{\mbr}{\nestedmathbold{r}}

\newcommand{\mbv}{\nestedmathbold{v}}
\newcommand{\mbw}{\nestedmathbold{w}}
\newcommand{\mbx}{\nestedmathbold{x}}
\newcommand{\mby}{\nestedmathbold{y}}
\newcommand{\mbz}{\nestedmathbold{z}}

\newcommand{\mbA}{\nestedmathbold{A}}
\newcommand{\mbB}{\nestedmathbold{B}}
\newcommand{\mbC}{\nestedmathbold{C}}

\newcommand{\mbI}{\nestedmathbold{I}}

\newcommand{\mbM}{\nestedmathbold{M}}

\newcommand{\mbP}{\nestedmathbold{P}}

\newcommand{\mbS}{\nestedmathbold{S}}

\newcommand{\mbU}{\nestedmathbold{U}}
\newcommand{\mbV}{\nestedmathbold{V}}

\newcommand{\mbbeta}{\nestedmathbold{\beta}}

\newcommand{\mbphi}{\nestedmathbold{\phi}}

\newcommand{\mbpsi}{\nestedmathbold{\psi}}

\newcommand{\mbtheta}{\nestedmathbold{\theta}}

\DeclareRobustCommand{\KL}[2]{\ensuremath{\textsc{kl}\left[#1\;\|\;#2\right]}}
\DeclarePairedDelimiterX{\infdivx}[2]{[}{]}{%
  #1\;\delimsize\|\;#2%
}

\DeclareMathOperator*{\argmax}{arg\,max}
\DeclareMathOperator*{\argmin}{arg\,min}

\newcommand{\cL}{\mathcal{L}}
\newcommand{\cN}{\mathcal{N}}

\newcommand{\cR}{\mathcal{R}}

\newcommand{\cH}{\mathcal{H}}
\newcommand{\cW}{\mathcal{W}}

\newcommand{\E}{\mathbb{E}}

\newcommand{\bbR}{\mathbb{R}}
\newcommand{\bbS}{\mathbb{S}}

\newcommand{\defeq}{\stackrel{\text{\tiny def}}{=}}

          \newcommand{\xinput}{\mbx}
            \newcommand{\xobs}{\mbx}
        \newcommand{\xgen}{\hat\mbx}

\newcommand{\latent}{\mbz}

\title{Model Selection for Bayesian Autoencoders}

\author{%
  Ba-Hien Tran \\
  \small{EURECOM} \\ \small{(France)}
  \And 
  Simone Rossi \\
  \small{EURECOM} \\ \small{(France)}
  \And 
  Dimitrios Milios \\
  \small{EURECOM} \\ \small{(France)}
  \AND
  Pietro Michiardi \\
  \small{EURECOM} \\ \small{(France)}
  \And 
  Edwin V. Bonilla \\
  \small{CSIRO's Data61 and }\\\small{The University of Sydney} \\ \small{(Australia)}
  \And 
  Maurizio Filippone\\
  \small{EURECOM} \\ \small{(France)}
}

\begin{document}

\maketitle

\begin{abstract}

 We develop a novel method for carrying out model selection for Bayesian autoencoders (BAEs) by means of prior hyper-parameter optimization. Inspired by the common practice of type-II maximum likelihood optimization and its equivalence to Kullback-Leibler divergence minimization, we propose to optimize the distributional sliced-Wasserstein distance (DSWD) between the output of the autoencoder and the empirical data distribution. The advantages of this formulation are that we can estimate the DSWD based on samples and handle high-dimensional problems. We carry out posterior estimation of the BAE parameters via stochastic gradient Hamiltonian Monte Carlo and turn our BAE into a generative model by fitting a flexible Dirichlet mixture model in the latent space. Consequently, we obtain a powerful alternative to variational autoencoders, which are the preferred choice in modern applications of autoencoders for representation learning with uncertainty. We evaluate our approach qualitatively and quantitatively using a vast experimental campaign on a number of unsupervised learning tasks and show that, in small-data regimes where priors matter,  our approach provides state-of-the-art results, outperforming multiple competitive baselines.

\end{abstract}

\section{Introduction}

The problem of %
learning useful representations of  data that facilitate the solution of downstream tasks such as  clustering, generative modeling  and classification, is at the crux of the success of many machine learning applications \citep[see, e.g.,][and references therein]{bengio2013representation}. %
From a plethora of potential solutions to this problem, unsupervised approaches based on autoencoders \cite{Cottrell89}  are particularly appealing as, by definition, they do not require label information and have proved effective in tasks such as  dimensionality reduction and information retrieval \cite{Hinton06}.

Autoencoders are %
neural network models composed of two parts, usually referred to as the encoder and the decoder.
The encoder maps each input $\xobs_i$ to a set of lower-dimensional latent variables $\mbz_i$. %
The decoder maps the latent variables $\mbz_i$ back to the observations $\xobs_i$.
The bottleneck introduced by the low-dimensional latent space is what characterizes the compression and representation learning capabilities of autoencoders.
It is not surprising that these models have connections with principal component analysis \cite{BaldiHornik89}, factor analysis and density networks \cite{MacKayGibbs99}, and latent variable models \cite{Lawrence05}.

In applications where  quantification of uncertainty is a primary requirement or  where  data is scarce, it is important to carry out a Bayesian treatment of these models by specifying a prior distribution over their parameters, i.e., the weights of the encoder/decoder. 
However, estimating the posterior distribution over the parameters of these models, which we refer to as \glspl{BAE}, is generally intractable and requires approximations. 
Furthermore, the need to specify priors for a large number of parameters, coupled with the fact that autoencoders are not generative models, has motivated the development of \glspl{VAE} as an alternative that can overcome  these limitations \cite{Kingma14}. Indeed, 
\glspl{VAE} have found tremendous success and have become one of the  preferred methods  in modern machine-learning applications \citep[see, e.g.,][and references therein]{KingmaW19}.

To recap, three potential limitations of \glspl{BAE} hinder their widespread applicability in order to achieve a similar or superior adoption to their variational counterpart: 
(i) lack of generative modeling capabilities;
(ii) intractability of inference and
(iii)   difficulty of setting sensible priors over their parameters.    
 In this work we revisit \glspl{BAE} and deal with these limitations in a principled way.
 In particular, we address the  first limitation in (i) by employing density estimation in the latent space. %
 Furthermore, we deal with the  second limitation in (ii) by exploiting recent advances in \gls{MCMC} and, in particular, \gls{SGHMC} \cite{Cheni2014}. 
 Finally, we believe that the third limitation (iii), which we refer to as the difficulty of carrying out \emph{model selection},  requires a more detailed treatment because choosing sensible priors for Bayesian neural networks is an extremely difficult problem, and this is the main focus of this work.

\paragraph{Contributions.}
Specifically, in this paper we provide a novel, practical, and elegant way of performing model selection for \glspl{BAE}, which allows us to revisit these models for applications where \glspl{VAE} are currently the primary choice.
We  start by considering the common practice of estimating  prior (hyper-)parameters via type-II maximum likelihood, which is equivalent to minimizing the \gls{KL} between the distribution induced by the \gls{BAE} and the data generating distribution. %
Because of the intractability of this objective and the difficulty to estimate it through samples, we resort to an alternative formulation where we replace the \gls{KL} with the \gls{DSWD} between these two distributions.
The advantages of this formulation are that we can estimate the \gls{DSWD} based on samples and, thanks to the slicing, we can handle large dimensional problems.
Once \gls{BAE} hyper-parameters are optimized, we %
estimate  the posterior distribution over the \gls{BAE} parameters %
via \gls{SGHMC} \cite{Cheni2014}, which is a powerful sampler that operates on mini-batches and has proven effective for Bayesian deep/convolutional networks \cite{Tran20, ZhangLZCW20, Izmailov21}.
Furthermore, we  turn our \gls{BAE} into a generative model by fitting a flexible mixture model in the latent space, namely the \gls{DPMM}.
We evaluate our approach qualitatively and quantitatively using a vast experimental campaign on a number of unsupervised learning tasks, with particular emphasis on the challenging task of generative modeling when the number of observations is small. %

\subsection{Related work}
\glspl{VAE} provide a theoretically-grounded and popular framework for representation learning and deep generative modeling.
However, training \glspl{VAE} poses considerable practical and theoretical challenges yet to be solved.
In practice, the learned aggregated posterior distribution of the encoder rarely matches the latent prior, 
and this hurts the quality of generated samples.
Several methods have been proposed to deal with this problem by using a more expressive form of priors on the latent space \cite{NalisnickS17, Chen2017, TomczakW18, BauerM19}.
Similar to our work, there is a line of research that employs a form of ex-post density estimation on the learned latent space \cite{dai2018diagnosing, bohm2020probabilistic, Ghosh2020From}.
\glspl{WAE} \cite{tolstikhin2018wasserstein} impose a new form of regularization on latent space by reformulating the objective function as an \gls{OT} problem.
There have been previous attempts to apply the Bayesian approach to  \glspl{VAE}.
For example, \cite{Daxberger19} treats the parameters of \gls{VAE}'s encoder and decoder in a Bayesian manner to deal with out-of-distribution samples.
Most of these works focus on imposing prior or regularization on the latent or weight space of autoencoders.
In this work, we take a different route, as we aim to impose prior knowledge directly on the output space.
Indeed, our work is motivated by recent attempts to rethink prior specification for \glspl{BNN}.
It is extremely difficult to choose a sensible prior on the parameters of \glspl{BNN} \citep[see, e.g.,][and references therein]{Nalisnick18} because their effect on the distribution of the induced functions is difficult to characterize. 
Thus, recent attempts in the literature have turned their attention towards defining priors in the space of functions \cite{Nalisnick0H21, YangLGLD20, Hafner2019, Sun2019}.
Closest to our work is that of \cite{Tran20}, which matches the functional prior induced by \glspl{BNN} to \gls{GP} priors by means of the Kantorovich-Rubinstein dual form of the Wasserstein distance.
Different from this line of works, we consider a general framework to impose a functional prior for \glspl{BNN} in an unsupervised learning setting.

\section{Preliminaries on Bayesian Autoencoders}
An \gls{AE} is a neural network parameterized by a set of parameters $\mbw$, which transforms an unlabelled dataset, $\xobs \defeq \{\xobs_n\}_{n=1}^{N} $, into a set of reconstructions $\hat{\xobs} \defeq \{\xgen_{n}\}_{n=1}^{N} $, with  $\xobs_n,\xgen_n\in\bbR^D$.
An \gls{AE} is composed of two components: (1) an encoder $f_{\text{enc}}$ which maps an input sample $\xobs_n$ to a latent code $\latent _n\in \bbR^K, K \ll D$; and (2) a decoder $f_{\text{dec}}$ which maps the latent code to a reconstructed datapoint $\xgen_n$.
In short, $\xgen = f(\xinput; \mbw) = (f_{\text{dec}} \circ f_{\text{enc}}) (\xinput)$, where we denote $\mbw := \{\mbw_{\text{enc}}, \mbw_{\text{dec}}\}$ the union of parameters of the encoder and decoder.
The Bayesian treatment of \glspl{AE} dictates that a prior distribution $p(\mbw)$ is placed over all parameters of $f_{\text{enc}}$ and $f_{\text{dec}}$,  
and that this prior knowledge is transformed into a posterior distribution by means of Bayes' theorem, 
\begin{align}
   \label{eq:bayes_rule}
   p(\mbw \g \xobs) = \frac{p(\xobs \g \mbw)p(\mbw)}{p(\xobs)}, 
\end{align}
where $p(\xobs \g \mbw)$ is the conditional likelihood that factorizes as $ p(\xobs \g \mbw) =  \prod_{n=1}^N p(\xobs_n\g\mbw)$.
Note that each conditional likelihood term is %
determined by the model architecture, the choice of $\mbw$, and the input $\xinput_n$, 
but in order to keep the notation uncluttered, we write them simply as $p(\xobs_n\g\mbw)$.  %

\paragraph{Likelihood model.}
In the Bayesian scheme, the prior and likelihood are both modeling choices.
Before giving an in-depth treatment on priors for \glspl{BAE} in the next section, %
we briefly discuss the likelihood, which can be chosen according to the type of data.
In our experiments, we mainly investigate image datasets, where pixel values are normalized in the $[0, 1]$ range.
Therefore, we rely on the \emph{continuous Bernoulli} distribution \cite{Loaiza-GanemC19}:
\begin{align}
  p(\xobs_n \g \mbw) 
   = \prod_{i=1}^{D} K(\lambda_i) \lambda_i^{\xobs_{n,i}} (1-\lambda_i)^{1 - \xobs_{n,i}} \coloneqq	 p(\xobs_n \g \xgen_n) ,
   \label{eq:continuous_bernoulli_likelihod}
\end{align}
where $K(\lambda_i)$ is a properly defined normalization constant \cite[][]{Loaiza-GanemC19} and $\lambda_i = f_i(\mbx_{n}; \mbw) = \xgen_{n,i} \in [0, 1]$ is the $i$-th output from the \gls{BAE} given the input $\xinput_{n}$. 
We note that, as $\xgen_{n}$ depends deterministically on $\mbw$,  we will use the above expression to refer to both 
$p(\xobs_n \g \mbw)$  and $p(\xobs_n \g \xgen_n)$, where the latter term will be of crucial importance when we define the functional prior 
induced over the reconstruction $\xgen$.
\paragraph{Inference.}
Although the posterior of \glspl{BAE} is analytically intractable, it can be approximated by variational methods or using \gls{MCMC} sampling. %
Within the large family of approximate Bayesian inference schemes, \gls{SGHMC} \cite{Cheni2014} allows us to sample from the true posterior by efficiently simulating a Hamiltonian system \cite{Neal2011}.
Differently from more traditional methods, \gls{SGHMC} can scale up to large datasets by relying on noisy but unbiased estimates of the potential energy function $U(\mbw) = -\sum_{n=1}^N \log p(\xobs_n \g \mbw) - \log p(\mbw)$.
These can be computed by considering a mini-batch of size $M$ of the data and approximating
$\sum_{n=1}^N \log p(\xobs_n \g \mbw) \approx \frac{N}{M}\sum_{j\in\mathcal{I}_M} \log p(\xobs_j \g \mbw)$,
where $\mathcal{I}_M$ is a set of $M$ random indices.
More details on \gls{SGHMC} %
can be found in the Appendix.

\begin{minipage}{.475\textwidth}
   \paragraph{Pathologies of standard priors.}
   The choice of the prior is important for the Bayesian treatment of any model as it characterizes the hypothesis space \cite{Mackay1992,Murray2005a}. %
   Specifically for \glspl{BAE}, one should note that placing a prior on the parameters of the encoder and decoder has an implicit effect on the prior over the network output (i.e. the reconstruction).
   In addition, the highly nonlinear nature of these models implies that interpreting the effect of the architecture is theoretically intractable and practically challenging.
   Several works argue that a vague prior such as $\cN(0,1)$ is good enough for some tasks and models, like classification with \glspl{CNN} \cite{WilsonI20}.
\end{minipage}\hfill
\begin{minipage}{.5\textwidth}
   \centering
   \scalebox{.93}
   {\setlength\tabcolsep{2pt}
      {
         \tiny
         \fontfamily{phv}\selectfont

         \begin{tabular}{ r c c | c}
            \toprule
            && \textbf{Ouput with} & \textbf{Output with}\\
             & \textbf{Input}                                                                          & $\mathcal{N}(0,1)$ \textbf{Prior}                                                                      & 
             \textbf{Optimized Prior}                                                                         \\
            \midrule

            \raisebox{10pt}{MNIST}
             & \includegraphics[clip,width=0.12\linewidth]{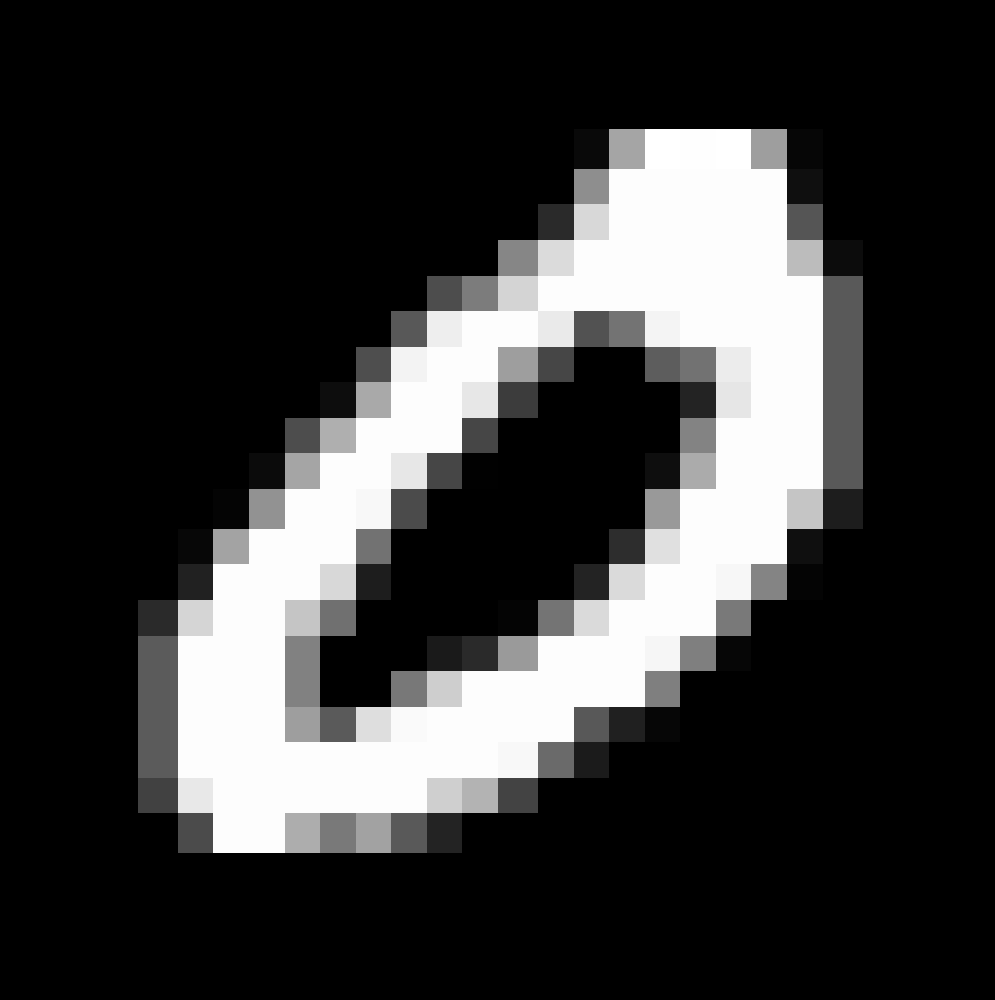}  & \includegraphics[clip,width=0.24\linewidth]{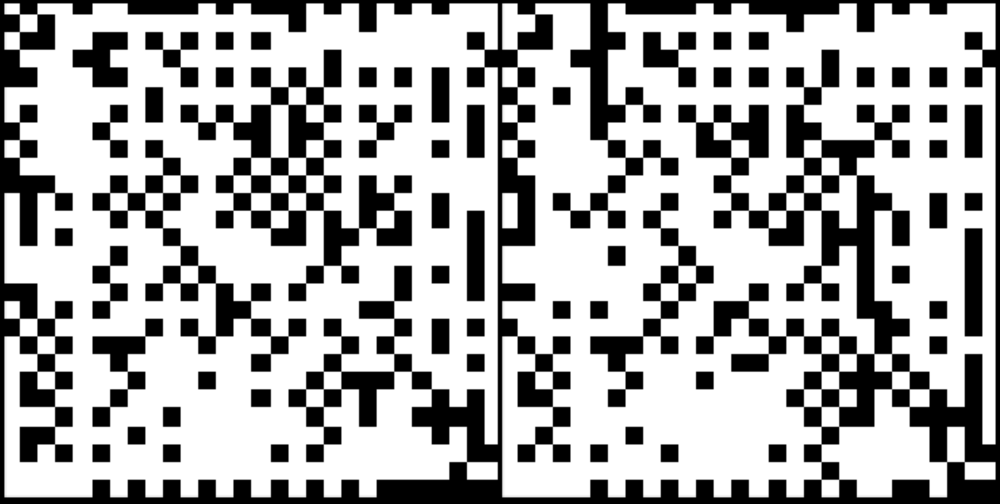}             & \includegraphics[clip,width=0.48\linewidth]{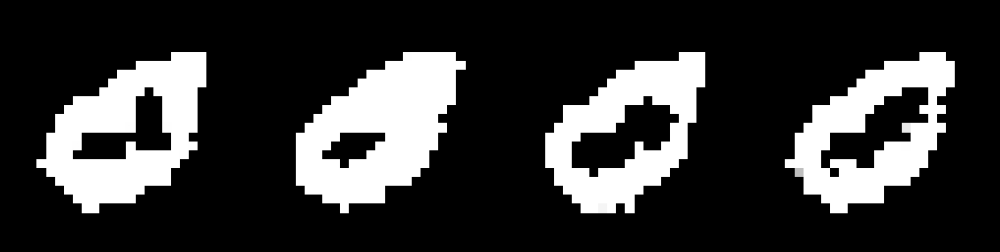}  \\

            \raisebox{10pt}{OOD}
             & \includegraphics[clip,width=0.12\linewidth]{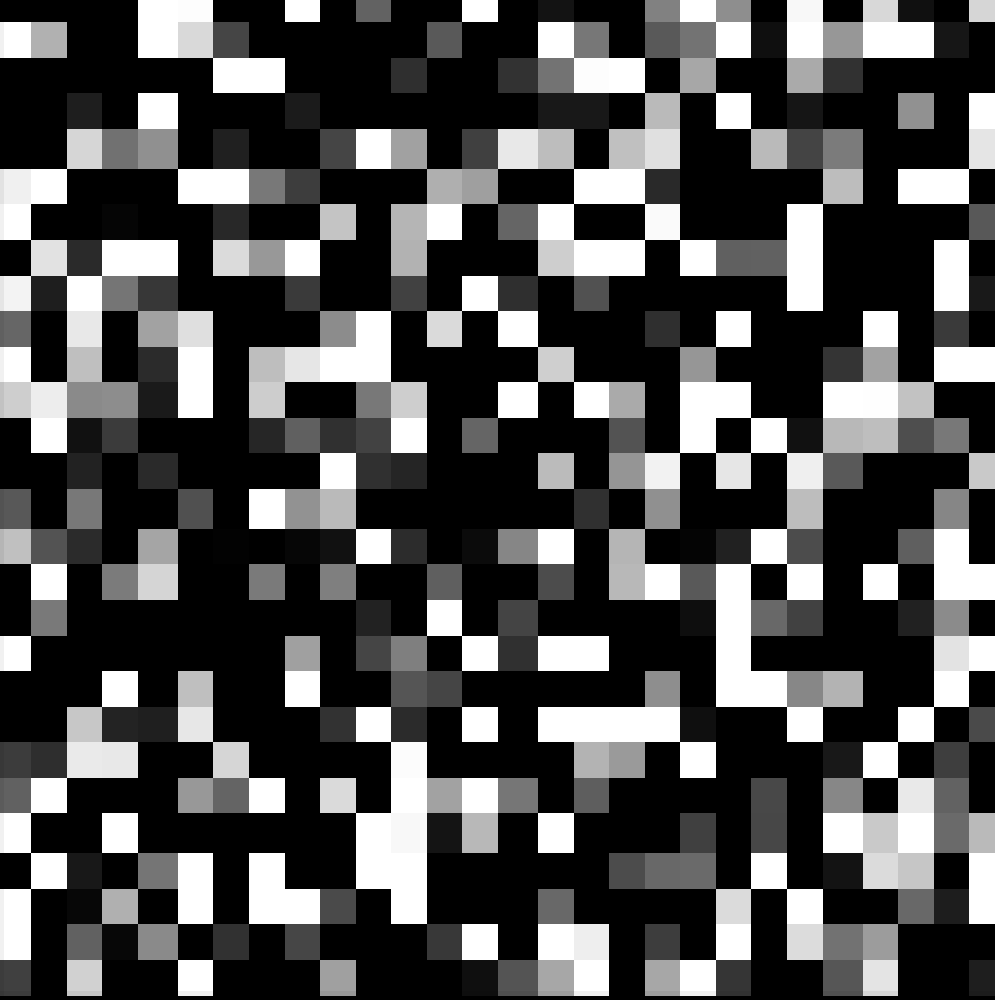}      & \includegraphics[clip,width=0.24\linewidth]{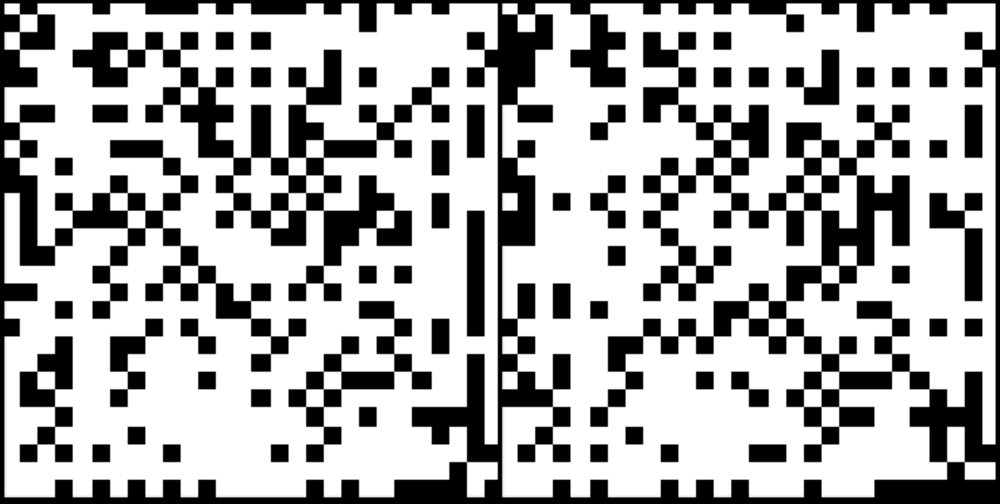}  & \includegraphics[clip,width=0.48\linewidth]{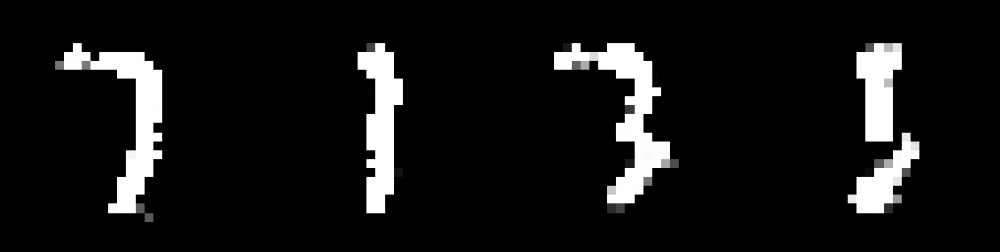}   \\

            \midrule

            \raisebox{10pt}{CELEBA}
             & \includegraphics[clip,width=0.12\linewidth]{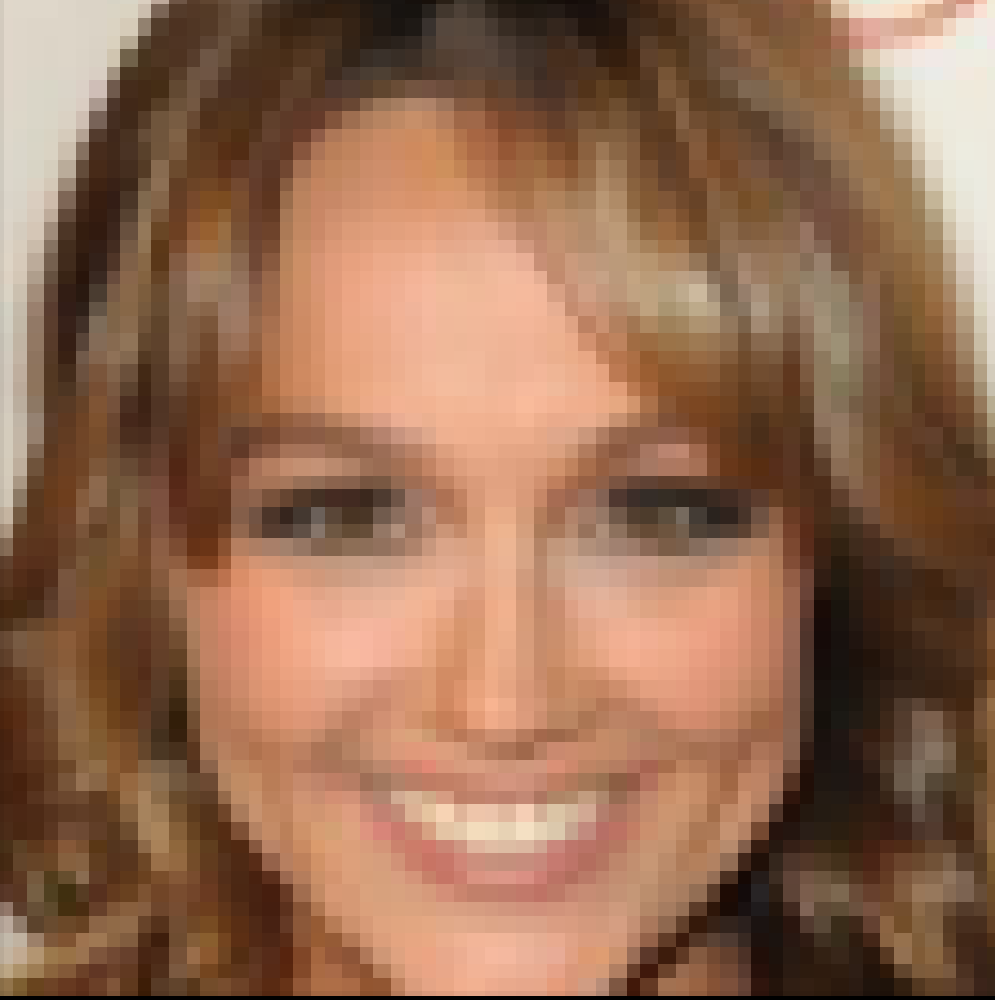} & \includegraphics[clip,width=0.24\linewidth]{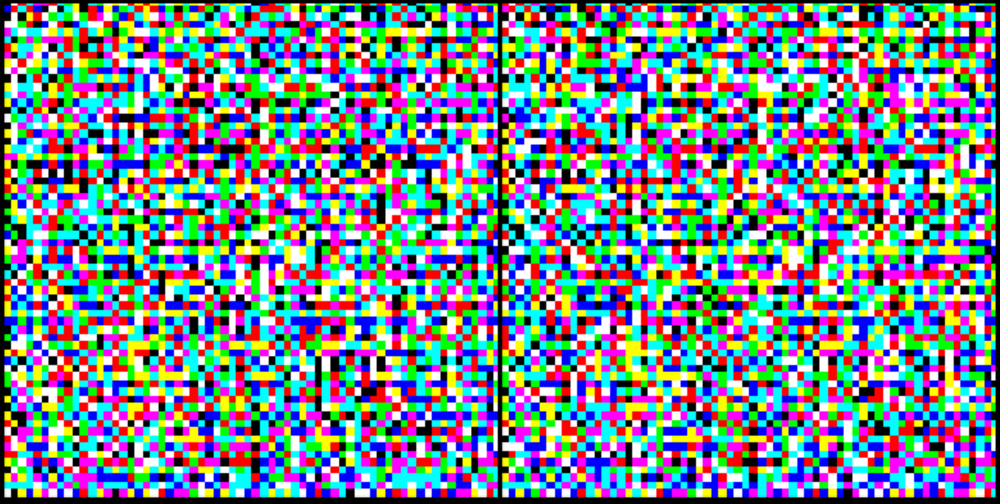}            & \includegraphics[clip,width=0.48\linewidth]{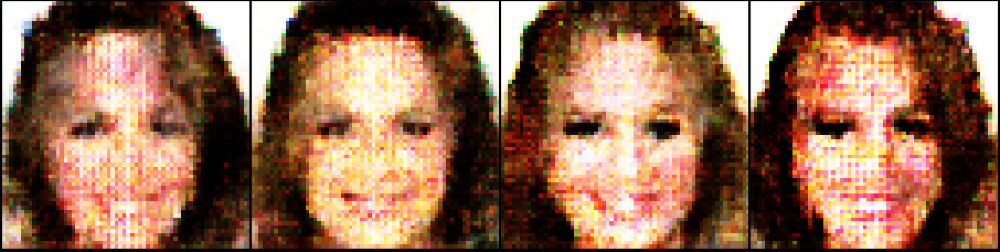} \\

            \raisebox{10pt}{OOD}
             & \includegraphics[clip,width=0.12\linewidth]{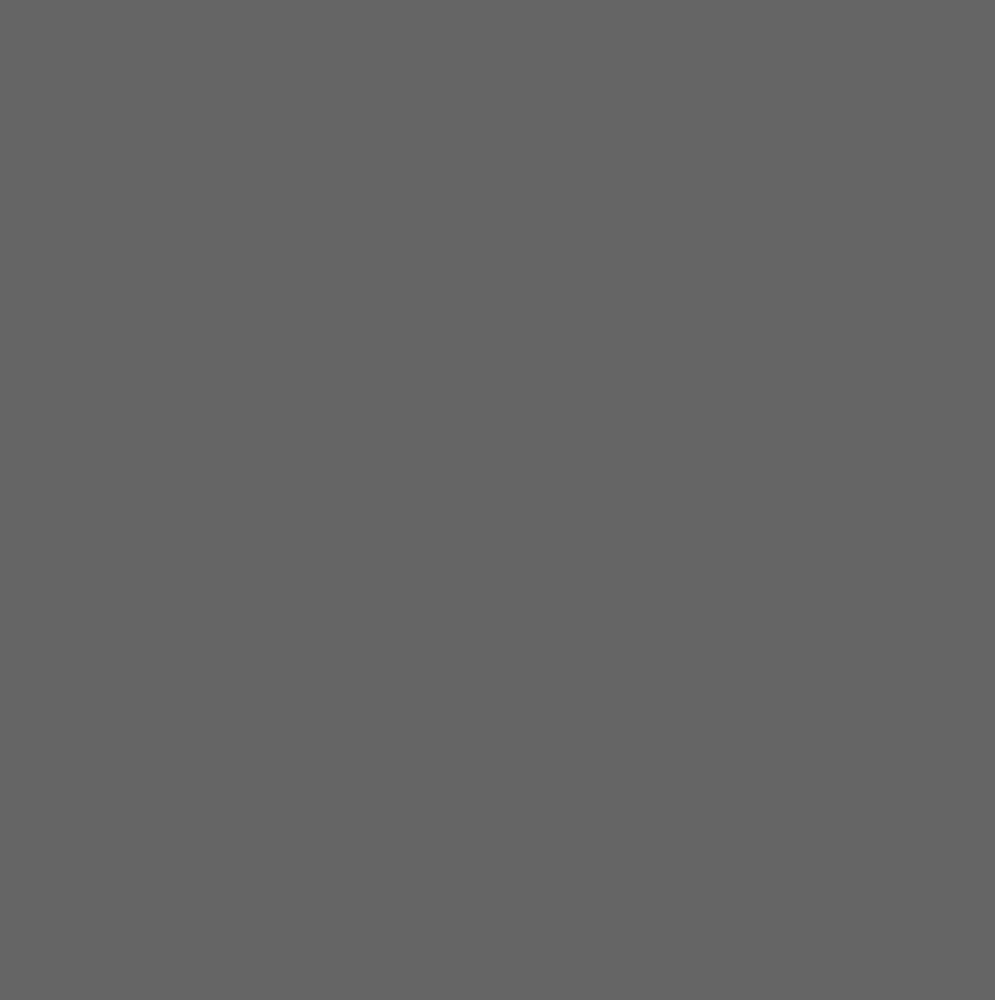} & \includegraphics[clip,width=0.24\linewidth]{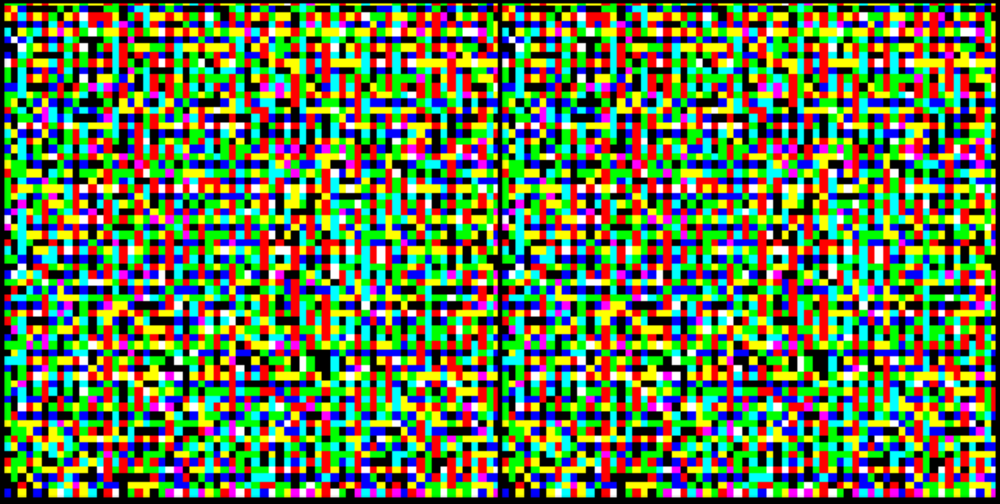} & \includegraphics[clip,width=0.48\linewidth]{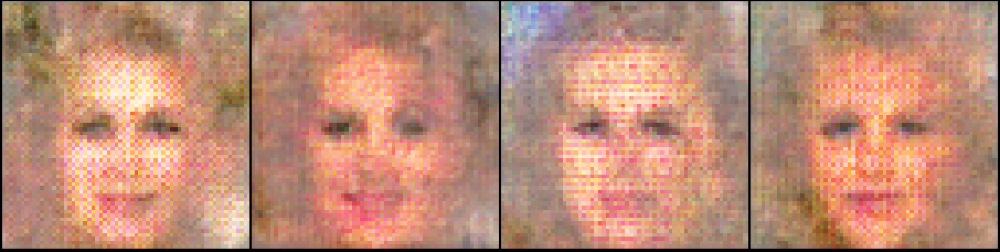}  \\
            \bottomrule
         \end{tabular}
      }
   }

   \captionof{figure}{Realizations sampled from different priors given an input image.
     \textsc{ood} stands for out-of-distribution. 
   }
   \label{fig:prior_samples}
\end{minipage}%

However, for \glspl{BAE} this is not enough, as illustrated in \cref{fig:prior_samples}.
The realizations obtained by sampling weights/biases from a $\cN(0,1)$ prior indicate that this choice provides poor inductive bias.
Meanwhile, by encoding better beliefs via an optimized  prior, which is the focus of  the next section, the samples can capture main characteristics intrinsic to the data, even when the model is fed with out-of-distribution inputs.

\section{Model Selection for Bayesian Autoencoders via  Prior Optimization}
One of the main advantages of the Bayesian paradigm is that we can incorporate prior knowledge into the model in a principled way.
Let us assume a prior distribution $p_\mbpsi(\mbw)$ on the parameters of the \gls{AE} network, where now we are explicit on  the set of (hyper-)parameters that determine the prior, i.e., $\mbpsi$.
Specifying this prior for the \gls{BAE} is not straightforward due to the complex nonlinear forms of $f_\text{enc}$ and $f_\text{dec}$, which induce 
a non-trivial effect on the %
output (functional) prior: %
\begin{align}
  p_\mbpsi({\xgen}) = \int f(\mbx; \mbw) p_\mbpsi(\mbw) \dd \mbw,
   \label{eq:func_prior}
\end{align}
where $\hat{\mbx} = f(\mbx; \mbw)$ %
is the functional output of the \gls{BAE}.
Although $p_\mbpsi({\xgen})$ cannot be evaluated analytically, it is possible to draw samples from it.

\paragraph{Prior parameterization.} The only two requirements needed to design a parameterization for the prior are: to be able to (1) draw samples from it and (2) to compute its log-density at any point.
The latter is required by many inference algorithms such as \gls{SGHMC}.
We consider a fully-factorized Gaussian prior over weights and biases at layer $l$:
\begin{align}
   p(w_{l}) = \cN(w_{l}; \mu_{l_{w}}, \sigma^{2}_{l_{w}}), \quad  p(b_{l}) = \cN(b_{l}; \mu_{l_{b}}, \sigma^{2}_{l_{b}}),
\end{align}
Notice that, as we shall see in \cref{sec:wd} and \cref{sec:summary-wd}, in order to estimate our prior hyper-parameters, we will require gradient back-propagation through the stochastic variables $w_{l}$ and $b_{l}$.
Thus, we treat these parameters in a deterministic manner by means of the %
 reparameterization trick \cite{rezende2014stochastic, Kingma14}. %

\subsection{Another route for Bayesian Occam's razor}
A common way to estimate hyper-parameters (i.e., prior parameters $\mbpsi$) is to rely on the Bayesian Occam's razor (a.k.a. \textit{empirical Bayes}), which dictates that the marginal likelihood $p_\mbpsi(\mbx)$ should be optimized with respect to $\mbpsi$.
There are countless examples where such simple procedure succeeds in practice \citep[see, e.g.,][]{Rasmussen06}.
The marginal likelihood is obtained by marginalizing out the outputs $\xgen$ and the model parameters $\mbw$,
\begin{align} \label{eq:marg:lik}
   p_\mbpsi(\xobs) = \int p(\xobs\g\xgen) p_\mbpsi(\xgen) \dd \xgen\,,
\end{align} 
where  $p(\xobs\g\xgen)$  and  $p_\mbpsi(\xgen)$  are given by  \cref{eq:continuous_bernoulli_likelihod} and \cref{eq:func_prior}, respectively. 
Unfortunately, in our context it is impossible to carry out this optimization due to the intractability of \cref{eq:marg:lik}.

Classic results in the statistics literature draw parallels between \gls{MLE} and \gls{KL} minimization \cite{Akaiko1973},
\begin{align}
   \argmax_\mbpsi \int  \pi(\xobs)\log p_\mbpsi(\xobs) \dd\xobs = \argmin_\mbpsi \underbrace{\int   \pi(\xobs) \log \frac{ \pi(\xobs)}{ p_\mbpsi(\xobs)} \dd\xobs}_{\KL{\pi(\xobs)}{p_\mbpsi(\xobs)}}\,,
   \label{eq:mle_is_kl}
\end{align}
where $\pi(\xobs)$ is the true data distribution.
This equivalence provides us with an interesting insight on an alternative view of marginal likelihood optimization as minimization of the divergence between the true data distribution and the marginal $p_\mbpsi(\xobs)$.

This alternative view still does not help us in obtaining a viable optimization strategy, even if we use $\xobs$ to estimate an empirical $\tilde{\pi}(\xobs)$; 
the empirical evaluation and optimization of \gls{KL} divergences is indeed a well-known challenging problem \cite{Flam2017},
although this is possible (for the \gls{KL} or any other $f$-divergence), for example, by leveraging results from convex analysis such as in the  convex minimization framework of \cite{Nguyen2010}.
However, we can now attempt to replace the intractable \gls{KL} divergence with another divergence to recover tractability. 
Inspired by recent works on deriving sensible priors for Bayesian neural networks \cite{Tran20}, %
we employ the Wasserstein distance, which, as we will see later, can be estimated efficiently using samples only, even for high-dimensional distributions.

To summarize: (1) we would like to do prior selection by carrying out type-II \gls{MLE}; (2) the \gls{MLE} objective is analytically intractable but the connection with \gls{KL} minimization allows us to (3) swap the divergence with the Wasserstein distance, yielding a practical framework for choosing priors.

\subsection{Matching the marginal distribution to the data distribution via Wasserstein distance minimization \label{sec:wd}}
Given the two probability measures $\pi$ and $p_\mbpsi$, both defined on $\bbR^D$ for simplicity, the $p$-Wasserstein distance between $\pi$ and $p_\mbpsi$ is given by 
\begin{align}
   W_{p}^p(\pi, p_\mbpsi) = \inf_{\gamma \in \Gamma(\pi, p_\mbpsi)} \int \|\xobs-\xobs^\prime\|^{p} \gamma(\xobs, \xobs^\prime)\dd\xobs \dd\xobs^\prime\,,
   \label{eq:wassertein_dist}
\end{align}
where $\Gamma (\pi, p_\mbpsi)$ is the set of all possible distributions $\gamma(\xobs,\xobs^\prime)$ such that the marginals are $\pi(\xobs)$ and $p_\mbpsi(\xobs^\prime)$ \cite{villani2008optimal}.
While usually analytically unavailable or computationally intractable, for $D=1$ the distance has a simple closed form solution, that can be easily estimated using samples only \cite{KolouriNSBR19}.

The \acrfull{DSWD} takes advantage of this result by projecting the estimation of distances for high-dimensional distributions into simpler estimation of multiple distances in one dimension.
The projection is done using the Radon transform $\cR$, an operator that maps a generic density function $\varphi$ defined in $\bbR^{D}$ to the set of its integrals over hyperplanes in $\bbR^D$, 
\begin{align}
   \cR \varphi(t, \mbtheta) := \int \varphi(\mbr) \delta (t - \mbr^\top \mbtheta) \dd\mbr\,, \quad \forall t \in \bbR\,,\;\; \forall \mbtheta \in \bbS^{D-1}\,,
   \label{eq:radon_transform}
\end{align}
where $\bbS^{D-1}$ is the unit sphere in $\bbR^D$ and $\delta(\cdot)$ is the Dirac delta \cite{helgason2010integral}.
Using the Radon transform, for a given direction (or \textit{slice}) $\mbtheta$ we can project the two densities $\pi$ and $p_\mbpsi$ into one dimension and we can solve the optimal transport problem in this projected space.
Furthermore, to avoid unnecessary computations, instead of considering all possible directions in $\bbS^{D-1}$, \gls{DSWD} proposes to find the optimal probability measure of slices $\sigma(\mbtheta)$ on the unit sphere $\bbS^{D-1}$,
\begin{align}
   \label{eq:dual_dsw}
   {DSW}_{p}(\pi, p_\mbpsi) := \sup_{\sigma \in \mathbb{M}_C} \Big( \mathbb{E}_{\sigma(\mbtheta)}  W_{p}^p \big( \cR {\pi}(t, \mbtheta), \cR {p_\mbpsi}(t, \mbtheta) \big)  \Big)^{{1}/{p}},
\end{align}
where, for $C > 0$, $\mathbb{M}_C$ is the set of probability measures $\sigma$ such that $\mathbb{E}_{\mbtheta, \mbtheta^\prime\sim \sigma} \big[ \mbtheta^\top\mbtheta^\prime \big] \leq C$ (a constraint that aims to avoid directions to lie in only one small area).
The direct computation of $DSW_p$ in \cref{eq:dual_dsw} is still challenging but  admits an equivalent dual form,
\begin{align}
   \hspace{-1ex} \sup_{h \in \cH} \left\{ \Big( \E_{\bar\sigma(\mbtheta)} \big[ W_{p}^p \big( \cR \pi(t, h(\mbtheta)), \cR p_\mbpsi(t, h(\mbtheta)) \big) \big] \Big)^{{1}/{p}} \hskip-2ex - \lambda_{C} \mathbb{E}_{\mbtheta, \mbtheta^\prime \sim \bar\sigma}\Big[ \big| h (\mbtheta)^{\top} h (\mbtheta^\prime) \big| \Big] \right\} + \lambda_C C \,,
   \label{eq:dsw_duality}
\end{align}
where $\bar\sigma$ is a uniform distribution in $\bbS^{D-1}$, $\cH$ is the set of functions $h:\bbS^{D-1}\rightarrow\bbS^{D-1}$ and $\lambda_C$ is a regularization hyper-parameter.
The formulation in \cref{eq:dsw_duality} is obtained by employing the Lagrangian duality theorem and by reparameterizing $\sigma(\mbtheta)$ as push-forward transformation of a uniform measure in $\bbS^{D-1}$ via $h$.
Now, by parameterizing $h$ using a deep neural network
with parameters $\mbphi$, defined as $h_{\mbphi}$,
\cref{eq:dsw_duality} becomes an optimization problem with respect to the network parameters.
The final step is to approximate the analytically intractable expectations with Monte Carlo integration,
\begin{align}
      & \max_\mbphi \Bigg\{\left[ \frac{1}{K} \sum_{i=1}^{K} \big[ W_{p}^p \big( \cR {\pi}(t, h_{\mbphi}(\mbtheta_i)), \cR {p_\mbpsi}(t, h_{\mbphi}(\mbtheta_i)) \big) \big] \right]^{{1}/{p}} 
      \hskip-3ex - \frac{\lambda_{C}}{K^2} \sum_{\substack{i,j = 1}}^K  |h_{\mbphi}(\mbtheta_i)^{\top} h_{\mbphi}(\mbtheta_j)|\Bigg\} + \lambda_C C \nonumber\,,
\end{align}
with $\mbtheta_i\sim\bar\sigma(\mbtheta)$. %
Finally, we can use stochastic gradient methods to update $\mbphi$ and then use the resulting optima for the estimation of the original distance.
We encourage the reader to check the detailed explanation of this formulation, including its derivation and some practical considerations for implementation, available in the Appendix. 

\subsection{Summary \label{sec:summary-wd}}
We aim at learning the prior on the \gls{BAE} parameters by optimizing the marginal $p_\mbpsi(\xobs)$ obtained after integrating out the weights from the joint $p_\mbpsi(\xobs, \mbw)$.
The connection with \emph{empirical Bayes} and \gls{KL} minimization suggests that we can find the optimal $\mbpsi^\star$ by minimizing 
the \gls{KL} between the true data distribution $\pi(\xobs)$ and the marginal $p_\mbpsi(\xobs)$ .
However, matching these two distributions is non-trivial due to their high dimensionality and the unavailability of their densities. 
To overcome this problem, we propose a sample-based approach using the distributional sliced 2-Wasserstein distance (\cref{eq:dsw_duality}) as objective:
\begin{align}
   \mbpsi^\star = \argmin_{\mbpsi} \Big[ DSW_2 \big( p_\mbpsi(\xobs), \pi(\xobs) \big) \Big].
\end{align}

This objective function is flexible and does not require the closed-form of  either $p_\mbpsi({\xobs})$ or $\pi(\xobs)$.
The only requirement is that we can draw samples from these two distributions.
Note that we can sample from $p_\mbpsi({\xobs})$, by first computing $\xgen$ after sampling from $p_\mbpsi(\mbw)$ and then perturbing the generated $\xgen$ by sampling from the likelihood $p(\xobs\g\xgen)$.
For the continuous Bernoulli likelihood this operation can be implemented by using the  reparameterization form that allows to backpropagate gradients \cite{Loaiza-GanemC19}.

\section{Experiments}

\textbf{Competing approaches.}
We compare our proposal with a wide selection of methods from the literature.
For autoencoding methods, we choose the 
vanilla 
\textbf{\gls{VAE}} \cite{Kingma14}, the \textbf{$\mbbeta$-\gls{VAE}} \cite{HigginsMPBGBML17} and \textbf{\wae} (Wasserstein AE) \cite{tolstikhin2018wasserstein}.
In addition, we consider models with more complex encoders (\textbf{\gls{VAE} + Sylvester flows} \cite{BergHTW18}), generators (\textbf{2-stage \gls{VAE}} \cite{dai2018diagnosing}), and priors (\textbf{\gls{VAE} + VampPrior} \cite{TomczakW18}).
For \celeba we also include a comparison with \glspl{GAN}, with the 
vanilla 
setup of \textbf{\textsc{ns}-\gls{GAN}} \cite{Ian2014, LucicKMGB18} and the more recent \textbf{DiffAugment-\gls{GAN}} \cite{ZhaoLLZ020, KarrasLAHLA20}.
Finally, we also compare against \gls{BAE} with the standard $\mathcal{N}(0, 1)$ {prior}.
Unless otherwise stated, all models---including ours---share the same latent dimensionality ($K=50$).
We defer a more detailed description of these models and architectures to the Appendix.

\textbf{Generative process.}
Differently from \glspl{VAE} and other methods, deterministic and Bayesian \glspl{AE} are not generative models.
To generate new samples with \glspl{BAE} we employ ex-post density estimation over the learned latent space, by fitting a density estimator $p_{\vartheta}(\latent)$ to $\{\latent_i = \mathbb{E}_{p({\mbw}_{\text{enc}} \g  \xobs)}[f_{\text{enc}}(\xinput_i ;  \mbw_{\text{enc}})] \}$.
In this work, we employ a nonparametric model for density estimation based on \acrfull{DPMM} \cite{Blei06}, so that its complexity is automatically adapted to the data; see also \cite{Bengioetal2013} for alternative ways to turn \glspl{AE} into generative models.
After estimating $p_{\vartheta}(\latent)$, a new sample can be generated by drawing $\latent_{\text{new}}$ from $p_{\vartheta}(\latent)$ and $\xgen_{\text{new}} = \mathbb{E}_{p(\mbw_{\text{dec}} \g  \xobs) }[f_{\text{dec}}({\latent_\text{new}} ; \mbw_{\text{dec}})]$. %

\textbf{Evaluation metrics.}
To evaluate the reconstruction quality, we use the test \gls{LL},
which tells us how likely the test targets are generated by the corresponding model.
The predictive log-likelihood is a proper scoring rule that depends on both the accuracy of predictions and their uncertainty \cite{gneiting2007strictly}.
To assess the quality of the generated images, instead, we employ the widely used \gls{FID} \cite{HeuselRUNH17}.
We note that, as \glspl{GAN} are not inherently equipped with an explicit likelihood model, we only report  their \gls{FID} scores.
Finally, all our experiments and evaluations are repeated four times, with different random training splits.

\subsection{Analysis of the effect of the prior}
To demonstrate the effect of our model selection strategy, we consider scenarios %
in the small-data regime 
where the prior might not be necessarily tuned on the training set. %
In this way we are able to impose inductive bias beyond what is available in the training data. 
We investigate two cases:
\begin{itemize}[noitemsep,topsep=-3pt,leftmargin=15pt]
   \item \mnist \cite{lecun1998gradient}: We use $100$ examples of the $0$ digits to tune the prior.
         The training set consists of examples of $1$-$9$ digits, whereas the test set contains $10\,000$ instances of all digits.
         We aim to demonstrate the ability of our approach to incorporate prior knowledge about completely unseen data with different characteristics into the model.
   \item \freyyale \cite{DaiDGL15}: We use $1\,956$ examples of \frey faces to optimize the prior.
         The training set and test set are comprised of \yale faces.
         We demonstrate the benefit of using a different dataset but from the same domain (e.g. face images) to specify the prior distribution.
\end{itemize}

\begin{figure}[t]
   \centering
   \begin{minipage}{.73\textwidth}
      \hspace{-2ex}
          \centering
    \scalebox{.99}
    {\setlength\tabcolsep{0.5pt}
         {
            \tikzexternaldisable
            \tiny
            \fontfamily{phv}\selectfont

          \begin{tabular}{ r c c c c}
             \toprule
             & \multicolumn{2}{c}{\mnist (N\,=\,200)} & \multicolumn{2}{c}{\freyyale (N\,=\,500)}                                                                                                                                                                    \\
             \cmidrule(r){2-3} \cmidrule(r){4-5}
              & {Reconstructed}                                                                  & {Generated}             & {Reconstructed} & {Generated}                                                     \\
             \midrule
 
             \raisebox{5pt}{Ground Truth}
              & \includegraphics[clip,width=0.205\linewidth]{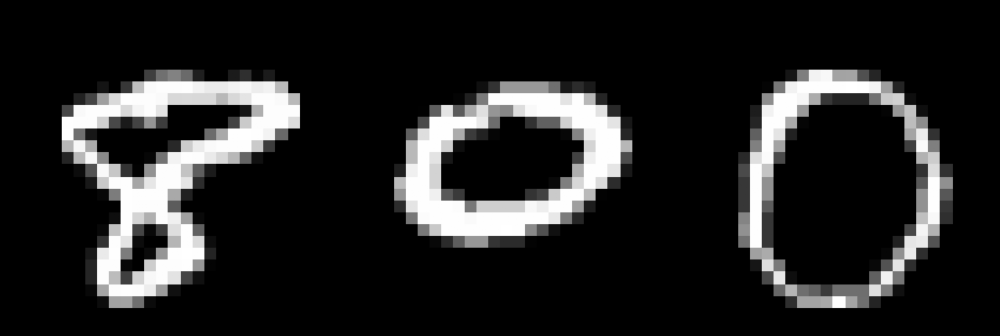} & 
              & \includegraphics[clip,width=0.205\linewidth]{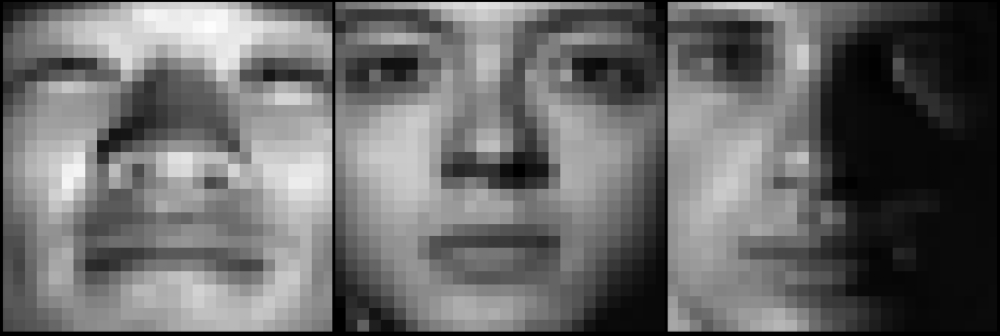}  &   \\[-3.5pt]

             \raisebox{5pt}{VAE}
              & \includegraphics[clip,width=0.205\linewidth]{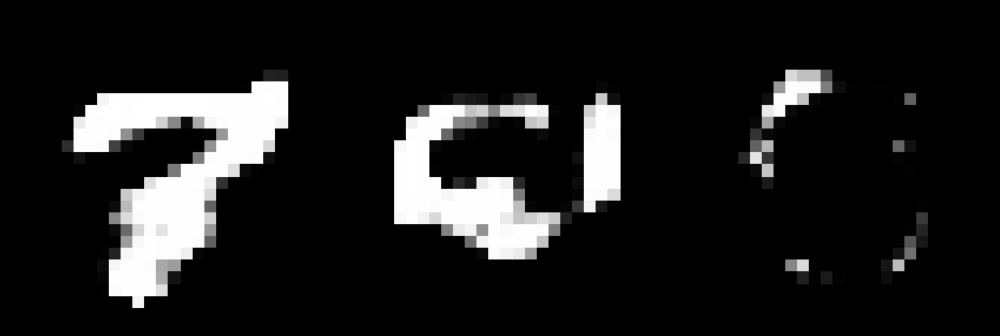} & \includegraphics[clip,width=0.205\linewidth]{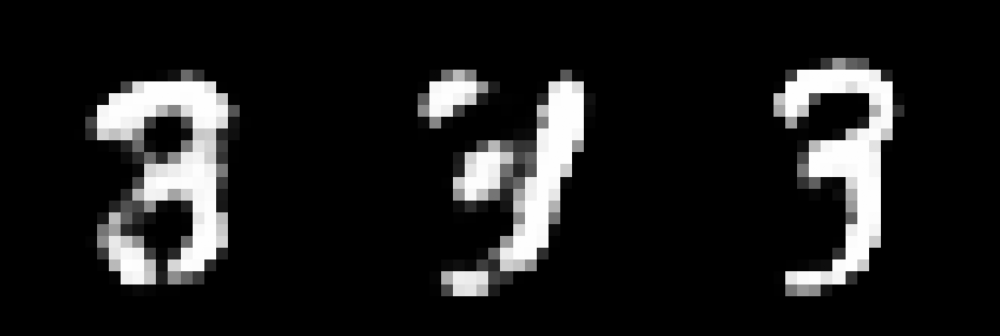}
              & \includegraphics[clip,width=0.205\linewidth]{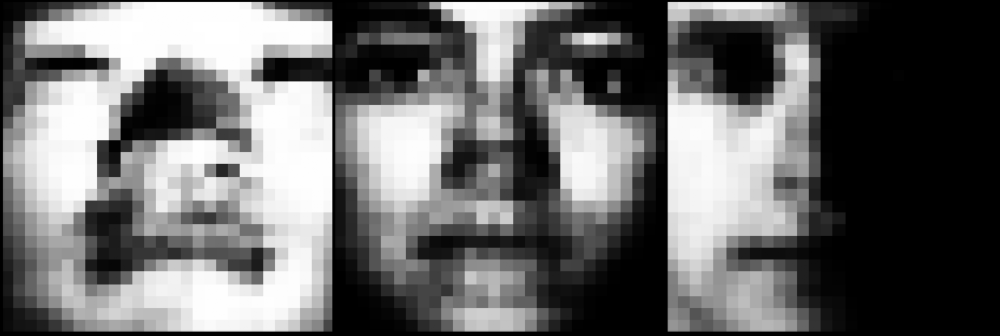}  & \includegraphics[clip,width=0.205\linewidth]{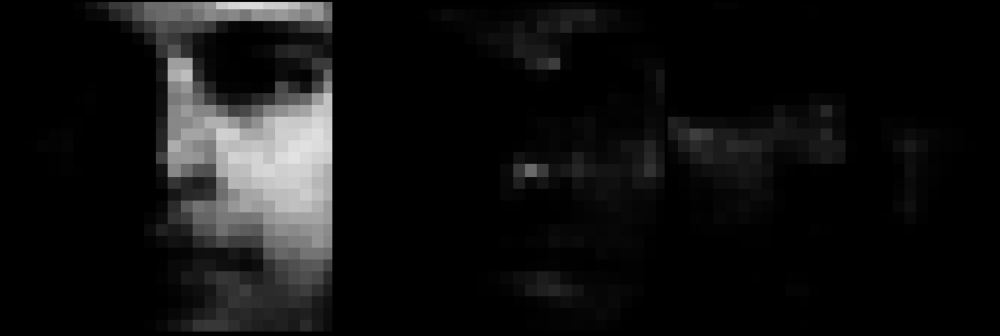}  \\[-3.5pt]
 
              \raisebox{5pt}{\textcolor{black}{$\bigstar$} VAE}
              & \includegraphics[clip,width=0.205\linewidth]{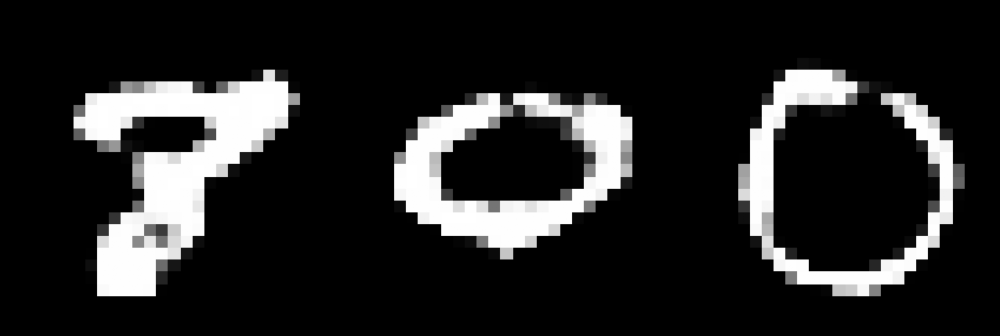} & \includegraphics[clip,width=0.205\linewidth]{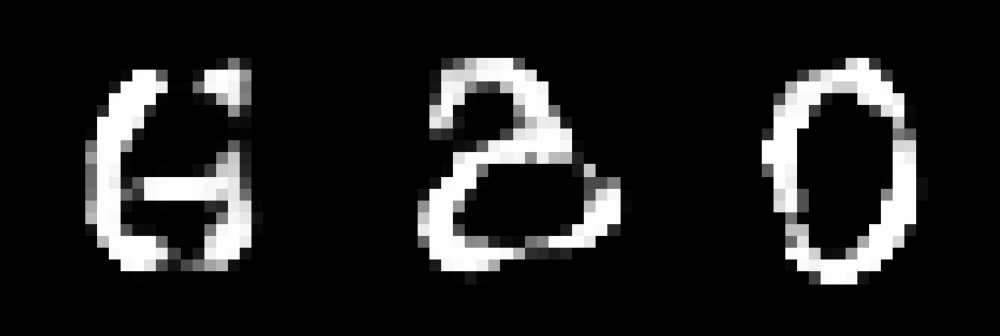}
              & \includegraphics[clip,width=0.205\linewidth]{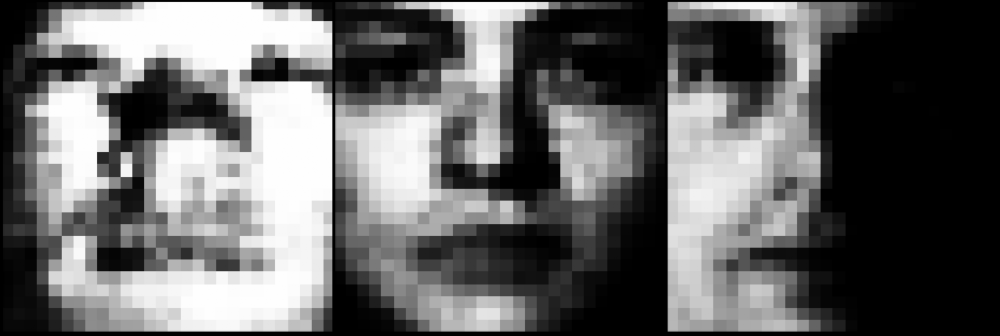}  & \includegraphics[clip,width=0.205\linewidth]{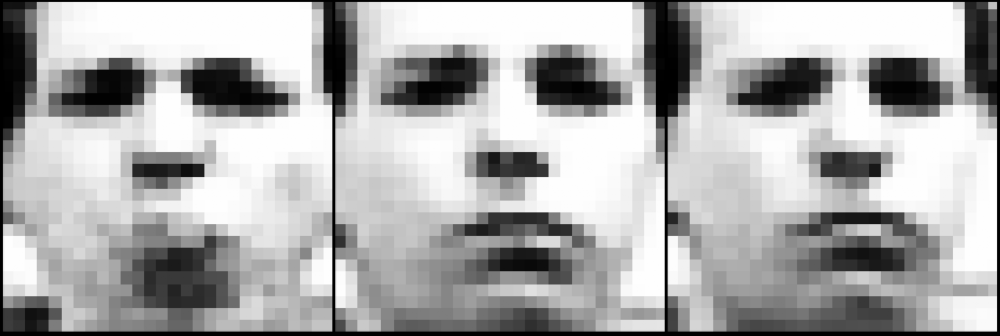}  \\[-3.5pt]

              \raisebox{5pt}{$\mathcal{N}(0, 1)$ BAE}
               & \includegraphics[clip,width=0.205\linewidth]{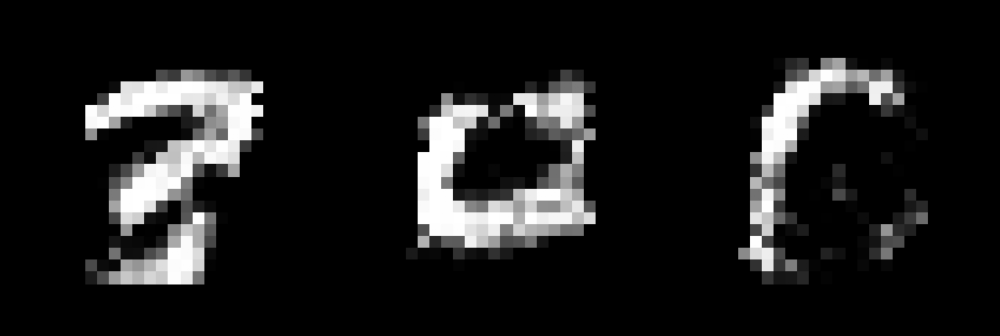} & \includegraphics[clip,width=0.205\linewidth]{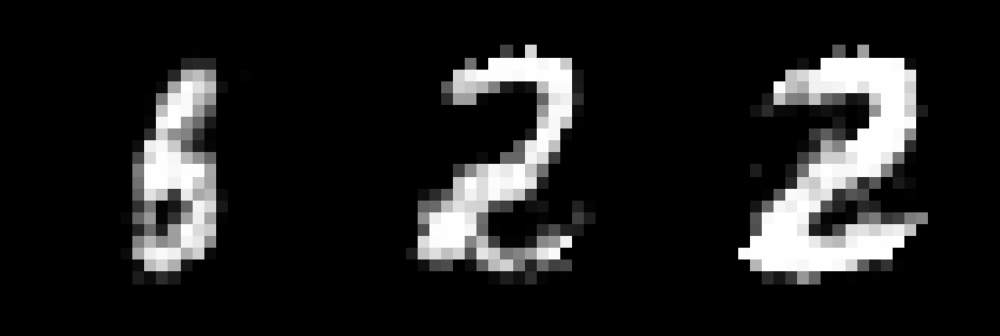}
               & \includegraphics[clip,width=0.205\linewidth]{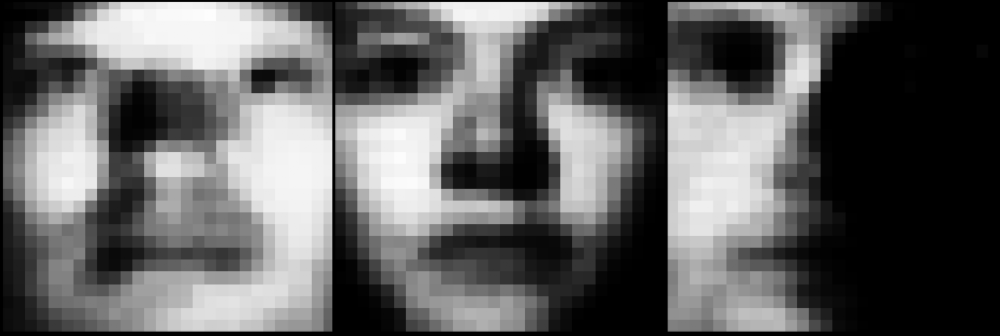}  & \includegraphics[clip,width=0.205\linewidth]{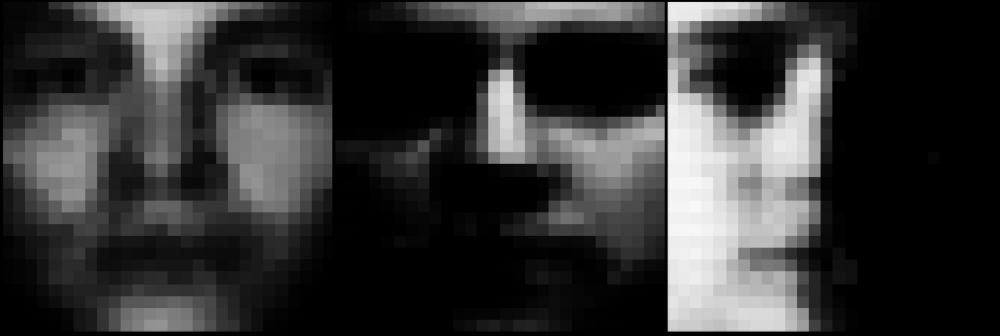}  \\[-3.5pt]
  
               \raisebox{5pt}{\textcolor{black}{$\bigstar$} $\mathcal{N}(0, 1)$ BAE}
               & \includegraphics[clip,width=0.205\linewidth]{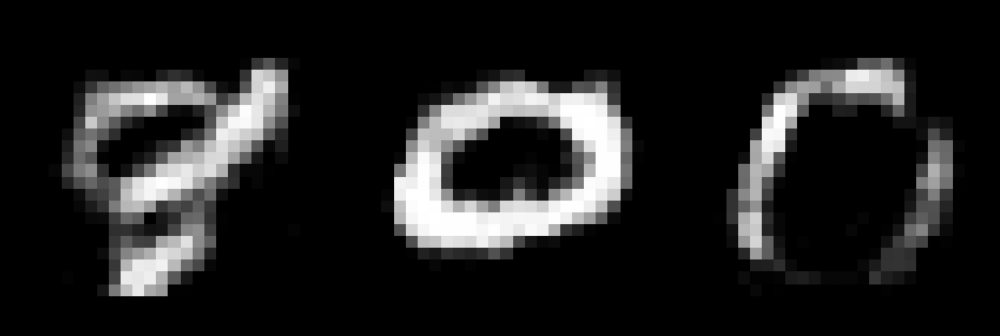} & \includegraphics[clip,width=0.205\linewidth]{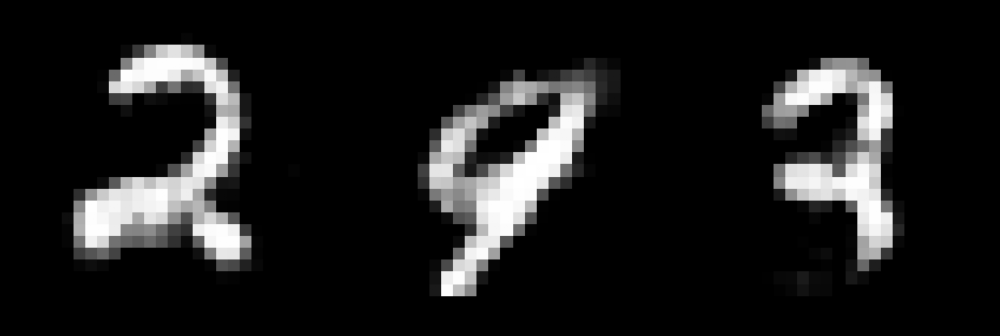}
               & \includegraphics[clip,width=0.205\linewidth]{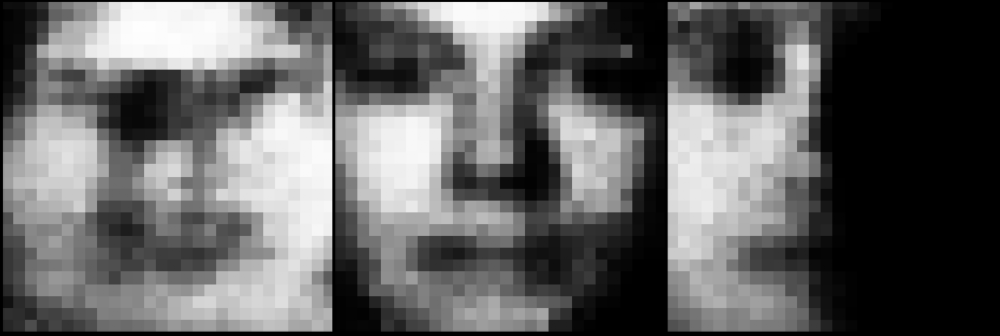}  & \includegraphics[clip,width=0.205\linewidth]{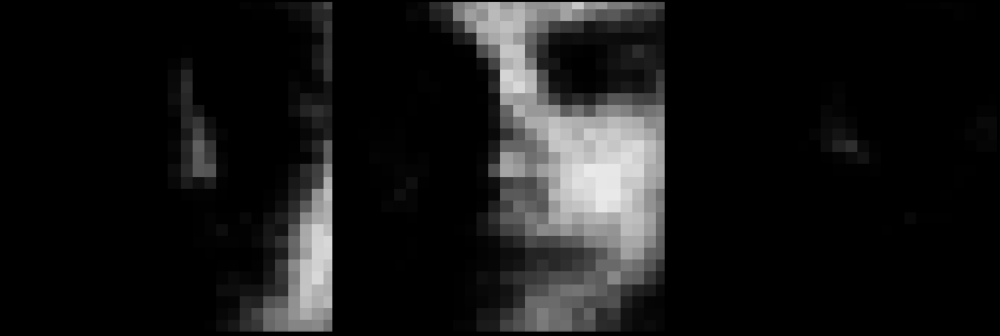}  \\[-3.5pt]
 
              \raisebox{5pt}{\textit{BAE + Optim. prior}}
               & \includegraphics[clip,width=0.205\linewidth]{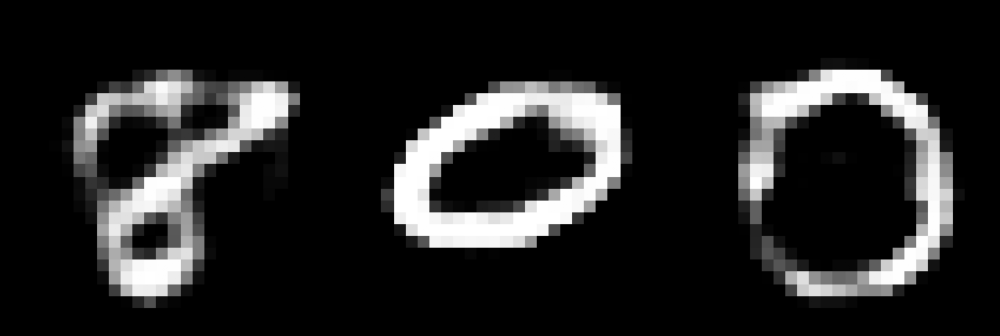} & \includegraphics[clip,width=0.205\linewidth]{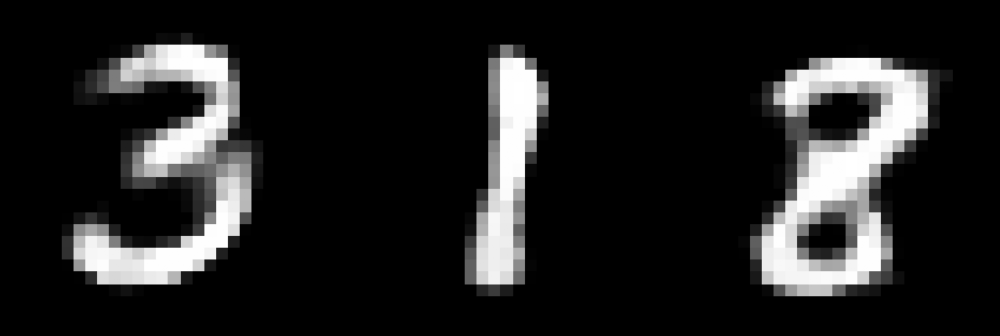}
               & \includegraphics[clip,width=0.205\linewidth]{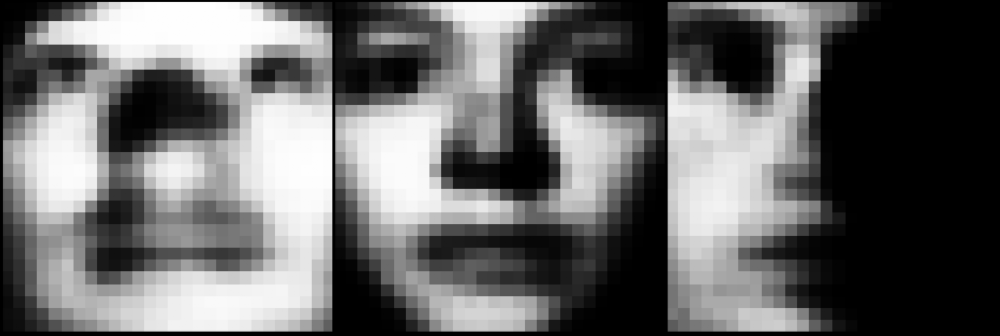}  & \includegraphics[clip,width=0.205\linewidth]{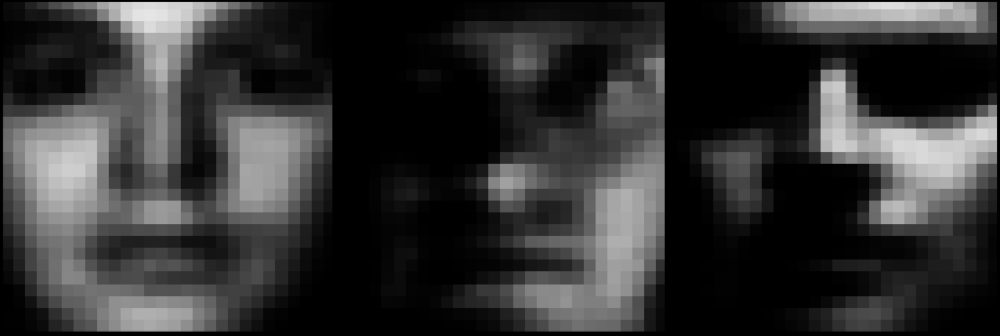}  \\[-3.5pt]
 
               \raisebox{5pt}{\textit{Uncertainty}}
                & \includegraphics[clip,width=0.205\linewidth]{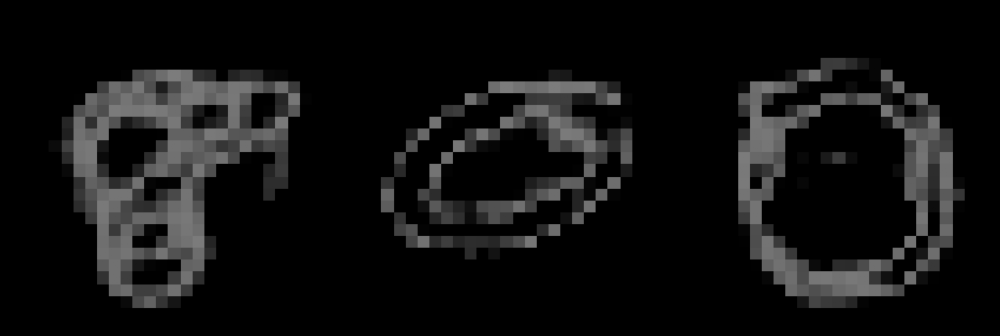} & \includegraphics[clip,width=0.205\linewidth]{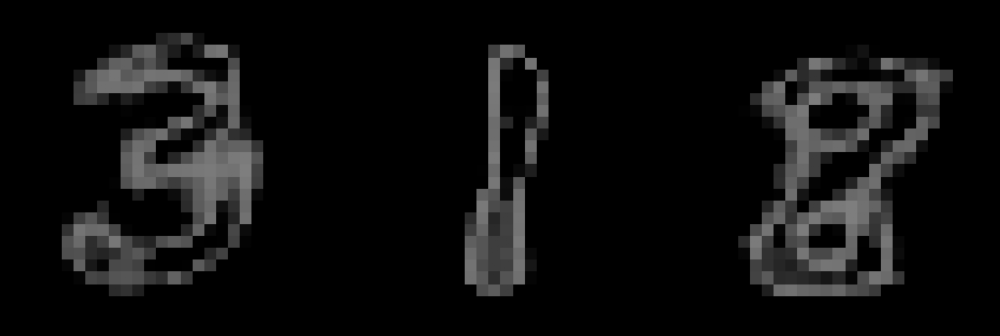}
                & \includegraphics[clip,width=0.205\linewidth]{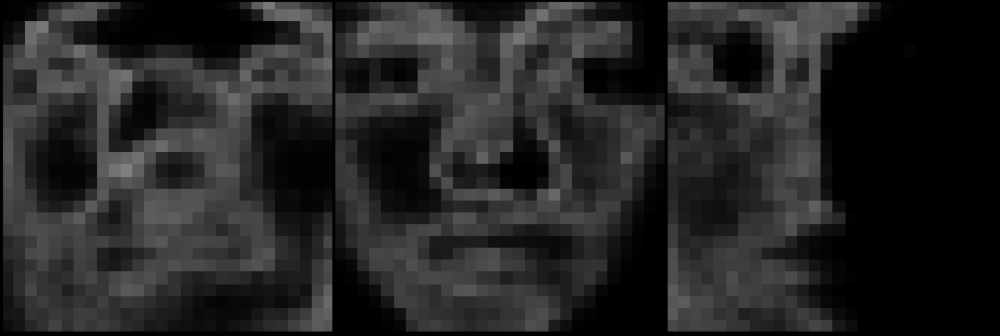}  & \includegraphics[clip,width=0.205\linewidth]{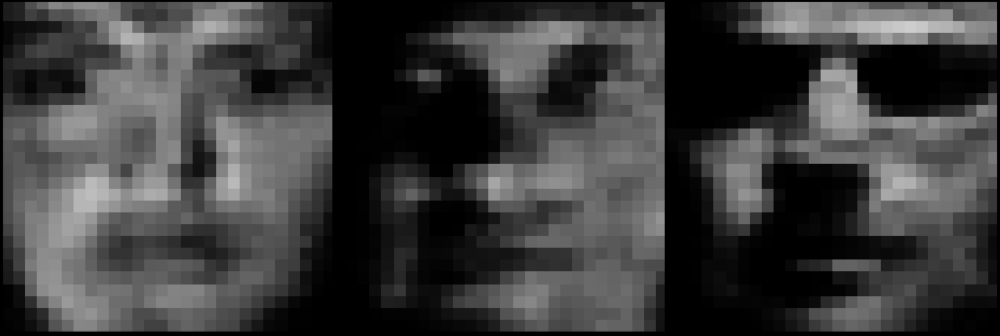}  \\[-3.5pt]

             \bottomrule
          \end{tabular}
          \tikzexternalenable
      }
       }

      \captionof{figure}{Qualitative evaluation for \mnist and \yale.
         Here, \textcolor{black}{$\bigstar$} indicates using the union of the training data and the data used to optimize prior to train the model.
         The last row depicts standard deviation of reconstructed/generated images estimated by \gls{BAE} using the
         optimized prior.
         \label{fig:mnist_face_results_pictures}
      }
   \end{minipage}%
   \hfill%
   \begin{minipage}{.25\textwidth}
      \centering
      \tiny
      \setlength{\figurewidth}{3.8cm}
      \setlength{\figureheight}{3.2cm}
      \includegraphics{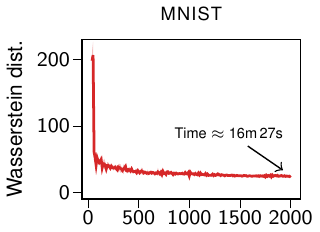}\\[1ex]
      \includegraphics{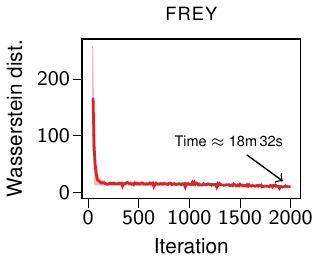}
      \captionof{figure}{
         Convergence of the proposed Wasserstein minimization scheme.
         \label{fig:covergence_wasserstein}
      }
   \end{minipage}
\end{figure}

\begin{figure}[t]
   \centering
   \tiny
   \setlength{\figurewidth}{3.6cm}
   \setlength{\figureheight}{3.2cm}
   \begin{tabular}{cc|c}
      \includegraphics{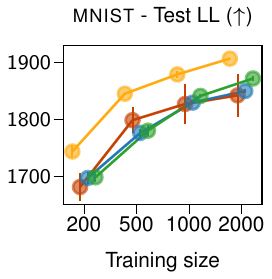} &
      \includegraphics{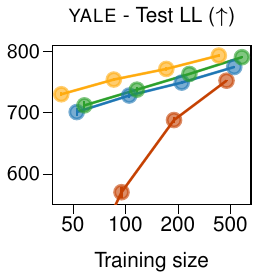}  &
      \setlength{\figurewidth}{7cm}%
      \setlength{\figureheight}{3.2cm}%
      \includegraphics{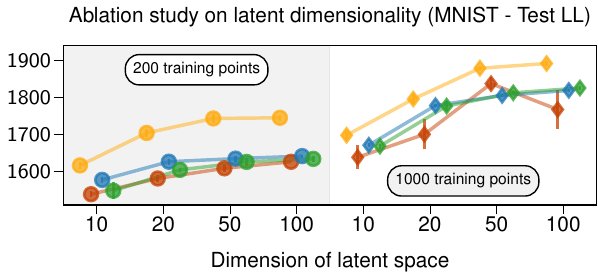}
   \end{tabular}
   \\[2ex]
   
    \definecolor{vae_combined}{rgb}{0.12156862745098,0.466666666666667,0.705882352941177}
    \definecolor{vae_beta_combined}{rgb}{0.172549019607843,0.627450980392157,0.172549019607843}
    \definecolor{bae_optim}{HTML}{ffab0f}
    \definecolor{bae_std_combined}{HTML}{c54102}
\tikzexternaldisable
\fontfamily{phv}\selectfont

 \tikzexternaldisable \setlength\tabcolsep{1pt} \begin{tabular}{lp{2ex}lp{2ex}lp{2ex}l}
    \toprule
    {\protect\tikz[baseline=-1ex]\protect\draw[thick, color=vae_combined, fill=vae_combined, mark=*, opacity=0.6, mark size=2.0pt, line width=0.8pt] plot[] (-.0, 0)--(.25,0)--(-.25,0);}  {VAE $\bigstar$} & & 
    {\protect\tikz[baseline=-1ex]\protect\draw[thick, color=vae_beta_combined, fill=vae_beta_combined, mark=*, opacity=0.6, mark size=2.0pt, line width=0.8pt] plot[] (-.0, 0)--(.25,0)--(-.25,0);}  {$\beta$-VAE $\bigstar$}  &&
    {\protect\tikz[baseline=-1ex]\protect\draw[thick, color=bae_std_combined, fill=bae_std_combined, mark=*, opacity=0.6, mark size=2.0pt, line width=0.8pt] plot[] (-.0, 0)--(.25,0)--(-.25,0);}  {BAE + $\mathcal{N}(0,1)$ Prior $\bigstar$} & &
    {\protect\tikz[baseline=-1ex]\protect\draw[thick, color=bae_optim, fill=bae_optim, mark=*, opacity=0.6, mark size=2.0pt, line width=0.8pt] plot[] (-.0, 0)--(.25,0)--(-.25,0);}  {BAE + Optim. Prior} \\
    \bottomrule
\end{tabular}\tikzexternalenable
   \caption{Test \acrfull{LL} of \mnist and \yale.
      \textit{Left:} test \gls{LL} as a function of training size;
      \textit{Right:} test \gls{LL} as a function of latent dimensionality.
      \label{fig:mnist_face_results_plots}}
\end{figure}

\textbf{Visual inspection.}
\cref{fig:mnist_face_results_pictures} shows some qualitative results (additional images are available in the Appendix),  while \cref{fig:covergence_wasserstein} shows the convergence of the Wasserstein distance during prior optimization in our proposal.
From a visual inspection we see that, on \mnist, by encoding knowledge about the ``$0$'' digit into the prior, the \gls{BAE} can reconstruct this digit fairly well although we only use ``$1$'' to ``$9$'' digits for inference (differently from the \gls{BAE} with standard prior).
Similarly, on \freyyale, we see that  by encoding knowledge from another dataset in the same domain, the optimized prior can impose a softer constraint compared to using directly this dataset for inference.
In addition, if we use directly the union of \frey and \yale faces for training (methods denoted with a $\bigstar$),
\gls{VAE} yields images that are similar to \frey instead of \yale faces,
while generated images from \gls{BAE} with $\cN(0,1)$ prior are of lower quality.
This again highlights the advantage of our approach to specifying an informative prior compared to using that data for training.
Another important benefit of our Bayesian treatment of \glspl{AE} is that we can quantify the \emph{uncertainty} for both reconstructed and generated images.
The last row of \cref{fig:mnist_face_results_pictures} illustrates the uncertainty estimate corresponding to the \gls{BAE} with optimized prior on \mnist and \yale datasets.
Our model exhibits increased uncertainty for semantically and visually challenging pixels such as the left part of the second ``$0$'' digit image in the \mnist example.
We also observe that the uncertainty is greater for generated images compared to reconstructed images as illustrated in the \yale example.
This is reasonable because the reconstruction process is guided by the input data rather than synthesizing new data according to a random latent code.

\begin{minipage}{.63\textwidth}
   \textbf{Visualization of inductive bias on MNIST.}
   To have an intuition of the inductive bias induced by the optimized prior, we visualize a low-dimensional projection of parameters sampled from the prior and the posterior \cite[]{IzmailovMKGVW19}.
   As we see in \cref{fig:mnist_conv_subspace}, the hypothesis space induced by the $\mathcal{N}(0,1)$ prior is huge, compared to where the true solution should lie. 
   Effectively this is another visualization of the famous Bayesian Occam's razor plot by David MacKay \cite[]{MacKay03}, where the model has very high complexity and poor inductive biases. 
   On the other hand, by considering our proposal to do \textit{model selection}, the hypothesis space of the optimized prior is reduced to regions close to the full posterior. %
   Additional visualizations are available in the Appendix.

\end{minipage}
\hfill
\begin{minipage}{.34\textwidth}
   \centering
   \tiny
   \setlength{\figurewidth}{3.7cm}
   \setlength{\figureheight}{3.7cm}
   \hspace{-3ex}
   \includegraphics{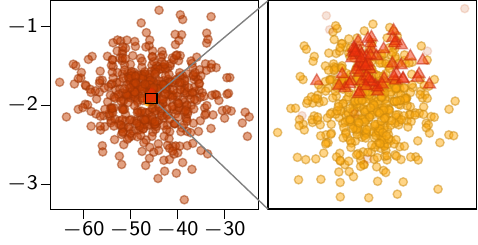}
   {
      \tiny
      \definecolor{color_0}{rgb}{0.99609375,0.49609375,0.0546875}
\definecolor{color_1}{rgb}{0.8359375,0.15234375,0.15625}
\definecolor{color_2}{rgb}{0.12109375,0.46484375,0.703125}
\definecolor{color_3}{rgb}{0.578125,0.40234375,0.73828125}
\definecolor{color_4}{rgb}{0.171875,0.625,0.171875}

\definecolor{color_2}{HTML}{c54102}
\definecolor{color_0}{HTML}{ffab0f}
\definecolor{color_4}{HTML}{f4320c}

\tikzexternaldisable
{\setlength{\tabcolsep}{1.8pt}
    \begin{tabular}{lll}
        \toprule
        {\protect\tikz[baseline=-1ex]\protect\draw[color=color_2, fill=color_2, mark=*, opacity=0.5, mark size=1.7pt, line width=0.0pt] plot[] (-0.1,0);}  \textsf{$\mathcal{N}(0,1)$ Prior} &   &
        {\protect\tikz[baseline=-1ex]\protect\draw[color=color_4, fill=color_4, mark=triangle*, opacity=0.5, mark size=1.7pt, line width=0.0pt] plot[] (-0.1,0);}  \textsf{Samples from true posterior}
        \\
        {\protect\tikz[baseline=-1ex]\protect\draw[color=color_0, fill=color_0, mark=*, opacity=0.5, mark size=1.7pt, line width=0.0pt] plot[] (-0.1,0);}  \textsf{Optim. Prior}             &
        \\
        \bottomrule
    \end{tabular}\tikzexternalenable
}
   }
   \captionof{figure}{Visualization in 2D of samples from priors and posteriors of \gls{BAE}'s parameters. The setup is the same as before with \mnist. \label{fig:mnist_conv_subspace}
   }

\end{minipage}

\textbf{Quantitative evaluation.}
For a quantitative analysis we rely on \cref{fig:mnist_face_results_plots}, where we study the effect on the reconstruction quality of different training sizes (on the \textit{left}) and different latent dimensions (on the \textit{right}).
Since we observed that the results of \gls{VAE} variants are not significantly different, we only show the results for $\beta$-\gls{VAE} and we leave the extended results to the Appendix.
From this experiment we can draw important conclusions.
The \gls{BAE} with optimized priors clearly outperforms the competing methods (and the \gls{BAE} with standard prior) in the inference task for all training sizes, with slightly diminishing effect for larger sets, as expected.
Also, this pattern is true when looking at different latent dimensions (\cref{fig:mnist_face_results_plots}, \textit{left}), where regardless of the dimensionality of the latent space, \glspl{BAE} with optimized priors deliver higher performances.

\begin{figure}[t]
   \begin{minipage}{.64\textwidth}
      \hspace{-11ex}
      \centering
\scalebox{.99}
{\setlength\tabcolsep{1pt}
    {

        \definecolor{color0}{HTML}{0b7734}
        \definecolor{color1}{HTML}{37a055}
        \definecolor{color2}{HTML}{75c477}
        \definecolor{color3}{HTML}{105ba4}
        \definecolor{color4}{HTML}{3787c0}
        \definecolor{color5}{HTML}{6caed6}
        \definecolor{color6}{HTML}{c54102}
        \definecolor{color7}{HTML}{ffab0f}

        \tikzexternaldisable
        \tiny
        \fontfamily{phv}\selectfont
        \begin{tabular}{ rr c c}
            \toprule
                                                                              &                                                                                                                                                                                                     & {Reconstructions}                                                                      & {Generated Samples} \\
            \midrule
            \raisebox{7pt}{Ground Truth}                                      &
                                                                              & \includegraphics[clip,width=0.28\linewidth]{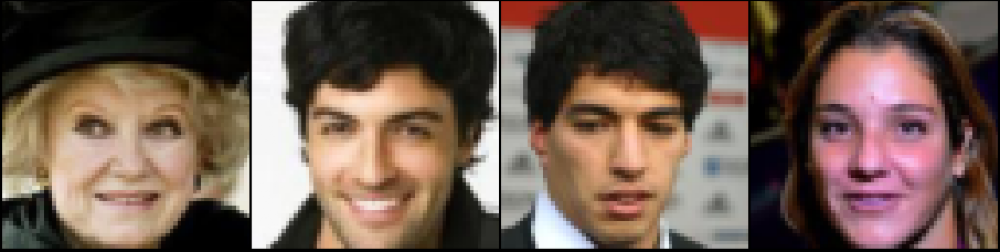}
            \\[-3.3pt]

            \raisebox{7pt}{ WAE \cite{tolstikhin2018wasserstein}}   & {\protect\tikz[baseline=-1ex]\protect\draw[thick, draw=color0, fill=color0!70, mark=*, draw opacity=1, mark size=2.2pt, line width=1pt] plot[] (.3, 0.2);}
                                                                              & \includegraphics[clip,width=0.28\linewidth]{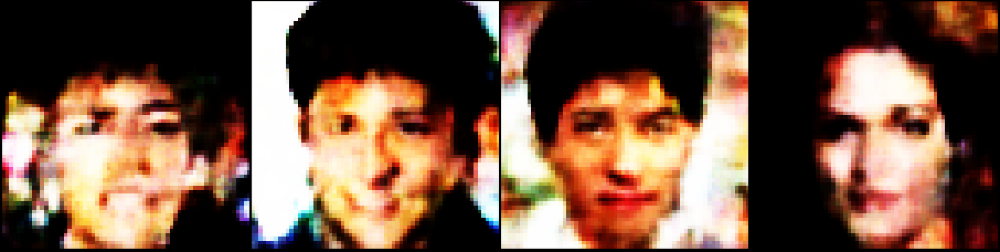}                                                                                                                     & \includegraphics[clip,width=0.28\linewidth]{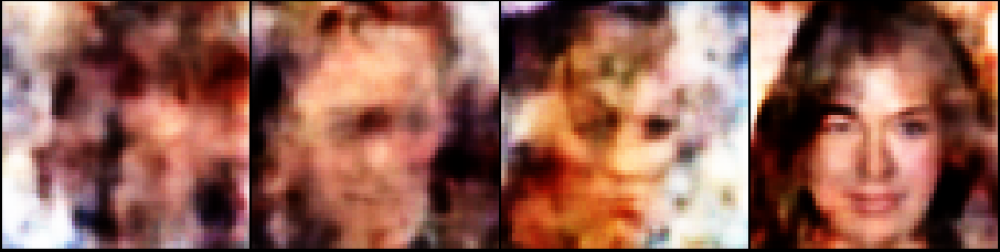}                                 \\[-2.5pt]

            \raisebox{7pt}{ VAE \cite{Kingma14}}                    & {\protect\tikz[baseline=-1ex]\protect\draw[thick, draw=color1, fill=color1!70, mark=*, draw opacity=1, mark size=2.2pt, line width=1pt] plot[] (.3, 0.2);}
                                                                              & \includegraphics[clip,width=0.28\linewidth]{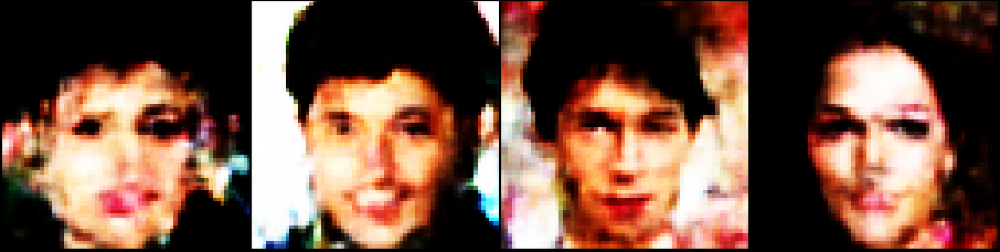}                                                                                                                     & \includegraphics[clip,width=0.28\linewidth]{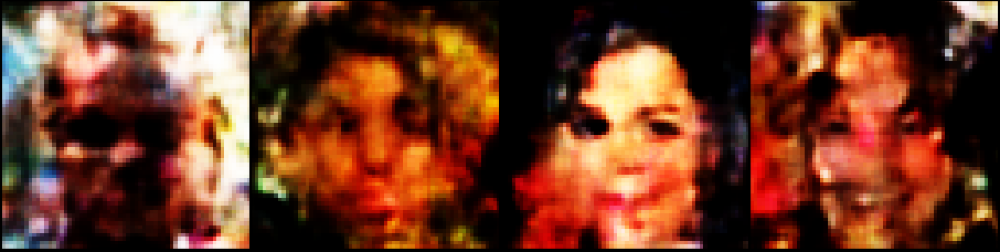}                                 \\[-2.5pt]

            \raisebox{7pt}{ $\beta$-VAE \cite{HigginsMPBGBML17}}    & {\protect\tikz[baseline=-1ex]\protect\draw[thick, draw=color2, fill=color2!70, mark=*, draw opacity=1, mark size=2.2pt, line width=1pt] plot[] (.3, 0.2);}
                                                                              & \includegraphics[clip,width=0.28\linewidth]{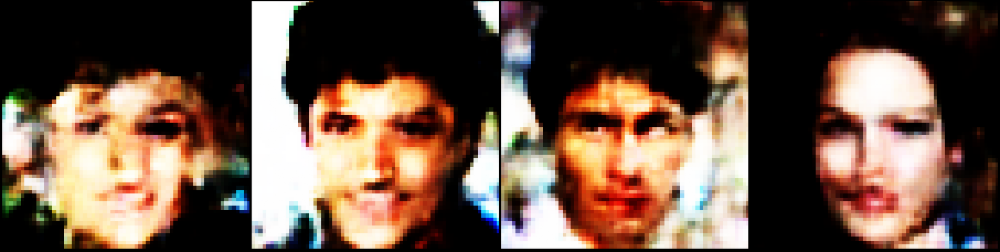}                                                                                                                & \includegraphics[clip,width=0.28\linewidth]{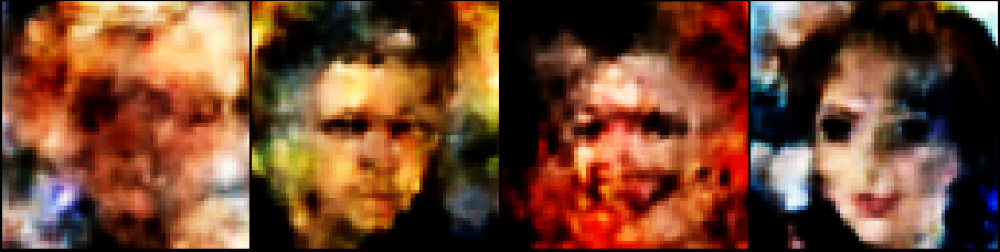}                            \\[-3.5pt]

            \raisebox{7pt}{ VAE + Sylveser Flows \cite{BergHTW18} } & {\protect\tikz[baseline=-1ex]\protect\draw[thick, draw=color3, fill=color3!70, mark=*, draw opacity=1, mark size=2.2pt, line width=1pt] plot[] (.3, 0.2);}
                                                                              & \includegraphics[clip,width=0.28\linewidth]{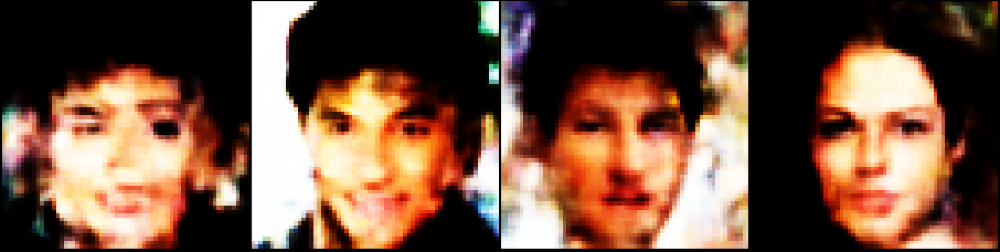}                                                                                                           & \includegraphics[clip,width=0.28\linewidth]{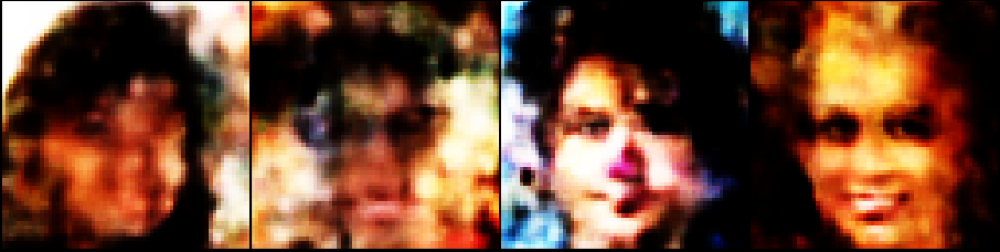}                       \\[-2.5pt]

            \raisebox{7pt}{ VAE + VampPrior \cite{TomczakW18}}      & {\protect\tikz[baseline=-1ex]\protect\draw[thick, draw=color4, fill=color4!70, mark=*, draw opacity=1, mark size=2.2pt, line width=1pt] plot[] (.3, 0.2);}
                                                                              & \includegraphics[clip,width=0.28\linewidth]{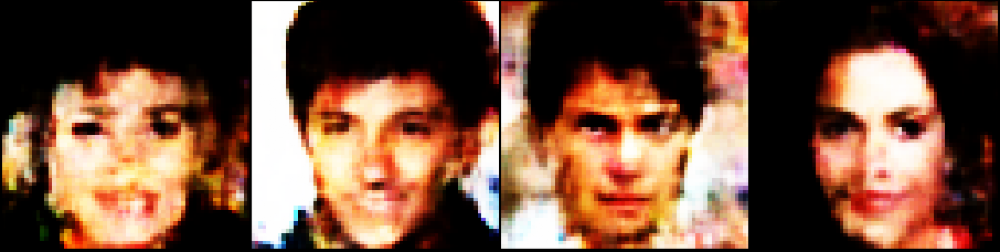}                                                                                                                & \includegraphics[clip,width=0.28\linewidth]{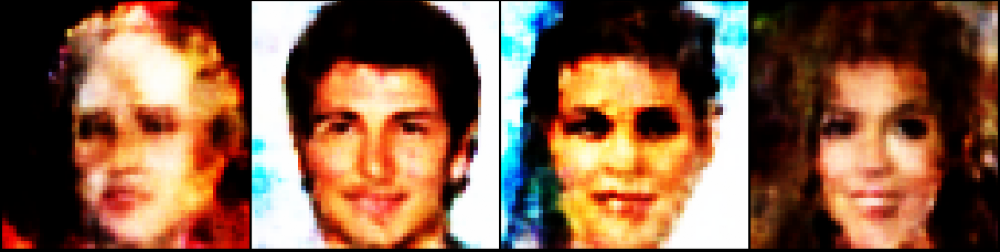}                            \\[-2.5pt]

            \raisebox{7pt}{ 2-Stage VAE \cite{dai2018diagnosing}}   & {\protect\tikz[baseline=-1ex]\protect\draw[thick, draw=color5, fill=color5!70, mark=*, draw opacity=1, mark size=2.2pt, line width=1pt] plot[] (.3, 0.2);}
                                                                              & \includegraphics[clip,width=0.28\linewidth]{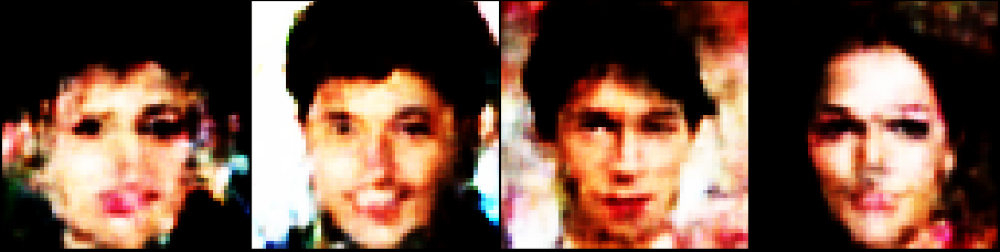}                                                                                                              & \includegraphics[clip,width=0.28\linewidth]{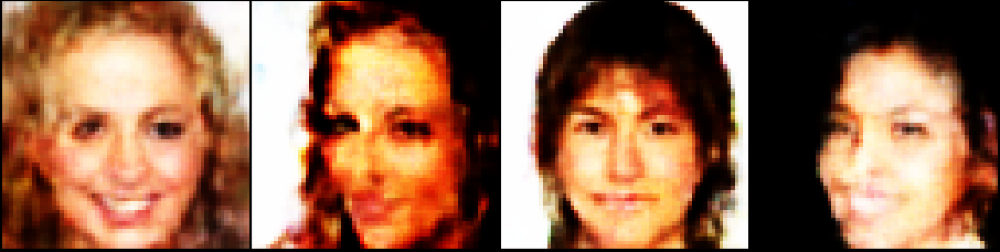}                          \\[-2.5pt]

            \raisebox{7pt}{ BAE + $\mathcal{N}(0,1)$ Prior }        & {\protect\tikz[baseline=-1ex]\protect\draw[thick, draw=color6, fill=color6!70, mark=*, draw opacity=1, mark size=2.2pt, line width=1pt] plot[] (.3, 0.2);}
                                                                              & \includegraphics[clip,width=0.28\linewidth]{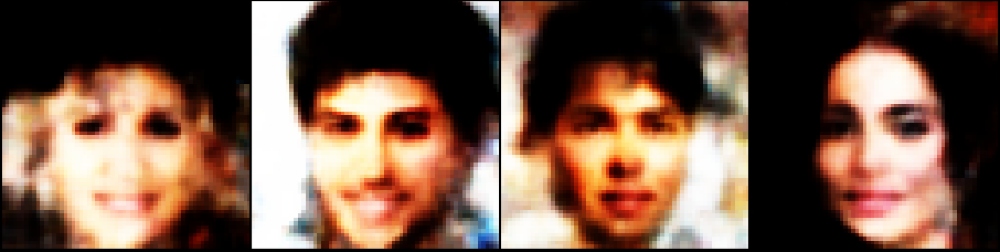}                                                                                                                 & \includegraphics[clip,width=0.28\linewidth]{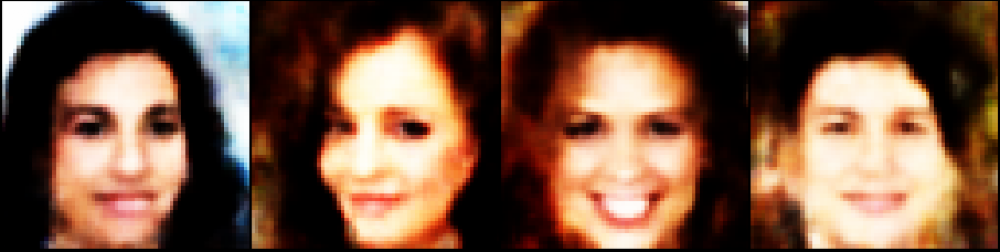}                             \\[-2.5pt]

            \raisebox{7pt}{BAE + Optim. Prior (\textbf{Ours})}                & {\protect\tikz[baseline=-1ex]\protect\draw[thick, draw=color7, fill=color7!70, mark=*, draw opacity=1, mark size=2.2pt, line width=1pt] plot[] (.3, 0.2);}
                                                                              & \includegraphics[clip,width=0.28\linewidth]{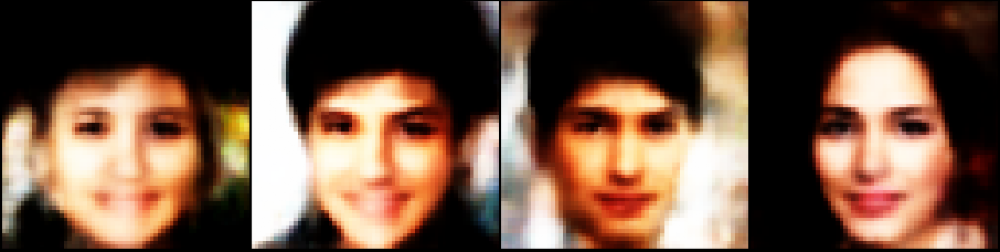}
                                                                              & \includegraphics[clip,width=0.28\linewidth]{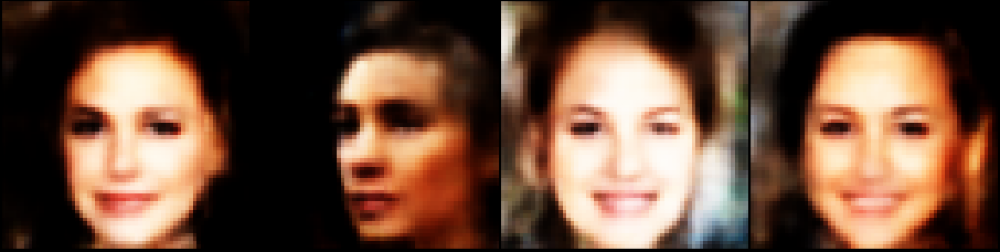}                                                                                                                                                                                                                             \\[-2.5pt]
            \raisebox{7pt}{ NS-GAN \cite{Ian2014}}                  & {\protect\tikz[baseline=-1ex]\protect\draw[thick, draw=white!58.8235294117647!black, fill=white!58.8235294117647!black, mark=*, draw opacity=1, mark size=2.2pt, line width=1pt] plot[] (.3, 0.2);}
                                                                              &                                                                                                                                                                                                     & \includegraphics[clip,width=0.28\linewidth]{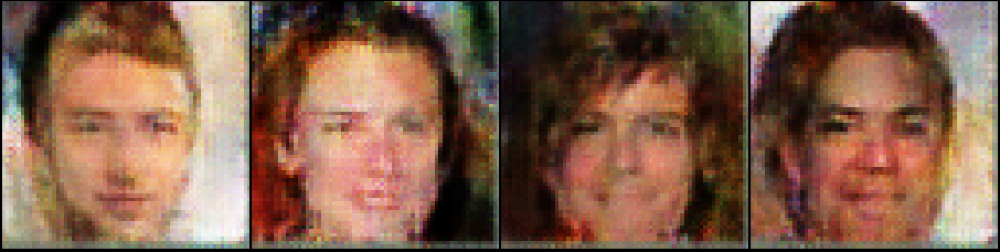}                                 \\[-2.5pt]

            \raisebox{7pt}{ DiffAugment-GAN \cite{ZhaoLLZ020}}      & {\protect\tikz[baseline=-1ex]\protect\draw[thick, draw=white!32.156862745098!black, fill=white!32.156862745098!black, mark=*, draw opacity=1, mark size=2.2pt, line width=1pt] plot[] (.3, 0.2);}
                                                                              &                                                                                                                                                                                                     & \includegraphics[clip,width=0.28\linewidth]{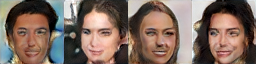}                        \\[-2.5pt]

            \bottomrule
        \end{tabular}
        \tikzexternalenable

    }}

   \end{minipage}
   \begin{minipage}{.34\textwidth}
      \tiny
      \setlength{\figurewidth}{6.3cm}
      \setlength{\figureheight}{4.6cm}
      \hspace{-9.5ex}
      \vspace{-3ex}
      \includegraphics{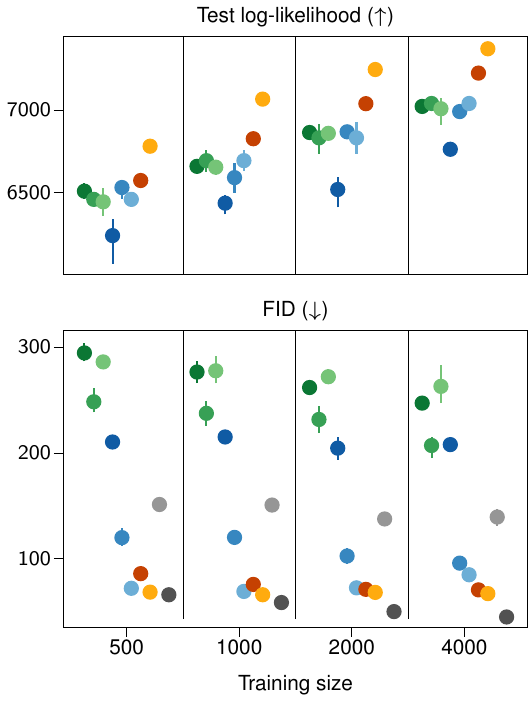}
   \end{minipage}
   \vspace{-1ex}
   \caption{Qualitative (\textit{left}) and quantitative evaluation (\textit{right}) on \celeba.
      The markers and bars represent the means and one standard deviations, respectively.
      In the (\textit{left}) figure, the sizes of training data and the data for optimizing prior are $500$ and $1000$, respectively.
      The higher the \acrfull{LL} and the lower \gls{FID} the better.  }
   \label{fig:celeba_results}
\end{figure}

\subsection{Reconstruction and generation of CELEBA}
We now look at a more challenging benchmark, the \celeba dataset \cite{LiuLWT15}.
For our proposal, we  use $1\,000$ examples that are randomly chosen from the original training set to learn the prior distribution.
The test set consists of about $20\,000$ images.
The goal of this experiment is to evaluate whether sacrificing part of the training data to specify a good prior is beneficial when compared to using that data for training the model.
\cref{fig:celeba_results} shows qualitative results  for the competing methods, their corresponding test \glspl{LL} and
\glspl{FID} for different training dataset sizes.
In terms of test \acrfullpl{LL} (\cref{fig:celeba_results}, \textit{top right}), we observe two clear patterns: (i) that  \gls{BAE} approaches perform considerably better than other methods and (ii) the \gls{VAE} with Sylvester flows performs consistently poor across dataset sizes.
This latter observation indicates that having a more expressive posterior for the encoder is not helpful when considering the small training sizes used in our experiments.
More importantly, we see that the \gls{BAE} using the optimized prior significantly outperforms other methods despite using less data for inference. These results largely agree with the quality of the reconstructions (first column of images in \cref{fig:celeba_results}, \textit{left}) in that \gls{BAE} methods provide more visually appealing reconstructions when compared to other approaches.

We now evaluate the quality of the generated images (second column of images in \cref{fig:celeba_results}, \textit{left}) along with their \gls{FID} scores \cite{HeuselRUNH17}.
Visually, it is clear that images generated from \glspl{VAE} (standard, $\beta$, Sylvester and \gls{WAE}) are very poor.
This failure may originate from the fact that the aggregated posterior distribution of the encoder is not aligned with the prior on the latent space.
This problem is more prominent in the case of small training data, where the encoder is not well-trained.
The VampPrior tackles this problem by explicitly modeling the aggregated posterior, while 2-stage \gls{VAE} uses another \gls{VAE} to estimate the density of the learned latent space.
By reducing the effect of the aggregated posterior mismatch, these strategies improve the quality of the generated images remarkably.
These results are consistent with their corresponding \gls{FID} scores (\cref{fig:celeba_results}, \textit{bottom right}) where
we also see that \gls{BAE} using the optimized prior consistently outperforms all variants of \glspl{VAE} and \textsc{ns-gan}.
Finally, we see that DiffAugment-\gls{GAN}, with the exception of  using a training size of 500,
yields better \gls{FID} scores. However, this is not surprising as this model uses much more complex network architectures  \cite{KarrasLAHLA20}, combined with a powerful differentiable augmentation scheme.
More importantly, it is clear that with few training samples our method generates more semantically meaningful images  then all other approaches, including DiffAugment-\gls{GAN}.

\subsection{Prior adjustment versus posterior tempering}

\begin{minipage}{.62\textwidth}
We have shown that the proposed framework for adjusting the prior is compatible with standard Bayesian practices, as it emulates type-II maximum likelihood.
   In other words, the distribution fitting that we induce by means of Wasserstein distance minimization relates  to the marginal output of \glspl{BAE}, very much in the same spirit of marginal likelihood maximization.
   The distribution is fit considering \emph{all} possible functions, when marginalized through the likelihood, creating an implicit regularization effect.
   Our scheme does not give more weight to particular training instances, but it simply restricts the hypothesis space.
   This is unlike \emph{posterior tempering} \cite{ZhangSDG18, IzmailovMKGVW19, Wenzel2020, aitchison2021a,zeno2021why},
   which is commonly defined as
   $
      p_{\tau}(\mbw \g \xobs) \propto {{p(\xobs \g \mbw)}}^{1/\tau} {p(\mbw)}
   $,
   where $\tau > 0$ is a \emph{temperature} value.
   With $\tau<1$, tempering is known to improve performance in the case of small training data and using miss-specified priors, but it corresponds to artificially sharpening the posterior by over-counting the data $\tau$ times.
\end{minipage}\hfill%
\begin{minipage}{.35\textwidth}
   \centering
   \definecolor{color0}{HTML}{ffab0f}
    \definecolor{color1}{HTML}{3d7afd}
   \tiny
   \includegraphics{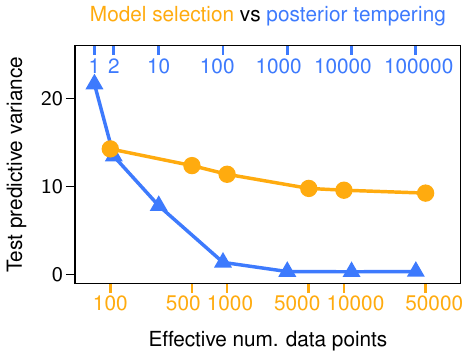}
   \captionof{figure}{
      Average test predictive variance as a function of the \textcolor{color0}{number of data points used to optimize the prior}, and the temperature (i.e. \textcolor{color1}{how many times the data points are over-counted}).
   }
   \label{fig:pred_variance_opt_vs_ts}
\end{minipage}

\begin{minipage}[t]{.375\textwidth}
   \vspace{0pt}
   \centering
   \tiny
   \setlength{\figurewidth}{5.5cm}
   \setlength{\figureheight}{3cm}
   \includegraphics{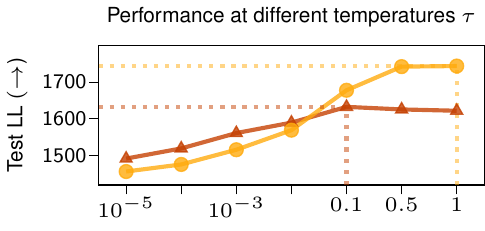}\\
   
\tikzexternaldisable
\fontfamily{phv}\selectfont

\definecolor{0}{rgb}{0.12109375,0.46484375,0.703125}
\definecolor{1}{rgb}{0.8359375,0.15234375,0.15625}
\definecolor{2}{rgb}{0.171875,0.625,0.171875} 
\definecolor{0}{HTML}{c54102}
\definecolor{1}{HTML}{ffab0f}
\tikzexternaldisable\begin{tabular}{ll}
\toprule    
{\protect\tikz[baseline=-1ex]\protect\draw[thick, color=0, fill=0, mark=triangle*, opacity=0.6, mark size=1.8pt, line width=0.8pt] plot[] (-.0, 0)--(.25,0)--(-.25,0);}  \textsf{\gls{BAE} + $\mathcal{N}(0, 1)$ Prior} \\  {\protect\tikz[baseline=-1ex]\protect\draw[thick, color=1, fill=1, mark=*, opacity=0.6, mark size=1.8pt, line width=0.8pt] plot[] (-.0, 0)--(.25,0)--(-.25,0);}  \textsf{\gls{BAE} + Optim. Prior}   \\
\bottomrule
\end{tabular}\tikzexternalenable

   \captionof{figure}{
      Test performance for temperature scaling with different priors. The dotted lines indicate the best performance.
   }
   \label{fig:mnist_temperature_scaling_main}
\end{minipage}%
\hfill%
\begin{minipage}[t]{.6\textwidth}
   \vspace{0pt}
   To demonstrate the differences with our proposal, we setup a comparison on \mnist. %
   In the empirical comparison of \cref{fig:pred_variance_opt_vs_ts}, we consider different temperatures and different sets of data points used to optimize the prior.
   As expected, the tempered posterior quickly collapses on the mode, while the posterior after our treatment retains a sufficiently constant variance, regardless  of the number of data points used.
   It is also interesting to notice that with the $\cN(0, 1)$ prior, the best temperature is $\tau=0.1$, while for our approach that optimizes the prior  is $\tau=1$, further confirming that the model now is well specified (\cref{fig:mnist_temperature_scaling_main}). 
\end{minipage}%

\section{Conclusions}

In this work, we have reconsidered the Bayesian treatment of autoencoders (\gls{AE}) in light of recent advances in Bayesian neural networks.
We have addressed the main challenge of \glspl{BAE}, so that they can be rendered as viable alternative to generative models such as \glspl{VAE}.
More specifically, we have found that the main limitation of \glspl{BAE} lies in the difficulty of specifying meaningful priors in the context of highly-structured data, which is ubiquitous in modern machine learning applications.
Consequently, we have proposed to specify  priors over the autoencoder weights by means of a novel optimization of prior hyper-parameters. %
Inspired by connections with marginal likelihood optimization, 
we derived a practical and efficient optimization framework, based on the minimization of the distributional sliced-Wasserstein distance between the distribution induced by the \gls{BAE} and the data generating distribution. 
The resulting hyper-parameter optimization strategy leads to a novel way to perform model selection for \glspl{BAE}, and we showed its advantages in an extensive experimental campaign.

\paragraph{Limitations and ethical concerns.}
Even if theoretically justified and empirically verified with extensive experimentation, our proposal for {model selection} still remains a \textit{proxy} to the true marginal likelihood maximization. 
The \gls{DSWD} formulation has nice properties of asymptotic convergence and computational tractability, but it may represent only one of the possible solutions. 
At the same time, we stress that the current literature does not cover this problem of \glspl{BAE} at all, and we believe our approach is a considerable step towards the development of practical Bayesian methods for representation learning in modern applications characterized by large-scale structured data (including tabular and graph data, which are currently not covered).
At the same time, the accessibility to these models to a wider audience and different kind of data might help to widespread harmful applications, which is a concern shared among all generative modeling approaches.
An ethical analysis of the consequences of Bayesian priors in unsupervised learning scenarios is also worth an in-depth investigation, which goes beyond the scope of this work.

\begin{ack}
      MF gratefully acknowledges support from the AXA Research Fund and the Agence Nationale de la Recherche (grant ANR-18-CE46-0002 and ANR-19-P3IA-0002).
\end{ack}

{
\small
\bibliographystyle{abbrv}

}

\newpage
\appendix
\section{Derivation of Distributional Sliced-Wasserstein Distance}
In this section, we review some key results on the Wasserstein distance.
%
Given two probability measures $\pi$, $\rho$, both defined on $\bbR^D$ for simplicity, the $p$-Wasserstein distance between $\pi$ and $\rho$ is given by 
\begin{align}
   W_{p}^p(\pi, \rho) = \inf_{\gamma \in \Gamma(\pi, \rho)} \int \|\mbx-\mby\|^{p} \gamma(\mbx, \mby)\dd\mbx \dd\mby\,,
   \label{eq:append:wassertein_dist}
\end{align}
where $\Gamma (\pi, \rho)$ is the set of all possible distributions $\gamma(\mbx,\mby)$ such that the marginals are $\pi(\mbx)$ and $\rho(\mby)$ \cite{villani2008optimal}.
While usually analytically unavailable, for $D=1$ the distance has the following  closed form
solution,
\begin{align}
   W_{p}^p(\pi, \rho) = \int_{0}^{1} | F^{-1}_{\pi}(z) - F^{-1}_{\rho}(z) |^{p} \dd z\,,
   \label{eq:append:wassertein_dist_1d}
\end{align}
where $F_{\pi}$ and $F_{\rho}$ are the \glspl{CDF} of $\pi$ and $\rho$, respectively.

\subsection{(Distributional) Sliced-Wasserstein Distance}
The main idea underlying the \gls{DSWD} is to project the challenging estimation of distances for high-dimensional distributions into simpler estimation of multiple distances in one dimension,
which all have closed-form solution (\cref{eq:append:wassertein_dist_1d}).
The projection is done using the Radon transform $\cR$, an operator that maps a density function $\varphi$ defined in $\bbR^{D}$ to the set of its integrals over hyperplanes in $\bbR^D$, 
\begin{align}
   \cR \varphi(t, \mbtheta) := \int \varphi(\mbz) \delta (t - \mbz^\top \mbtheta) \dd\mbz\,, \quad \forall t \in \bbR\,,\;\; \forall \mbtheta \in \bbS^{D-1}\,,
   \label{eq:append:radon_transform}
\end{align}
where $\bbS^{D-1}$ is the unit sphere in $\bbR^D$ and $\delta(\cdot)$ is the Dirac delta \cite{helgason2010integral}.
Using the Radon transform, for a given $\mbtheta$ we can project the two densities $\pi$ and $\rho$ into one dimension,
\begin{align}
   W_{p}^p(\pi, \rho) = \int_{\bbS^{D-1}} W_{p}^p \left( \cR \pi(t, \mbtheta), \cR \rho(t, \mbtheta) \right) \dd\mbtheta 
   \approx \frac{1}{K} \sum_{i=1}^{K} W_{p}^p \big( \cR \pi(t, \mbtheta_i), \cR \rho(t, \mbtheta_i) \big)\,,
   \label{eq:append:sliced_wassertein_dist_mc_approx}
\end{align}
where the approximation comes from using Monte-Carlo integration by sampling $\mbtheta_i$ uniformly in $\mathbb{S}^{D-1}$ \cite{bonneel2015sliced}. 
While having significant computational advantages, this approach might require to draw many unimportant projections that are 
computationally exhausting 
and that provide a minimal improvement on the overall distance approximation. 

The \textit{distributional sliced-Wasserstein distance} (DSW) \cite{nguyen2021distributional} solves this issue by finding the optimal probability measure of slices $\sigma(\mbtheta)$ on the unit sphere $\bbS^{D-1}$ and it's defined as follows,
\begin{align}
   \label{eq:append:dual_dsw}
   {DSW}_{p}(\pi, \rho; C) := \sup_{\sigma \in \mathbb{M}_C} \Bigg( \mathbb{E}_{\sigma(\mbtheta)}  W_{p}^p \big( \cR {\pi}(t, \mbtheta), \cR {\rho}(t, \mbtheta) \big)  \Bigg)^{{1}/{p}},
\end{align}
where, for $C > 0$, $\mathbb{M}_C$ is the set of probability measures $\sigma$ such that $\mathbb{E}_{\mbtheta, \mbtheta^\prime\sim \sigma} \big[ \mbtheta^\top\mbtheta^\prime \big] \leq C$ (a constraint that aims to avoid directions to lie in only one small area).
Critically, the definition of \gls{DSWD} in \cref{eq:append:dual_dsw} does not suffer from the curse of dimensionality, indeed
\cite{nguyen2021distributional} showed that the statistical error of this estimation scales down with 
$C_D \cdot n^{-\frac{1}{2}}$ , where $C_D$ is a constant depending on dimension $D$. 
Furthermore, while generally we have that $DSW_p(\pi,\rho)\le W_p(\pi,\rho)$, it can be proved that under mild assumptions on $C$, the two distances are topological equivalent, i.e. converging in distribution on $DSW_p$ implies the convergence on $W_p$ \cite[see Theorem 2 in][]{nguyen2021distributional}.

The direct computation of $DSW_p$ in \cref{eq:append:dual_dsw} is still challenging but it admits an equivalent dual form,
\begin{align}
    \sup_{h \in \cH} \left\{ \Big( \E_{\bar\sigma(\mbtheta)} \big[ W_{p}^p \big( \cR \pi(t, h(\mbtheta)), \cR \rho(t, h(\mbtheta)) \big) \big] \Big)^{{1}/{p}} \hskip-2ex - \lambda_{C} \mathbb{E}_{\mbtheta, \mbtheta^\prime \sim \bar\sigma}\Big[ \big| h(\mbtheta)^{\top} h(\mbtheta^\prime) \big| \Big] \right\} + \lambda_C C \,,
   \label{eq:append:dsw_duality}
\end{align}
where $\bar\sigma$ is a uniform distribution in $\bbS^{D-1}$, $\cH$ is a class of all Borel measurable functions $\bbS^{D-1}\rightarrow\bbS^{D-1}$ and $\lambda_C$ is a regularization hyper-parameter.
The formulation in \cref{eq:append:dsw_duality} is obtained by employing the Lagrangian duality theorem and by reparameterizing $\sigma(\mbtheta)$ as push-forward transformation of a uniform measure in $\bbS^{D-1}$ via $h$.
Now, by parameterizing $h$ using a deep neural network\footnote{We use a single multi layer perceptron (MLP) layer with normalized output as the $h$ function.} with parameters $\mbphi$, defined as $h_{\mbphi}$,
\cref{eq:append:dsw_duality} becomes an optimization problem with respect to the network parameters.
The final step is to approximate the analytically intractable expectations with Monte Carlo integration,
\begin{align}
   &DSW_{p}(
      \pi, \rho) \approx  \nonumber \\ 
      & \max_\mbphi \Bigg\{\left[ \frac{1}{K} \sum_{i=1}^{K} \big[ W_{p}^p \big( \cR {\pi}(t, h_{\mbphi}(\mbtheta_i)), \cR {\rho}(t, h_{\mbphi}(\mbtheta_i)) \big) \big] \right]^{{1}/{p}} 
      \hskip-3ex - \frac{\lambda_{C}}{K^2} \sum_{\substack{i,j = 1}}^K  |h_{\mbphi}(\mbtheta_i)^{\top} h_{\mbphi}(\mbtheta_j)| + \lambda_C C\Bigg\}\,,
      \label{eq:mc_approx_dsw_duality}
\end{align}
where $\mbtheta_i$ are uniform samples from the unit sphere $\mathbb{S}^{D-1}$ and $\forall t \in\bbR$.
Finally, we can use stochastic gradient methods to update $\mbphi$ and then use the resulting optima for the estimation of the original distance.

\section{Numerical Implementation of Sliced-Wasserstein Distance}

\subsection{Wasserstein distance between two empirical 1D distributions}

The Wasserstein distance between two one-dimensional distributions $\pi$ and $\rho$ is defined as in \cref{eq:append:wassertein_dist_1d}.
The integral in this equation can be numerically estimated by using the midpoint Riemann sum:
\begin{align}
    \int_{0}^{1} |F^{-1}_{\pi}(z) - F^{-1}_{\rho}(z)|^{p} dz \approx \frac{1}{M} \sum_{m=1}^{M} |F^{-1}_{\pi}(z_m) - F^{-1}_{\rho}(z_m)|^{p},
\end{align}
where $z_m = \frac{2m-1}{M}$, $M$ is the number of points used to approximate the integral.
If we only have samples from the distributions, $x_{m} \sim \pi$ and $y_{m} \sim \rho$, we can obtain the empirical densities as follows
\begin{align}
    \pi(x) \approx \pi_{M}(x)   & = \frac{1}{M} \sum_{m=1}^{M} \delta(x - x_m), \\
    \rho(y) \approx \rho_{M}(y) & = \frac{1}{M} \sum_{m=1}^{M} \delta(y - y_m),
\end{align}
where $\delta$ is the Dirac delta function.
The corresponding empirical cumulative density functions are
\begin{align}
    F_{\pi}(z) \approx F_{\pi, M}(z)   & = \frac{1}{M} \sum_{m=1}^{M} u(z - x_m), \\
    F_{\rho}(z) \approx F_{\rho, M}(z) & = \frac{1}{M} \sum_{m=1}^{M} u(z - y_m),
\end{align}
where $M$ is the number of samples, $u(\cdot)$ is the step function.

Calculating the Wasserstein distance with the empirical distribution function is computationally attractive.
To do that, we first sort $x_m s$ in an ascending order, such that $x_{i[m]} \leq x_{i[m+1]}$, where $i[m]$ is the index of the sorted $x_m s$.
It is straightforward to show that $F^{-1}_{\pi, M}(z_m) = x_{i[m]}$.
Thus, the Wasserstein distance can be approximated as follows
\begin{align}
    W^{p}_{p}(\pi, \rho) \approx \frac{1}{M} \sum_{m=1}^{M} |x_{i[m]} - y_{j[m]}|^{p}.
    \label{eq:empirical_1d_wasser_dist}
\end{align}

\vspace{-20pt}

\subsection{Slicing empirical distribution}
According to the equation \cref{eq:append:radon_transform}, the marginal densities (i.e. slices) of the distribution $\pi$ can be obtained as follows
\begin{align}
    \cR \pi(t, \mbtheta) = \int \pi(\mbx) \delta (t - \mbx^{\top} \mbtheta) d\mbx , \quad \forall t \in \bbR.
\end{align}
Because, in practice, only samples from the distributions are available we aim to calculate a Radon slice of the empirical distribution of $M$ samples $\pi_{M} = \frac{1}{M} \sum_{m=1}^{M} \delta(\mbx - \mbx_m)$:
\begin{align}
    \mathcal{R}\pi(t, \mbtheta) & \approx \frac{1}{M} \sum_{m=1}^{M} \int \delta (\mbx - \mbx_m) \delta(t - \mbx^{\top}\mbtheta) d\mbx  \\
                                & = \frac{1}{M} \sum_{m=1}^{M} \delta(t - \mbx_{m}^{\top} \mbtheta).
    \label{eq:empirical_radon_transform}
\end{align}

By using the approximation in \cref{eq:empirical_radon_transform} and the empirical implementation of 1D Wasserstein distance (\cref{eq:empirical_1d_wasser_dist}), we are able to compute a proxy to the original distance in \cref{eq:append:dual_dsw}.

\vspace{-5pt}
\section{Pseudocode of Prior Optimization Procedure}

\vspace{-10pt}

\cref{alg:prior_optimization}  desribes the procedure of prior optimization for \glspl{BAE}.

\begin{algorithm}[]
    \caption{Prior Optimization}
    \label{alg:prior_optimization}

    \KwInput{Empirical distribution $\tilde{\pi}(\mbx)$; prior over parameters $p_{\mathbold{\psi}}(\mbw)$;
        number of prior samples $N_S$; mini-batch size $N_B$;
        number of random projections $K$; regularization coefficient $\lambda_{C}$.}
    \KwOutput{The optimized prior's parameters $\mathbold{\psi}$}

    \While{$\mbpsi$ has not converged}{
    Sample $\mbx = \{\mbx_{i}\}_{i=1}^{N_{B}}$ from $\tilde{\pi}(\mbx)$ \tcp{Sample input data}
    Sample $\cW = \{\mbw_{i}\}_{i=1}^{N_S}$ from $p_{\mathbold{\psi}}(\mbw)$ \tcp{Sample parameters from the prior} 
    \ForEach{$\mbw_i \in \cW$}{
        \tcc{Following steps are performed in a batch manner} 
        $\hat{\mbx}_i$ = $(f_{\text{dec}} \circ f_{\text{enc}})(\mbx)$ \tcp{Compute the functional outputs from Autoencoder }
        Sample $\tilde{\mbx}_i$ from $p(\mbx \g \hat{\mbx}_i)$  \tcp{Sample from the likelihood}
    }
    Gather samples $\tilde{\mbx} = \cup \{\tilde{\mbx}_i\}_{i=1}^{N_s}$ \\
    $\cL = DSW_2(\mbx, \tilde{\mbx}; K, \lambda_C)$ \tcp{Compute the $DSW_2$ distance using \cref{eq:mc_approx_dsw_duality}}
    $\mathbold{\psi} \leftarrow \text{Optimizer}(\mathbold{\psi}, \nabla_{\mathbold{\psi}} \cL)$ \tcp{Update prior's parameters}
    }

    \textbf{Return:} {$\mathbold{\psi}$}

\end{algorithm}
\section{Details on Stochastic gradient Hamiltonian Monte Carlo}

\gls{HMC} \cite{Neal2011} is a highly-efficient Markov Chain Monte Carlo (MCMC) method used to generate samples from the posterior $\mbw \sim p(\mbw \g \mbx)$.
\gls{HMC} considers the joint log-likelihood as a pontential energy function $U(\mbw) = -\log p(\mbx \g \mbw) - \log p(\mbw)$, and introduces a set of auxilary momentum variable $\mbr$.
Samples are generated from the joint distribution $p(\mbw, \mbr)$ based on the Hamiltonian dynamics:
\begin{align}
    \begin{cases}
        d\mbw & = \mbM^{-1} \mbr dt,  \\
        d\mbr & = -\nabla U(\mbw) dt,
    \end{cases}
\end{align}
where, $\mbM$ is an arbitrary mass matrix that plays the role of a preconditioner.
In practice, this continuous system is approximated by means of $\varepsilon$-discretized numerical integration, and followed by Metropolis steps to accommodate numerical errors stemming from the integration.

However, HMC is not practical for large datasets due to the cost of computing the gradient $\nabla U(\mbw) = \nabla \log (\mbx \g \mbw)$ on the entire dataset.
To mitigate this issue, \cite{Cheni2014} proposed \gls{SGHMC}, which uses a noisy, unbiased estimate of the gradient $\nabla \tilde{U}(\mbw)$ which is computed from a mini-batch of the data.
The discretized Hamiltonian dynamics are then updated as follows
\begin{align}
    \begin{cases}
        \Delta \mbw & = \varepsilon \mbM^{-1} \mbr ,                                                                                         \\
        \Delta \mbr & = - \varepsilon \nabla \tilde{U}(\mbw) - \varepsilon \mbC \mbM^{-1} \mbr + \cN(0, 2\varepsilon (\mbC - \tilde{\mbB})),
    \end{cases}
    \label{eq:discretized_hamilton_dynamics}
\end{align}
where $\varepsilon$ is an step size, $\mbC$ is an user-defined friction matrix, $\tilde{\mbB}$ is the estimate for the noise of the gradient evaluation.
To choose these hyper-parameters, we use a scale-adapted version of \gls{SGHMC} \citep{Springenberg2016}, where the hyper-parameters are adjusted automatically during a burn-in phase.
After this period, all hyperparamteters stay fixed.

\paragraph{Estimating $\mbM$.} We set the mass matrix $\mbM^{-1} = \mathrm{diag}\left(\hat{V}_{\mbw}^{-1/2} \right)$, where $\hat{V}_{\mbw}^{-1/2}$ is an estimate of the uncentered variance of the gradient, $\hat{V}_{\mbw}^{-1/2} \approx \mathbb{E}[( \nabla \tilde{U}(\mbw))^2] $, which can be estimated by using exponential moving average as follows
\begin{equation}
    \Delta \hat{V}_{\mbw} = -\tau^{-1} \hat{V}_{\mbw} + \tau^{-1} \nabla (\tilde{U}(\mbw))^2,
\end{equation}
where $\tau$ is a parameter vector that specifies the moving average windows. This parameter can be automatically chosen by using an adaptive estimate \citep{Springenberg2016} as follows
\begin{equation}
    \Delta \tau = -g_{\mbw}^2 \hat{V}^{-1}_{\mbw} \tau + 1, \quad \text{and}, \quad \Delta g_{\mbw} = -\tau^{-1} g_{\mbw} + \tau^{-1} \nabla \tilde{U}(\mbw),
\end{equation}
where $g_{\mbw}$ is a smoothed estimate of the gradient $\nabla U(\mbw)$.

\paragraph{Estimating $\tilde{\mbB}$.} The estimate for the noise of the gradient evaluation, $\tilde{\mbB}$ should be ideally the estimate of empirical Fisher information matrix of $U(\mbw)$, which is prohibitively expensive to compute.
Therefore, we use a diagonal approximation, $\tilde{\mbB} = \frac{1}{2} \varepsilon \hat{V}_{\mbw}$, which is already available from the step of estimating $\mbM$.

\paragraph{Choosing $\mbC$.} In practice, one can simply set the friction matrix as $\mbC = C \mbI$, i.e. the same independent noise for each elements of $\mbw$.

\paragraph{The discretized Hamiltonian dynamics.} By substituting $\mbv := \varepsilon \hat{V}_{\mbw}^{-1/2} \mbr$, the dynamics \cref{eq:discretized_hamilton_dynamics} become
\begin{align}
    \begin{cases}
        \Delta \mbw & = \mbv,                                                                                                                                                                           \\
        \Delta \mbv & = -\varepsilon^{2} \hat{V}_{\mbw}^{-1/2} \nabla \tilde{U}(\mbw) - \varepsilon C \hat{V}_{\mbw}^{-1/2}  \mbv + \cN(0, 2 \varepsilon^3 C \hat{V}_{\mbw}^{-1} - \varepsilon^4 \mbI).
    \end{cases}
\end{align}
Following \citep{Springenberg2016}, we choose $C$ such that $\varepsilon C \hat{V}_{\mbw}^{-1/2} = \alpha \mbI$. This is equivalent to using a constant momentum coefficient of $\alpha$. The final discretized dynamics are then
\begin{align}
    \begin{cases}
        \Delta \mbw & = \mbv,                                                                                                                                                     \\
        \Delta \mbv & = -\varepsilon^{2} \hat{V}_{\mbw}^{-1/2} \nabla \tilde{U}(\mbw) - \alpha  \mbv + \cN(0, 2 \varepsilon^2 \alpha \hat{V}_{\mbw}^{-1/2} - \varepsilon^4 \mbI).
    \end{cases}
\end{align}
\section{PCA of the SGD Trajectory}
Inspired by \cite{IzmailovMKGVW19}, we use the subspace spanned by the SGD trajectory to visualize neural network's parameters in a low-dimensional space.
This subspace is cheap to construct and can capture many of the sharp directions of the loss surface \citep{IzmailovMKGVW19, Li0TSG18, MaddoxIGVW19}.
More specifically, we perform SGD starting from a MAP solution with a constant learning rate.
Here, the loss function is the negative log joint likelihood of the \gls{BAE}:
\begin{align}
    \cL(\mbw) = -\frac{N}{M} \sum_{i=1}^{M} \log p(\mbx_i \g \mbw) - \ \log p(\mbw),
\end{align}
where $M$ is the mini-batch size and $N$ is the size of training data.
We store the deviations $\mba_i = \overline{\mbw} - \mbw_i$ for the last $M$ epochs, where $\overline{\mbw}$ is the running average of the first moment, $M$  is determined
by the amount of memory we can use.
Then we perform PCA based on randomized SVD \cite{HalkoMT11} on the matrix $\mbA$ comprised of vectors $\mba_1, ..., \mba_M$ to construct the subspace.
The procedure is summarized in \cref{alg:pca_sgd_trajectory}.

\begin{algorithm}[]
    \caption{Subspace construction with PCA}
    \label{alg:pca_sgd_trajectory}

    \KwInput{Pretrained paremeters $\mbw_{\text{MAP}}$; learning rate $\eta$;
        number of steps $\tau$; momentum update frequency $c$;
        maximum number of columns $M$ in deviation matrix $\mbA$.}
    \KwOutput{Shift vector $\overline{\mbw}$; projection matrix $\mbP$ for subspace.}

    $\overline{\mbw} \leftarrow \mbw_{\text{MAP}}$ \tcp{Initialize mean}

    \For{$i \leftarrow 1, 2, ..., T$}{
        $\mbw_i \leftarrow \mbw_{i-1} - \eta \grad_{\mbw} \cL(\mbw_{i-1})$ \tcp{Perform SGD update}
        \If{$\mathtt{MOD}(i,c) = 0$}{
            $n \leftarrow i / c$ \tcp{Number of models}
            $\overline{\mbw} \leftarrow \frac{n\overline{\mbw} + \mbw_{i}}{n+1}$ \tcp{Update mean}
            \If{$\mathtt{NUM\_COLS}(\mbA)=M$}{
                $\mathtt{REMOVE\_COL(\mbA[:, 1])}$
            }
            $\mathtt{APPEND\_COL(\mbA, \mbw_i - \overline{\mbw})}$ \tcp{Store deviation}
        }
    }
    $\mbU, \mbS, \mbV^{\top} \leftarrow SVD(\mbA)$ \tcp{Perform truncated SVD}
    \textbf{Return:} {$\overline{\mbw}$, $\mbP = \mbS \mbV^{\top}$}
\end{algorithm}
\newpage
\section{Additional Details on Experimental Settings}

\subsection{Experimental environment}
In our experiments, we use 4 workstations, which have the following specifications:
\begin{itemize}
    \item \textbf{GPU}: NVIDIA Tesla P100 PCIe 16 GB.
    \item \textbf{CPU}: Intel(R) Xeon(R) (4 cores) @ 2.30GHz.
    \item \textbf{Memory}: 25.5 GiB (DDR3).
\end{itemize}

\subsection{Preprocessing data}

\begin{itemize}
    \item \mnist \cite{lecun1998gradient}: The dataset is publicly available at \url{http://yann.lecun.com/exdb/mnist}. We keep the original resolution of $1\times28\times28$ of the \mnist dataset.
    \item \freyyale \cite{DaiDGL15}: The \frey and \yale datasets are publicly availaibe at \url{http://cs.nyu.edu/~roweis/data.html} and \url{http://vision.ucsd.edu/extyaleb/CroppedYaleBZip}, respectivey. All the images of \frey and \yale datasets are resized to the $1\times28\times28$ resolution.
    \item \celeba \cite{LiuLWT15}: The dataset is publicly available at \url{http://mmlab.ie.cuhk.edu.hk/projects/CelebA.html}. According to \cite{DinhSB17}, we pre-process \celeba images by first taking a $148\times148$ center crop and then resizing to the $3\times64\times64$ resolution.
\end{itemize}

\subsection{Network architectures}
In our experiments, we use convolutional networks for modeling both encoders and decoders.
For a fair comparison, we employ the same network architecture for all models.
The network's parameters are initialized by using the default scheme in PyTorch \cite{paszke2019pytorch}.

\cref{tab:architecture_cnn} shows details on the network architectures used in our experimental campaign.

\begin{table}[H]
  \scriptsize
  \setlength{\tabcolsep}{5pt}
  \begin{sc}

    \begin{tabular}{c p{0.27\linewidth} p{0.27\linewidth} p{0.27\linewidth}}
      \toprule
               & MNIST                                                                         & Frey-Yale & CelebA \\
      \midrule
      \midrule
      Encoder: &
      $x \in \mathbb{R}^{1{\times}28{\times}28}$ \newline
      $\rightarrow \text{Conv}_{32} \rightarrow \text{Leaky RELU}$\newline
      $\quad \rightarrow \text{Conv}_{64}\rightarrow \text{Leaky RELU}$\newline
      $\quad \rightarrow \text{Conv}_{64}\rightarrow \text{Leaky RELU}$\newline
      $\quad \rightarrow \text{Conv}_{128}\rightarrow \text{Leaky RELU}$\newline
      $\quad \rightarrow \text{Flatten} \rightarrow \text{FC}_{50{\times}M}$
               & $x \in \mathbb{R}^{1{\times}28{\times}28}$ \newline
      $\rightarrow \text{Conv}_{64} \rightarrow \text{Leaky RELU}$\newline
      $\quad \rightarrow \text{Conv}_{128}\rightarrow \text{Leaky RELU}$\newline
      $\quad \rightarrow \text{Conv}_{128}\rightarrow \text{Leaky RELU}$\newline
      $\quad \rightarrow \text{Conv}_{256}\rightarrow \text{Leaky RELU}$\newline
      $\quad \rightarrow \text{Flatten} \rightarrow \text{FC}_{50{\times}M}$
               & $x \in \mathbb{R}^{3{\times}64{\times}64}$ \newline
      $\rightarrow \text{Conv}_{64} \rightarrow \text{Leaky RELU}$\newline
      $\quad \rightarrow \text{Conv}_{128}\rightarrow \text{Leaky RELU}$\newline
      $\quad \rightarrow \text{Conv}_{256}\rightarrow \text{Leaky RELU}$\newline
      $\quad \rightarrow \text{Conv}_{512}\rightarrow \text{Leaky RELU}$\newline
      $\quad \rightarrow \text{Flatten} \rightarrow \text{FC}_{50{\times}M}$                                        \\

      \midrule
      Decoder: &
      $z \in \mathbb{R}^{50} \rightarrow \text{FC}_{7{\times}7{\times}128}$\newline
      $\rightarrow \text{Leaky RELU}$\newline
      $\rightarrow \text{ConvT}_{128}\rightarrow \text{Leaky RELU}$\newline
      $\rightarrow \text{ConvT}_{64}\rightarrow \text{Leaky RELU}$\newline
      $\rightarrow \text{ConvT}_{64}\rightarrow \text{Leaky RELU}$\newline
      $\rightarrow \text{ConvT}_{1}\rightarrow \text{Sigmoid}$
               & $z \in \mathbb{R}^{50} \rightarrow \text{FC}_{7{\times}7{\times}256}$\newline
      $\rightarrow \text{Leaky RELU}$\newline
      $\rightarrow \text{ConvT}_{256}\rightarrow \text{Leaky RELU}$\newline
      $\rightarrow \text{ConvT}_{128}\rightarrow \text{Leaky RELU}$\newline
      $\rightarrow \text{ConvT}_{128}\rightarrow \text{Leaky RELU}$\newline
      $\rightarrow \text{ConvT}_{1}\rightarrow \text{Sigmoid}$
               & $z \in \mathbb{R}^{50} \rightarrow \text{FC}_{8{\times}8{\times}512}$\newline
      $\rightarrow \text{Leaky RELU}$\newline
      $\rightarrow \text{ConvT}_{512}\rightarrow \text{Leaky RELU}$\newline
      $\rightarrow \text{ConvT}_{256}\rightarrow \text{Leaky RELU}$\newline
      $\rightarrow \text{ConvT}_{128}\rightarrow \text{Leaky RELU}$\newline
      $\rightarrow \text{ConvT}_{1}\rightarrow \text{Sigmoid}$                                                                                \\
      \bottomrule
    \end{tabular}
  \end{sc}

  \caption{Convolutional Encoder-Decoder architectures. ${\scriptsize{\textsc{Conv}}}_{n}$ denotes a convolutional layer with $n$ filters, whereas ${\scriptsize{\textsc{FC}}}_{n}$ represents a fully-connected layer with $n$ units.
    All convolutions ${\scriptsize{\textsc{Conv}}}_{n}$ and transposed convolutions ${\scriptsize{\textsc{ConvT}}}_{n}$ have a filter size of
    $4{\times}4$ for \mnist and \freyyale and $5{\times}5$ for \celeba.
    $M=1$ for all models except for the \glspl{VAE} which have $M=2$ as the encoder has to yield both mean and variance for each input.}
  \label{tab:architecture_cnn}
\end{table}

\subsection{Prior optimiziation}
As done in \cite{nguyen2021distributional}, we use a single-layer \gls{MLP}, $h_{\mbphi}$, to represent the Borel measurable function in the dual form of \gls{DSWD} (\cref{eq:mc_approx_dsw_duality}).
At each iteration of \cref{alg:prior_optimization}, to find a local maxima, we optimize $h_{\mbphi}$ for $30$ epochs by using an Adam optimizer \cite{jlb2015adam} with a learning rate of $0.0005$.
We use another Adam optimizer with a learning rate of $0.001$ to update the prior's parameters.
We use a mini-batch size of $N_{B} = 64$ and then generate $N_{s} = 32$ prior samples given each data point.
By default, we use $K=1000$ random projections with a regularization coefficient $\lambda_{C}=100$ to estimate the 2-Wasserstein distance.
The convergences of prior optimization on \mnist, \frey and \celeba datasets are illustrated in \cref{fig:wasserstein_convergence}.

\subsection{SGHMC hyper-parameters}
In \cref{tab:sghmc_hyperparams} we report the hyper-parameters used in the experiments on \mnist, \yale and \celeba datasets.
As seen,  we always use a fixed step size of $0.003$, a momentum coefficient of $0.05$, and a mini-batch size of $64$.
The number of collected samples after thinning is $32$.
The number of burn-in iterations and the thinning interval are increased according to the size of the training set.

\begin{table}[htb]
    \centering
    \scalebox{.9}
    {\setlength{\tabcolsep}{5pt}
        \begin{sc}
            \small
            \rowcolors{4}{}{mylightgray}
            \begin{tabular}{lcccccccccccccc}
                \toprule
                                              & \multicolumn{4}{c}{MNIST} & \multicolumn{4}{c}{YALE} & \multicolumn{4}{c}{CELEBA}                                                                                                                                                                                                                                              \\
                \cmidrule(r){2-5} \cmidrule(r){6-9} \cmidrule(r){10-13}
                Training Size                 & \multicolumn{1}{c}{200}   & \multicolumn{1}{c}{500}  & \multicolumn{1}{c}{1000}   & \multicolumn{1}{c}{2000} & \multicolumn{1}{c}{50} & \multicolumn{1}{c}{100} & \multicolumn{1}{c}{200} & \multicolumn{1}{c}{500} & \multicolumn{1}{c}{500} & \multicolumn{1}{c}{1000} & \multicolumn{1}{c}{2000} & \multicolumn{1}{c}{4000} \\
                \midrule
                \midrule
                Mini-batch Size               & 64                        & 64                       & 64                         & 64                       & 64                     & 64                      & 64                      & 64                      & 64                      & 64                       & 64                       & 64                       \\
                Step Size $(10^{-3})$         & $3$                       & $3$                      & $3$                        & $3$                      & $3$                    & $3$                     & $3$                     & $3$                     & $3$                     & $3$                      & $3$                      & $3$                      \\
                Momentum $(10^{-2})$          & $5$                       & $5$                      & $5$                        & $5$                      & $5$                    & $5$                     & $5$                     & $5$                     & $5$                     & $5$                      & $5$                      & $5$                      \\
                Num. Burn-in Steps $(10^{3})$ & $6$                       & $6$                      & $6$                        & $6$                      & $6$                    & $6$                     & $6$                     & $6$                     & $6$                     & $20$                     & $20$                     & $20$                     \\
                Num. Samples                  & $32$                      & $32$                     & $32$                       & $32$                     & $32$                   & $32$                    & $32$                    & $32$                    & $32$                    & $32$                     & $32$                     & $32$                     \\
                Thinning Interval $(10^{3})$  & $1$                       & $1$                      & $1$                        & $2$                      & $1$                    & $1$                     & $1$                     & $1$                     & $1$                     & $2$                      & $3$                      & $5$                      \\
                \bottomrule
            \end{tabular}
        \end{sc}}
    \caption{\gls{SGHMC} hyper-parameters used in the experiments on \mnist, \yale and \celeba datasets.}
    \label{tab:sghmc_hyperparams}
\end{table}

\subsection{Competing approaches}
\begin{itemize}
    \item \textbf{\gls{VAE}} \cite{Kingma14}: The vanilla \gls{VAE} model employed with a Gaussian encoder and a standard Gaussian prior on the latent space.
    \item $\beta$-\textbf{\gls{VAE}} \cite{HigginsMPBGBML17}: The \gls{KL} term in the \gls{VAE}'s objective is weighted by $\beta=0.1$ to reduce the effect of the prior.
          This helps to avoid the over-regularization problem of \glspl{VAE} and improve reconstruction quality.
    \item \textbf{\gls{VAE} + Sylvester Flows} \cite{BergHTW18}: One of the state-of-the-art normalizing flows for the encoder of \glspl{VAE}, which has richer expressiveness than \gls{VAE}'s post-Gaussian encoder.
          As employed in \cite{BergHTW18}, we use Orthogonal Sylvester flows with $4$ transformations and $32$ orthogonal vectors.
    \item \textbf{\gls{VAE} + VampPrior} \cite{TomczakW18}: A flexible prior for \glspl{VAE}, which is a mixture of variational posteriors conditioned on learnable pseudo-observations.
          This allows the variational posterior to learn more a potential latent representation.
          Due to using small training data, we use $100$ trainable pseudo-observations in our experiments.
          We found that increasing more pseudo-observations may hurt the predictive performance because of overfitting.
    \item \textbf{2-Stage \gls{VAE}} \cite{dai2018diagnosing}:  A simple and practical method to improve the quality of generated images from \glspl{VAE} by performing a form of ex-post density estimation via a second \gls{VAE}.
          As employed in \cite{dai2018diagnosing}, for the second-stage \gls{VAE}, we use a \gls{MLP} having three 1024-dimensional hidden layers with ReLU activation function.
    \item \textbf{\wae} \cite{tolstikhin2018wasserstein} Wasserstein Autoencoder: This model is an alternative of \glspl{VAE}.
          By reformulating the objective function as an \gls{OT} problem, \gls{WAE} regularizes the averaged encoding distribution instead of each data point.
          This encourages the encoded training distribution to match the prior while still allowing to learn significant information from the data.
          As suggested in \cite{tolstikhin2018wasserstein}, we use WAE-MMD with the inverse multiquadratics kernel and a regularization coefficient $\lambda=10$ due to its stability compared to WAE-GAN.
          We impose the standard Gaussian prior on the latent space.
    \item \textbf{\textsc{ns}-\textsc{gan}} \cite{Ian2014}: a standard \gls{GAN} with the non-saturating loss,
          which has been shown to be robust to the choice of hyper-parameters on \celeba \cite{LucicKMGB18}.
          For a fair comparison, we reuse the encoder and decoder architectures for the discriminator and generator, respectively.
    \item \textbf{DiffAugment-\gls{GAN}} \cite{ZhaoLLZ020}: a more complex architecture \cite[\textsc{stylegan2}, see][]{KarrasLAHLA20} combined with a powerful differentiable augmentation scheme, specifically developed for low data regimes. 
    We refer to the original work of \cite{ZhaoLLZ020} and the implementation in \url{https://github.com/mit-han-lab/data-efficient-gans} for additional details on the network architecture. We use the same latent size of 50, a maximum of 64 feature maps, and all available augmentations (color, cutout and translation). The remaining parameters are left at default value.
\end{itemize}

All autoencoder models are trained for 200 epochs with an Adam optimizer \cite{jlb2015adam} using the default hyper-parameters in PyTorch, i.e. $\text{learning rate}=0.001$, $\beta_1 = 0.9, \beta_2=0.999$.
The \textsc{ns}-\textsc{gan} is trained for 200 epochs with a learning rate of $0.0002$.
The DiffAugment-\gls{GAN} is trained with $\text{learning rate of }0.001$ for 1 million steps (expect for the case of $4\,000$ training samples, which was extended for 2 millions steps).

\subsection{Performance evaluation}

\paragraph{Test log-likelihood.}
To evaluate the reconstruction quality, we use the mean predictive log-likelihood evaluated over the test set.
This metric tells us how probable it is that the test targets were generated using the test inputs and our model.
Notice that for the case of autoencoder models, the test targets are exactly the test inputs.
The predictive likelihood is a proper scoring rule \cite{gneiting2007strictly} that depends on both the accuracy of predictions and their uncertainty.

For \gls{BAE}, as done in the literature of \glspl{BNN} \cite{Izmailov21, OsawaSKJETY19}, we can estimate the predictive likelihood for an unseen datapoint, $\mbx^{*}$, as follows
\begin{align*}
    \mathbb{E}_{p(\mbw \g \mbx)}[p({\mbx^{*}} \g \mbw)] \approx \frac{1}{M} \sum_{i=1}^{M} p({\mbx^{*}} \g \mbw_{i}), \quad \mbw_{i} \sim p(\mbw \g \mbx),
\end{align*}
where $\mbw_i$ is a sample from the posterior $p(\mbw \g \mbx)$ obtained from the \gls{SGHMC} sampler.

For \glspl{VAE}, because the randomness comes from the latent code not the network's parameters, we can use MC approximation to estimate the predictive likelihood as follows
\begin{align*}
    \mathbb{E}_{q(\mbz \g \mbx^{*})}[p({\mbx^{*}} \g \mbz)] \approx \frac{1}{N} \sum_{i=1}^{N} p({\mbx^{*}} \g \mbz_{i}), \quad \mbz_{i} \sim q(\mbz \g \mbx^{*}),
\end{align*}
where $q(\mbz \g \mbx^{*})$ is the amortized approximate posterior. In our experiments, we use $N = 200$.

For completeness, we also report the test marginal log-likelihood $p(\mbx)$ of \glspl{VAE}, which is estimated by the importance weighted sampling (IWAE) method \cite{BurdaGS15}.
More specifically,
\begin{align*}
    \text{IWAE} = \log \bigg( \frac{1}{K} \sum_{i=1}^{K} \frac{p(\mbx^{*},\mbz_{i})}{q(\mbz_{i} \g \mbx^{*})} \bigg), \quad \mbz_{i} \sim q(\mbz \g \mbx^{*}).
\end{align*}
It can be shown that IWAE lower bounds $\log p(\mbx^{*})$ and can be arbitrarily close to the target as the number of samples $K$ grows.
We use $K=1000$ in the experiments.
The full results of test marginal log-likelihood are reported in Tables \ref{tab:marginal_ll_mnist}, \ref{tab:marginal_ll_face} and \ref{tab:marginal_ll_celeba}.

\paragraph{FID score.}
To assess the quality of the generated images, we employed the widely used Fr\'echet Inception Distance \cite{HeuselRUNH17}.
This metric is the Fr\'echet distance between two multivariate Gaussians, the generated samples and real data samples are compared through their
distribution statistics:
\begin{align}
    \text{FID} = \| \mu_{\text{real}} - \mu_{\text{gen}} \|^2 + \text{Tr}(\Sigma_{\text{real}} + \Sigma_{\text{gen}} - 2 \sqrt{\Sigma_{\text{real}} \Sigma_{\text{gen}}}).
\end{align}
Two distribution samples are calculated from the $2048$-dimensional activations of pool3 layer of Inception-v3 network \footnote{We use the original TensorFlow implementation of FID score which is available at \url{https://github.com/bioinf-jku/TTUR.}}.
In our experiments, the statistics of generated and real data are computed over $10000$ generated images and test data, respectively.

\newpage
\section{Additional Results of Comparison with Temperature Scaling}

In Bayesian deep learning, \emph{temperature scaling} is a practical technique to improve predictive performance \cite{ZhangSDG18, IzmailovMKGVW19, Wenzel2020}.
There are two main approaches to tempering the posterior, namely (1) \textit{partial tempering} and (2) \textit{full tempering} \cite{aitchison2021a,zeno2021why}.
In this section, we investigate rigorously the posteriors induced by the $\cN(0,1)$ prior and optimized prior under different tempering settings.
We use the same setup of \mnist as in the main paper, with $200$ examples for inference.
For the optimized prior, we use $100$ training samples for learning prior.
For the $\cN(0,1)$ prior, we use the union of $200$ training samples and the data used to optimized prior for training.

\subsection{Partial Tempering}

The \emph{partially tempered} posterior is defined as follows \cite{IzmailovMKGVW19, WilsonI20}
\begin{align*}
    p_{\tau_{\text{partial}}}(\mbw \g \mbx) \propto {\underbrace{p(\mbx \g \mbw)}_{\text{likelihood}}}^{1/\tau} \underbrace{p(\mbw)}_{\text{prior}},
\end{align*}
where $\tau > 0$ is a \emph{temperature} value. This parameter controls how the prior and likelihood interact in the posterior.
When $\tau=1$ the true posterior is recovered, and as $\tau$ becomes large, the tempered posterior approaches the prior.
In the case of small training data and using a misspecified prior such as $\mathcal{N}(0,1)$, we would use a small temperature value (e.g. $\tau < 1$) to \emph{reduce the effect of the prior}.
This corresponds to artificially sharpening the posterior by overcounting the data by a factor of $\tau$.

\begin{figure}[H]
    \centering
    \tiny
    \setlength{\figurewidth}{4.0cm}
    \setlength{\figureheight}{3.7cm}
    \begin{subfigure}[t]{0.48\textwidth}
        \centering
        \includegraphics{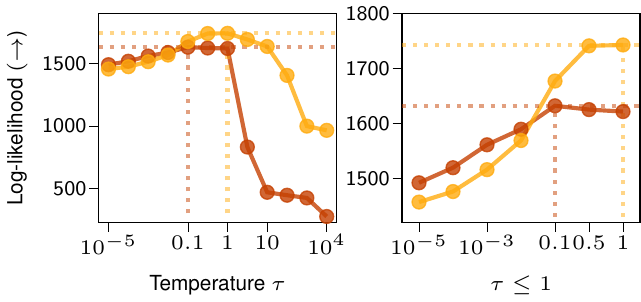}
        \vspace{-1ex}
        \caption{Partial tempering.}
        \label{fig:mnist_parital_tempering_ll}
    \end{subfigure}
    \hspace{1ex}
    \begin{subfigure}[t]{0.48\textwidth}
        \centering
        \includegraphics{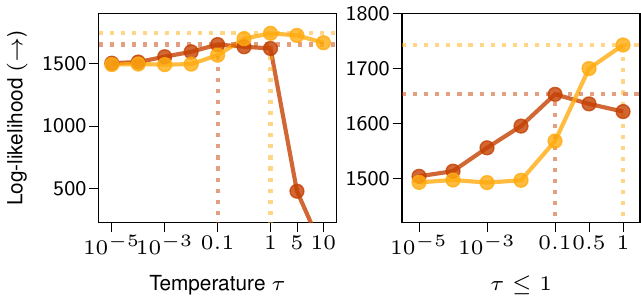}
        \vspace{-1ex}
        \caption{Full tempering.}
        \label{fig:mnist_full_tempering_ll}
    \end{subfigure}

    \vspace{1ex}
    
\tikzexternaldisable
\fontfamily{phv}\selectfont

\definecolor{0}{rgb}{0.12109375,0.46484375,0.703125}
\definecolor{1}{rgb}{0.8359375,0.15234375,0.15625}
\definecolor{2}{rgb}{0.171875,0.625,0.171875} 
\definecolor{0}{HTML}{c54102}
\definecolor{1}{HTML}{ffab0f}
\tikzexternaldisable\begin{tabular}{ll}
\toprule    
{\protect\tikz[baseline=-1ex]\protect\draw[thick, color=0, fill=0, mark=triangle*, opacity=0.6, mark size=1.8pt, line width=0.8pt] plot[] (-.0, 0)--(.25,0)--(-.25,0);}  \textsf{\gls{BAE} + $\mathcal{N}(0, 1)$ Prior} &  {\protect\tikz[baseline=-1ex]\protect\draw[thick, color=1, fill=1, mark=*, opacity=0.6, mark size=1.8pt, line width=0.8pt] plot[] (-.0, 0)--(.25,0)--(-.25,0);}  \textsf{\gls{BAE} + Optim. Prior}   \\
\bottomrule
\end{tabular}\tikzexternalenable

    \caption{Test \gls{LL} as a function of temperature on \mnist using \gls{BAE} with $\mathcal{N}(0,1)$ prior.
        The dotted lines indicate the best performance of \gls{LL}.
    }

    \label{fig:mnist_temperature_scaling}
\end{figure}

\cref{fig:mnist_parital_tempering_ll} shows the test \gls{LL} on \mnist for \gls{BAE} with $\mathcal{N}(0, 1)$ prior and different temperature values.
As expected, the predictive performance of the posterior obtained via low temperatures $\tau < 1$ is much better than those at high temperatures $\tau > 1$.
However, cooling the posterior only shows slight improvement compared to the true posterior induced from the optimized prior.
In addition, in case $\tau > 1$, where the influence of the posterior becomes stronger, the tempered posterior w.r.t. the optimized prior is significantly better than using the $\cN(0,1)$ prior.
This again shows clearly that $\cN(0, 1)$ is a poor prior for a deep \gls{BAE}.

\cref{fig:mnist_subspace_partial_tempering} illustrates samples from priors and posteriors in a low-dimensional space.
We also consider the posterior obtained from the \textit{entire} training data and the $\mathcal{N}(0,1)$ prior as ``oracle'' posterior.
In this case, the choice of the prior does not strongly affect the posterior as this is  dominated by the likelihood.
It can be seen that, for high-temperature values $\tau>1$, the \emph{warm posteriors} w.r.t. $\cN(0,1)$ prior are stretched out as the prior effect is too strong.
These posteriors are mismatched with the ``oracle'' posterior as further confirmed by very low test log-likelihood.
Meanwhile, due to the good inductive bias from the optimized prior, the corresponding tempered posterior is still located in regions nearby the ``oracle'' posterior.
For low temperature values $\tau < 1$, the \emph{cold posteriors} are more concentrated by overcounting evidence.
However, if we use a very small temperature (e.g. $\tau=10^{-5}$), the resulting posterior overly concentrates around the \gls{MLE}, becoming too constrained by the training data.

\newpage
\subsection{Full Tempering}
For the fullly tempered posterior, instead of scaling the likelihood term only, we scale the whole posterior as follows
\begin{align*}
    p_{\tau_{\text{full}}}(\mbw \g \mbx) \propto \big( {\underbrace{p(\mbx \g \mbw)}_{\text{likelihood}}} \underbrace{p(\mbw)}_{\text{prior}} \big)^{1/\tau}.
\end{align*}
The only difference between partial and full tempering is whether we scale the prior.
If we place Gaussian priors on the parameters, this scaling can be absorbed into the prior variance, $\sigma^{2}_{\text{full}} = \sigma^{2}_{\text{partial}} / \tau$.

\begin{figure}[]
    \centering
    \begin{subfigure}[b]{1\textwidth}
        \centering
        \includegraphics[width=0.32\textwidth]{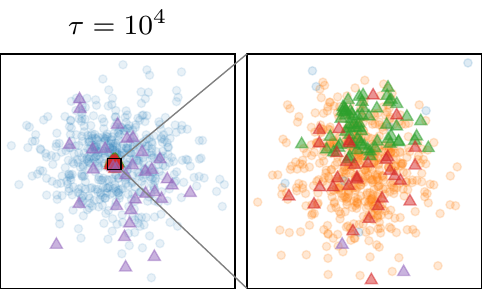}
        \hspace{0.1ex}
        \includegraphics[width=0.32\textwidth]{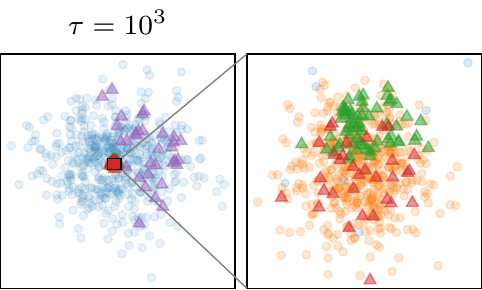}
        \hspace{0.1ex}
        \includegraphics[width=0.32\textwidth]{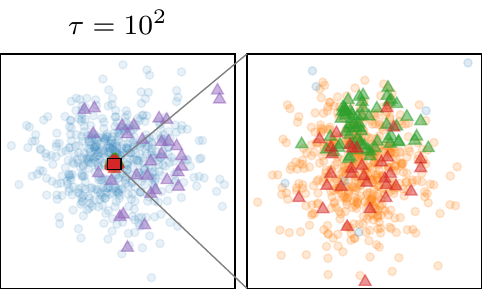} \\
        \vspace{0.1ex}

        \includegraphics[width=0.32\textwidth]{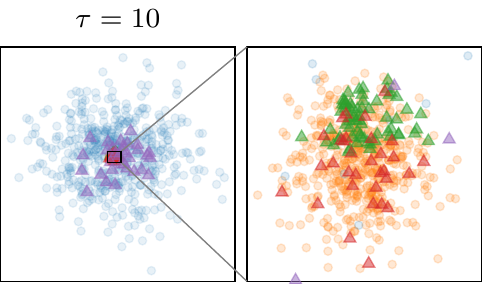}
        \hspace{0.1ex}
        \includegraphics[width=0.32\textwidth]{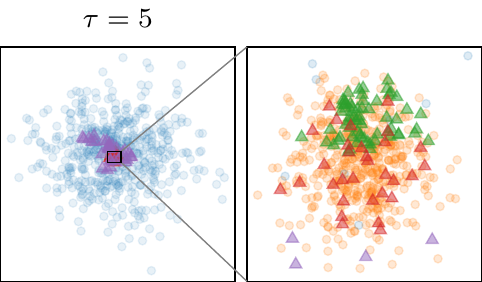}
        \hspace{0.1ex}
        \includegraphics[width=0.32\textwidth]{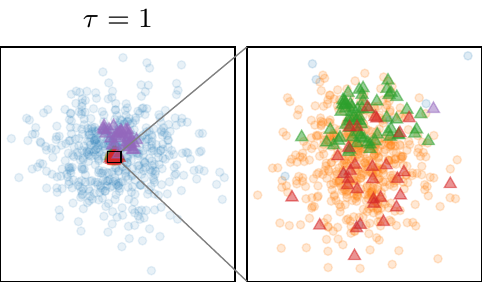} \\
        \vspace{0.1ex}

        \includegraphics[width=0.32\textwidth]{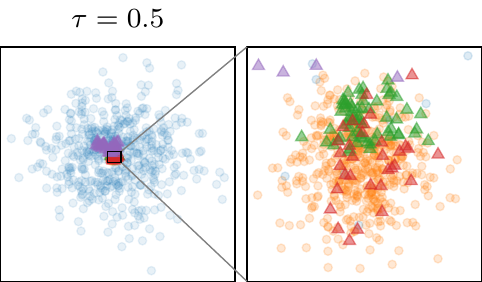}
        \hspace{0.1ex}
        \includegraphics[width=0.32\textwidth]{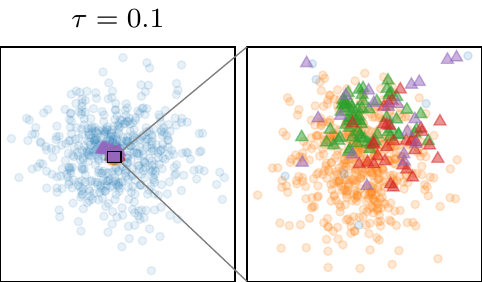}
        \hspace{0.1ex}
        \includegraphics[width=0.32\textwidth]{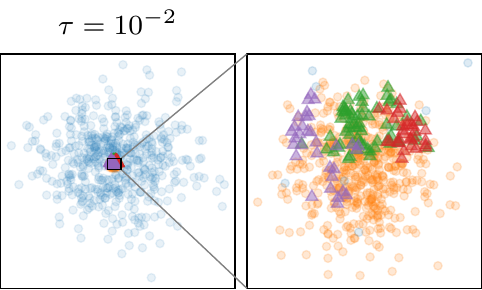} \\
        \vspace{0.1ex}

        \includegraphics[width=0.32\textwidth]{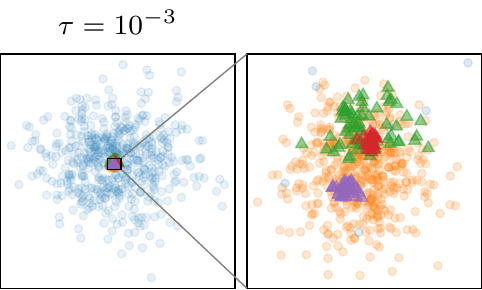}
        \hspace{0.1ex}
        \includegraphics[width=0.32\textwidth]{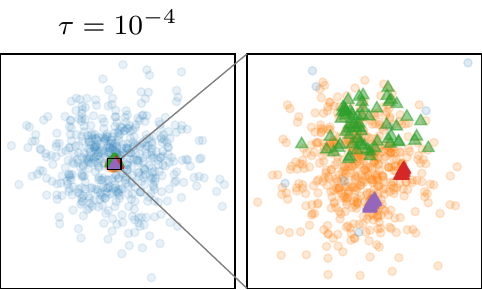}
        \hspace{0.1ex}
        \includegraphics[width=0.32\textwidth]{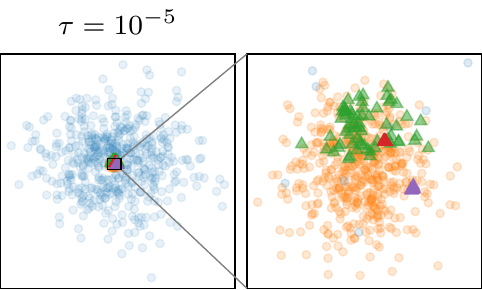} \\

        \caption{Partial tempering.}
        \label{fig:mnist_subspace_partial_tempering}
    \end{subfigure}

    \vspace{2ex}

    \begin{subfigure}[b]{1\textwidth}
        \includegraphics[width=0.32\textwidth]{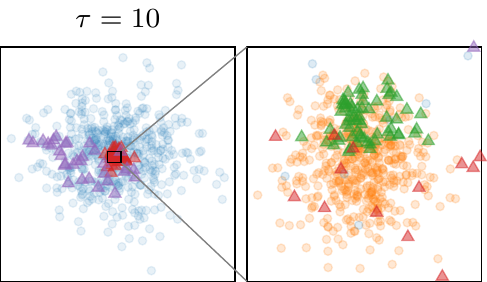}
        \hspace{0.1ex}
        \includegraphics[width=0.32\textwidth]{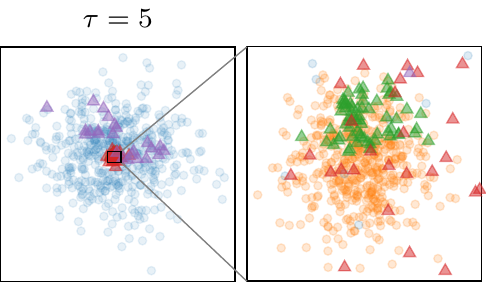}
        \hspace{0.1ex}
        \includegraphics[width=0.32\textwidth]{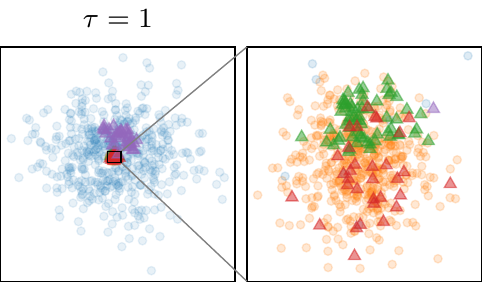} \\
        \vspace{-2ex}

        \includegraphics[width=0.32\textwidth]{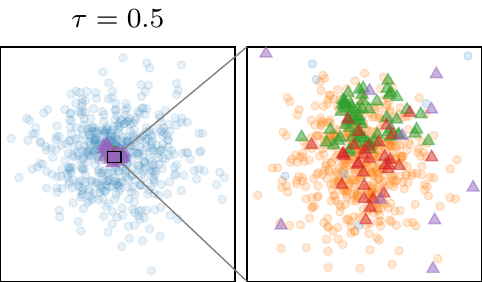}
        \hspace{0.1ex}
        \includegraphics[width=0.32\textwidth]{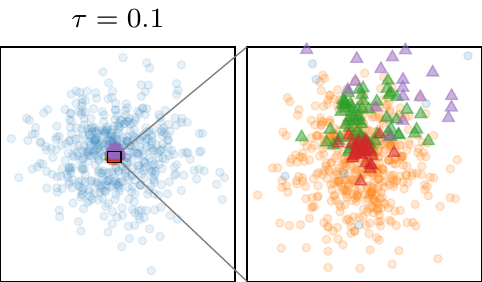}
        \hspace{0.1ex}
        \includegraphics[width=0.32\textwidth]{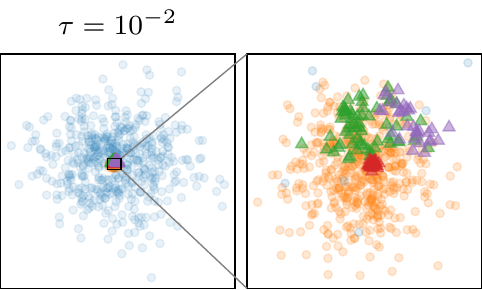} \\
        \vspace{-2ex}

        \includegraphics[width=0.32\textwidth]{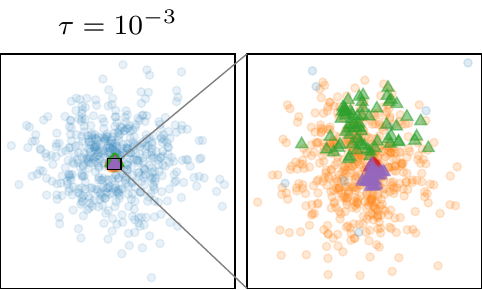}
        \hspace{0.1ex}
        \includegraphics[width=0.32\textwidth]{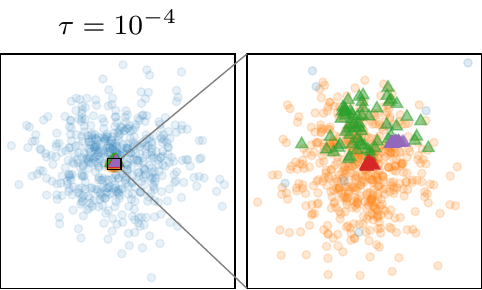}
        \hspace{0.1ex}
        \includegraphics[width=0.32\textwidth]{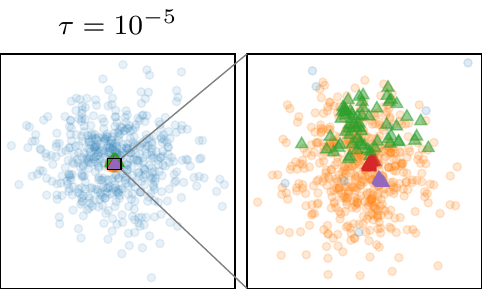} \\
        \vspace{-2ex}
        \caption{Full tempering.}
        \label{fig:mnist_subspace_full_tempering}
    \end{subfigure}

    \vspace{1.5ex}
    {
        \tiny
        \tikzexternaldisable
\fontfamily{phv}\selectfont

\definecolor{color_0}{rgb}{0.99609375,0.49609375,0.0546875}
\definecolor{color_1}{rgb}{0.8359375,0.15234375,0.15625}
\definecolor{color_2}{rgb}{0.12109375,0.46484375,0.703125}
\definecolor{color_3}{rgb}{0.578125,0.40234375,0.73828125}
\definecolor{color_4}{rgb}{0.171875,0.625,0.171875} \tikzexternaldisable\begin{tabular}{lll}
\toprule
{
\protect\tikz[baseline=-1ex]\protect\draw[color=color_0, fill=color_0, mark=*, opacity=0.5, mark size=1.7pt, line width=0.0pt] plot[] (-0.1,0);}  \textsf{Optim. Prior}  &    {\protect\tikz[baseline=-1ex]\protect\draw[color=color_1, fill=color_1, mark=triangle*, opacity=0.5, mark size=1.7pt, line width=0.0pt] plot[] (-0.1,0);}  \textsf{Posterior (Optim. Prior + $\blacklozenge$)} & {}\\
{\protect\tikz[baseline=-1ex]\protect\draw[color=color_2, fill=color_2, mark=*, opacity=0.5, mark size=1.7pt, line width=0.0pt] plot[] (-0.1,0);}  \textsf{$\mathcal{N}(0,1)$ Prior}  &    {\protect\tikz[baseline=-1ex]\protect\draw[color=color_3, fill=color_3, mark=triangle*, opacity=0.5, mark size=1.7pt, line width=0.0pt] plot[] (-0.1,0);}  \textsf{Posterior ($\mathcal{N}(0,1)$ Prior + $\bigstar$)}  & {\protect\tikz[baseline=-1ex]\protect\draw[color=color_4, fill=color_4, mark=triangle*, opacity=0.5, mark size=1.7pt, line width=0.0pt] plot[] (-0.1,0);}  \textsf{Posterior ($\mathcal{N}(0,1)$ Prior + $\blacksquare$)}  \\
\bottomrule
\end{tabular}\tikzexternalenable
    }
    \caption{Visualization of samples from priors and posteriors of \gls{BAE}'s parameters in the plane spanned by eigenvectors of the SGD trajectory.
        {\scriptsize $\blacklozenge$} indicates using $200$ samples for training;   {\scriptsize $\bigstar$} indicates using the union of these samples and $100$ samples used for learning the prior;
    {\scriptsize $\blacksquare$} denotes using all $60000$ training samples.
    Here, $\tau$ is the temperature value used for the {\scriptsize $\blacklozenge$} and {\scriptsize $\bigstar$} cases.
    All plots are produced using convolutional \gls{BAE} on \mnist.}
    \label{fig:mnist_subspace_temperature_scaling}
\end{figure}

Recently, \cite{Wenzel2020} argues that \glspl{BNN} require a cold posterior, where a $\tau < 1$ is employed, to obtain a good performance.
However, we hypothesize that \emph{the cold posterior effect} may originate from using a poor prior.
In this case, as shown in \cref{fig:mnist_full_tempering_ll}, the results of full tempering are similar to those of partial tempering.
Cooling the posterior only helps to increase slightly predictive performance for $\cN(0,1)$ prior.
We also observe that the \gls{MCMC} sampling is not converged if a very large $\tau$ is employed, thus we only consider small values of $\tau$ (e.g. $\tau\in \{5, 10\}$).
In these cases, as depicted in \cref{fig:mnist_subspace_full_tempering}, the samples from the posterior may be outside of the hypothesis space of the optimized prior.

In sum, the true posterior induced from our optimized prior is remarkably better than any types of tempered posteriors.
These results suggest that, in the small-data regime, we should choose carefully a more sensible prior rather than simply using a vague prior and overcounting the data.

\newpage
\section{Ablation Studies}

\subsection{Additional results of ablation study on the size of the dataset to optimize priors}

\begin{figure}[H]
    \centering
    \tiny
    \setlength{\figurewidth}{5.5cm}
    \setlength{\figureheight}{3.7cm}
    \begin{subfigure}[b]{0.38\textwidth}
        \centering
        \includegraphics{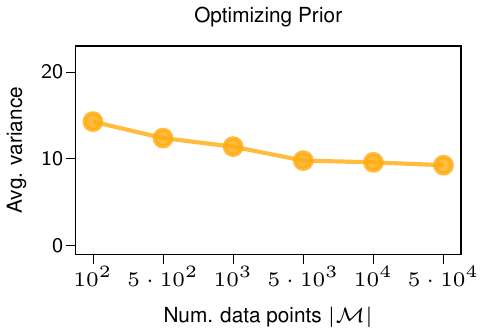}
        \caption{}
        \label{fig:avg_variance_optim_prior}
    \end{subfigure}
    \begin{subfigure}[b]{0.365\textwidth}
        \centering
        \includegraphics{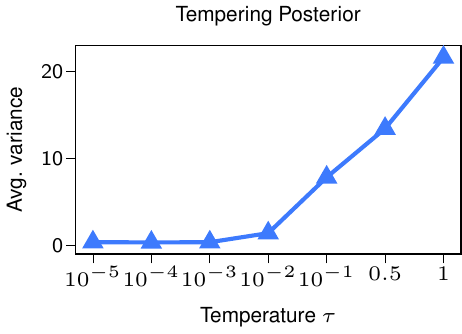}
        \caption{}
        \label{fig:avg_variance_tempered_posterior}
    \end{subfigure}
    \caption{The average predictive variance computed over test datapoints as a function of \textbf{(a)} the number of data points used to optimize prior, and \textbf{(b)} the temperature used for cooling the posterior.
        Here, we use $200$ datapoints from \mnist dataset for inference.
        In figure \textbf{(a)}, we use the optimized prior and consider the true poserior without any tempering.
        In figure \textbf{(b)}, we use the standard Gaussian prior and employ partial tempering for the posterior.}
    \label{fig:avg_variance}
\end{figure}

\begin{figure}[H]
    \centering
    \begin{subfigure}[b]{0.25\textwidth}
        \includegraphics[width=\textwidth]{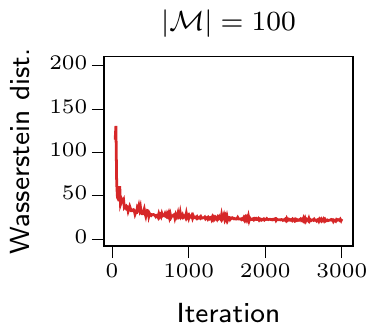}
    \end{subfigure}
    \hspace{-2ex}
    \begin{subfigure}[b]{0.23\textwidth}
        \includegraphics[width=\textwidth]{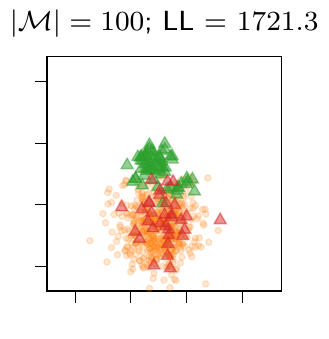}
    \end{subfigure}
    \begin{subfigure}[b]{0.25\textwidth}
        \includegraphics[width=\textwidth]{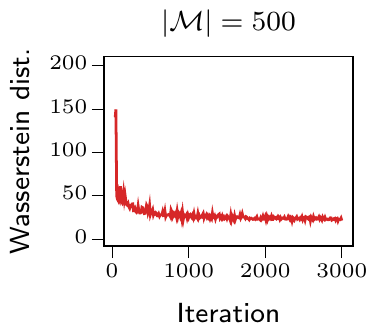}
    \end{subfigure}
    \hspace{-2ex}
    \begin{subfigure}[b]{0.23\textwidth}
        \centering
        \includegraphics[width=\textwidth]{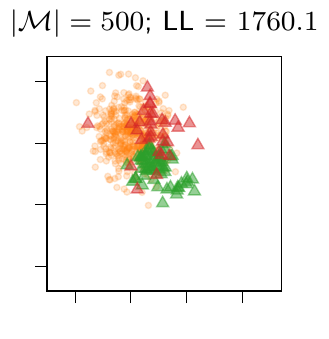}
    \end{subfigure}\\

    \vspace{1ex}

    \begin{subfigure}[b]{0.25\textwidth}
        \includegraphics[width=\textwidth]{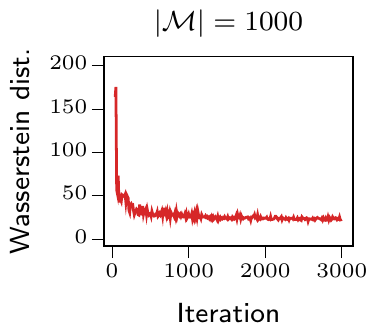}
    \end{subfigure}
    \hspace{-2ex}
    \begin{subfigure}[b]{0.23\textwidth}
        \includegraphics[width=\textwidth]{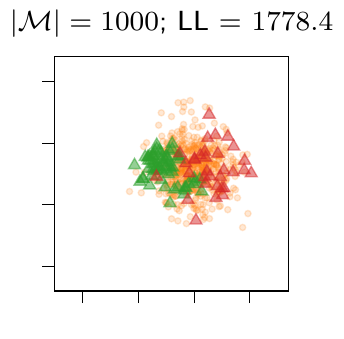}
    \end{subfigure}
    \begin{subfigure}[b]{0.25\textwidth}
        \includegraphics[width=\textwidth]{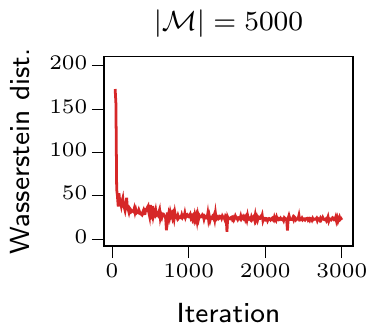}
    \end{subfigure}
    \hspace{-2ex}
    \begin{subfigure}[b]{0.23\textwidth}
        \includegraphics[width=\textwidth]{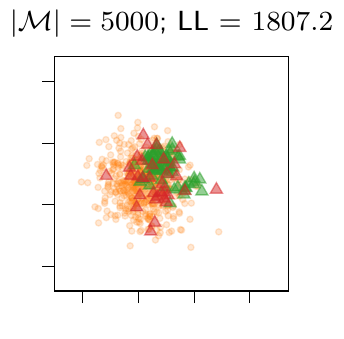}
    \end{subfigure}\\

    \vspace{1ex}

    \begin{subfigure}[b]{0.25\textwidth}
        \includegraphics[width=\textwidth]{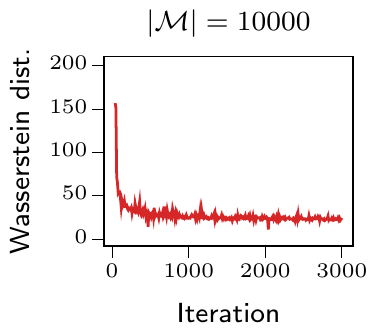}
    \end{subfigure}
    \hspace{-2ex}
    \begin{subfigure}[b]{0.23\textwidth}
        \includegraphics[width=\textwidth]{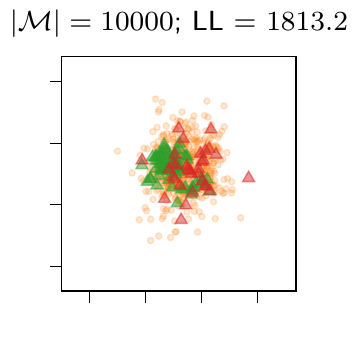}
    \end{subfigure}
    \begin{subfigure}[b]{0.25\textwidth}
        \includegraphics[width=\textwidth]{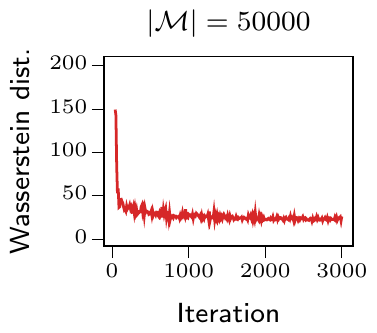}
    \end{subfigure}
    \hspace{-2ex}
    \begin{subfigure}[b]{0.23\textwidth}
        \includegraphics[width=\textwidth]{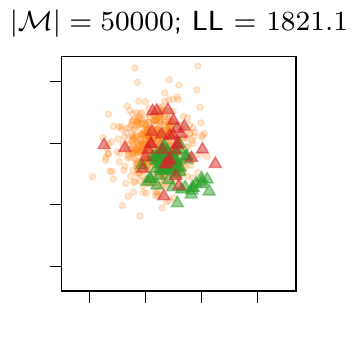}
    \end{subfigure}\\

    {
    \tiny
    \tikzexternaldisable
\fontfamily{phv}\selectfont

\definecolor{color_0}{rgb}{0.99609375,0.49609375,0.0546875}
\definecolor{color_1}{rgb}{0.8359375,0.15234375,0.15625}
\definecolor{color_2}{rgb}{0.12109375,0.46484375,0.703125}
\definecolor{color_3}{rgb}{0.578125,0.40234375,0.73828125}
\definecolor{color_4}{rgb}{0.171875,0.625,0.171875} \tikzexternaldisable\begin{tabular}{lll}
\toprule
{
\protect\tikz[baseline=-1ex]\protect\draw[color=color_0, fill=color_0, mark=*, opacity=0.5, mark size=1.7pt, line width=0.0pt] plot[] (-0.1,0);}  \textsf{Optim. Prior}  &    {\protect\tikz[baseline=-1ex]\protect\draw[color=color_1, fill=color_1, mark=triangle*, opacity=0.5, mark size=1.7pt, line width=0.0pt] plot[] (-0.1,0);}  \textsf{Posterior (Optim. Prior + $\blacklozenge$)} & {\protect\tikz[baseline=-1ex]\protect\draw[color=color_4, fill=color_4, mark=triangle*, opacity=0.5, mark size=1.7pt, line width=0.0pt] plot[] (-0.1,0);}  \textsf{Posterior ($\mathcal{N}(0,1)$ Prior + $\blacksquare$)}  \\
\bottomrule
\end{tabular}\tikzexternalenable

    }

    \caption{Visualization of convergence Wasserstein optimization, and samples from priors and posteriors of \gls{BAE}'s parameters in the plane spanned by eigenvectors of the SGD trajectory corresponding  to the first and second largest
    eigenvalues.
    Here, $|\mathcal{M}|$ is the size of dataset used for optimizing the prior;
    {\scriptsize $\blacklozenge$} indicates using $200$ training samples for inference;
    {\scriptsize $\blacksquare$} denotes using all $60000$ training samples for inference;
    \gls{LL} denotes the test log-likelihood performance of the posterior w.r.t. the optimized prior.
    All plots are produced using convolutional \gls{BAE} on \mnist.}
    \label{fig:mnist_subspace_different_training_size}
\end{figure}

In this experiment, we demonstrate that we can obtain a sensible result by using a small number of training instances to optimize the prior.
Here, we use a set of $200$ samples of $0$-$9$ digits for inference, and another dataset also consisting of $0$-$9$ digits for optimizing the prior.
\cref{fig:mnist_subspace_different_training_size} shows the predictive performance and samples from the posterior.
We observe that the performance gain by using more data is not significant.
We can achieve sensible results by using only about $10$-$50$ samples for each class.
In addition, as illustrated in the low-dimensional space (\cref{fig:mnist_subspace_different_training_size}), the hypothesis space of the prior is not collapsed as we increase the size of the dataset used to optimize the prior.
As a result, the predictive posterior is also not concentrated to the \gls{MLE} solutions as further demonstrated in \cref{fig:avg_variance}.
This behavior is very different from overcounting the data by using temperature scaling, where the posterior becomes more concentrated as the temperature is decreased.
This again demonstrates the practicality of our proposed method in the small-data regime.

\subsection{Effect of the dimensionality of latent space}
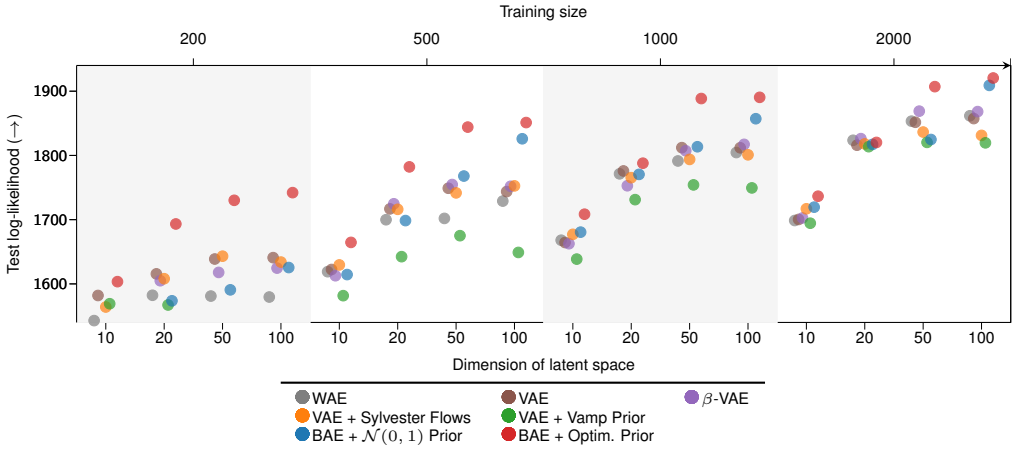
\begin{figure}[H]
    \centering
    \tiny
    \setlength{\figurewidth}{14cm}
    \setlength{\figureheight}{5cm}
    \begin{subfigure}[b]{1\textwidth}
        \centering
        \includegraphics{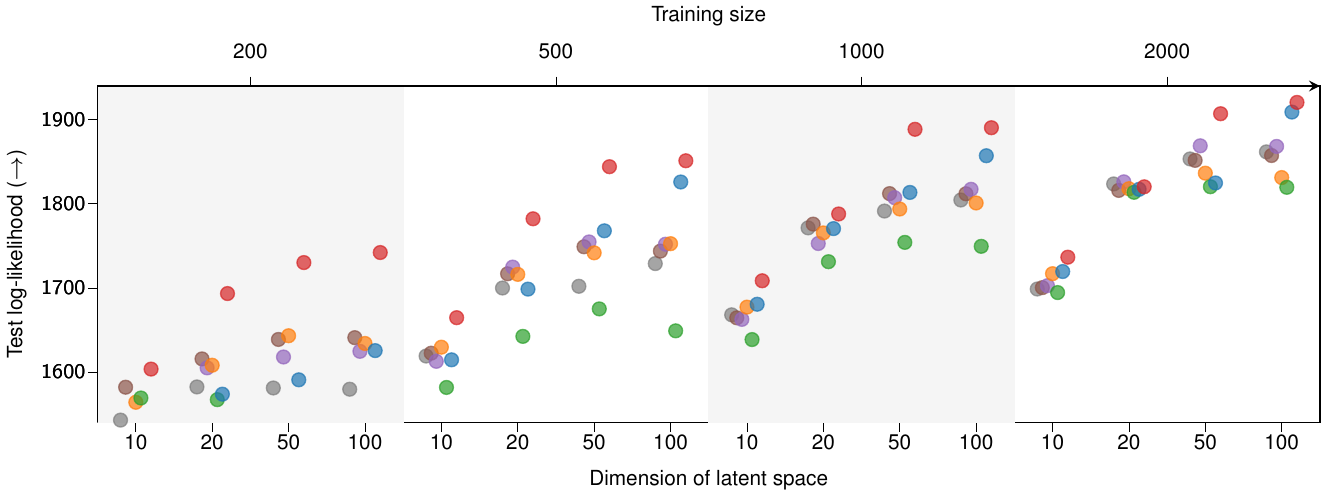}
        \vspace{3 ex}
        \definecolor{wae_combined}{rgb}{0.49609375,0.49609375,0.49609375}
\definecolor{vae_combined}{rgb}{0.546875,0.3359375,0.29296875}
\definecolor{vae_beta_combined}{rgb}{0.578125,0.40234375,0.73828125}
\definecolor{vae_sylvester_combined}{rgb}{0.99609375,0.49609375,0.0546875}
\definecolor{vae_vamp_combined}{rgb}{0.171875,0.625,0.171875}
\definecolor{bae_std_combined}{rgb}{0.12109375,0.46484375,0.703125}
\definecolor{bae_optim}{rgb}{0.8359375,0.15234375,0.15625}
\tikzexternaldisable\begin{tabular}{lll}
    \toprule
    {\protect\tikz[baseline=-1ex]\protect\draw[thick, color=wae_combined, fill=wae_combined, mark=*, opacity=0.6, mark size=2.2pt, line width=0.8pt] plot[] (.25,0);}  \textsf{WAE}                                       & {\protect\tikz[baseline=-1ex]\protect\draw[thick, color=vae_combined, fill=vae_combined, mark=*, opacity=0.6, mark size=2.2pt, line width=0.8pt] plot[] (.25,0);}  \textsf{VAE}                        & {\protect\tikz[baseline=-1ex]\protect\draw[thick, color=vae_beta_combined, fill=vae_beta_combined, mark=*, opacity=0.6, mark size=2.2pt, line width=0.8pt] plot[] (.25,0);}  \textsf{$\beta$-VAE} \\
    {\protect\tikz[baseline=-1ex]\protect\draw[thick, color=vae_sylvester_combined, fill=vae_sylvester_combined, mark=*, opacity=0.6, mark size=2.2pt, line width=0.8pt] plot[] (.25,0);}  \textsf{VAE + Sylvester Flows} & {\protect\tikz[baseline=-1ex]\protect\draw[thick, color=vae_vamp_combined, fill=vae_vamp_combined, mark=*, opacity=0.6, mark size=2.2pt, line width=0.8pt] plot[] (.25,0);}  \textsf{VAE + Vamp Prior}                                                                                                                                                                                                                         \\
    {\protect\tikz[baseline=-1ex]\protect\draw[thick, color=bae_std_combined, fill=bae_std_combined, mark=*, opacity=0.6, mark size=2.2pt, line width=0.8pt] plot[] (.25,0);}  \textsf{BAE + $\mathcal{N}(0,1)$ Prior}    & {\protect\tikz[baseline=-1ex]\protect\draw[thick, color=bae_optim, fill=bae_optim, mark=*, opacity=0.6, mark size=2.2pt, line width=0.8pt] plot[] (.25,0);}  \textsf{BAE + Optim. Prior}                                                                                                                                                                                                                                       \\
    \bottomrule
\end{tabular}\tikzexternalenable
    \end{subfigure}

    \caption{
       Ablation study on the test \gls{LL} on \mnist dataset for different sizes of the latent space and training sizes.
    }

    \label{fig:abl_study_latent_dim}
\end{figure}

\cref{fig:abl_study_latent_dim} illustrates the predictive performance of \glspl{VAE} and \glspl{BAE} in terms test \gls{LL} on \mnist for different size of the latent space and training size.
It is clear that \glspl{BAE} with optimized prior consistently outperforms other competitors across all dimensionalities of the latent space and training sizes.

\subsection{Visualizing 2-dimensional latent space}
\begin{figure}[htb]
    \centering
    \begin{subfigure}[b]{0.25\textwidth}
        \includegraphics[width=\textwidth]{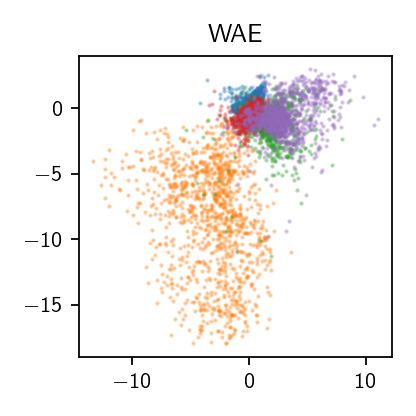}
    \end{subfigure}
    \hspace{-2ex}
    \begin{subfigure}[b]{0.25\textwidth}
        \includegraphics[width=\textwidth]{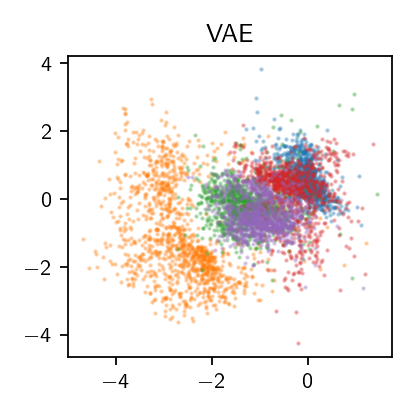}
    \end{subfigure}
    \hspace{-2ex}
    \begin{subfigure}[b]{0.25\textwidth}
        \includegraphics[width=\textwidth]{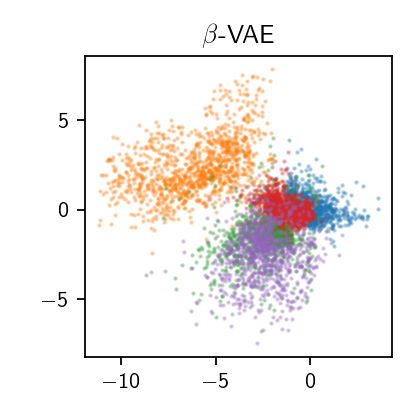}
    \end{subfigure}
    \hspace{-2ex}
    \begin{subfigure}[b]{0.25\textwidth}
        \includegraphics[width=\textwidth]{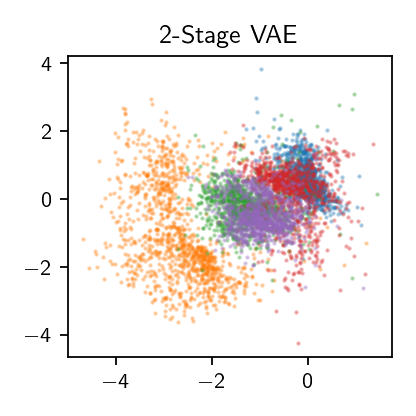}
    \end{subfigure}

    \begin{subfigure}[b]{0.25\textwidth}
        \includegraphics[width=\textwidth]{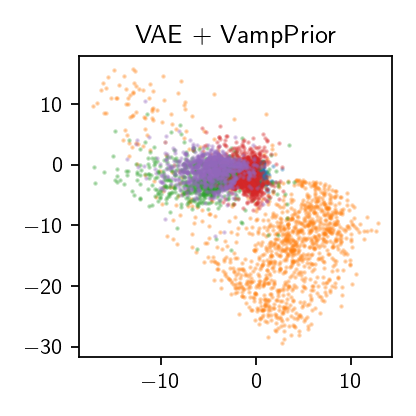}
    \end{subfigure}
    \hspace{-2ex}
    \begin{subfigure}[b]{0.25\textwidth}
        \includegraphics[width=\textwidth]{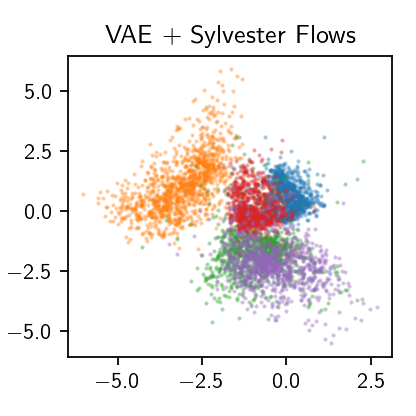}
    \end{subfigure}
    \hspace{-2ex}
    \begin{subfigure}[b]{0.25\textwidth}
        \includegraphics[width=\textwidth]{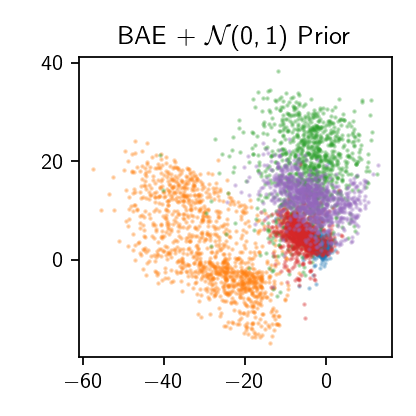}
    \end{subfigure}
    \hspace{-2ex}
    \begin{subfigure}[b]{0.25\textwidth}
        \includegraphics[width=\textwidth]{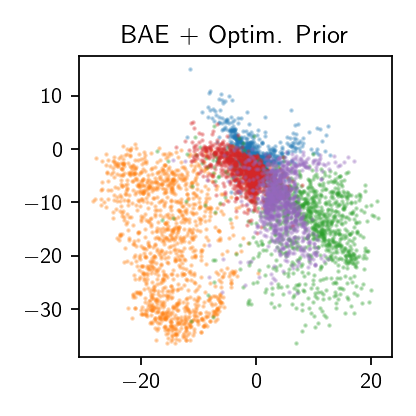}
    \end{subfigure}
    \caption{Visualization of 2D latent spaces of variants of autoencoders on \mnist test set where each color represents a digit classs.
    We consider only 5 classes for easier visualization and comparison.
    All models are trained on $1000$ training samples from \mnist dataset.
    }
    \label{fig:2d_latent_space_5_digits}
\end{figure}

We run several experiments with a low latent space ($K=2$) to test the efficacy of \glspl{VAE} and \glspl{BAE} as dimensionality reduction techniques.
\cref{fig:2d_latent_space_5_digits} shows the results, where each color represents an \mnist digit.
As seen, \gls{BAE} with optimized prior produces a more well-defined class structure in comparision with other methods.

\begin{figure}[htb]
    \centering
    \begin{subfigure}[b]{0.25\textwidth}
        \includegraphics[width=\textwidth]{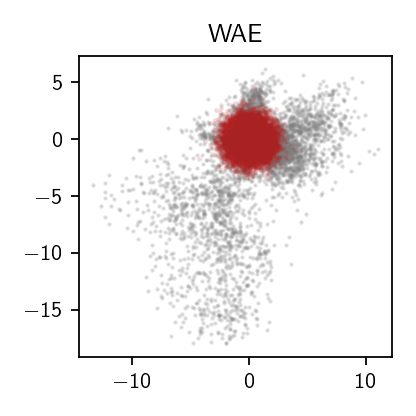}
    \end{subfigure}
    \hspace{-2ex}
    \begin{subfigure}[b]{0.25\textwidth}
        \includegraphics[width=\textwidth]{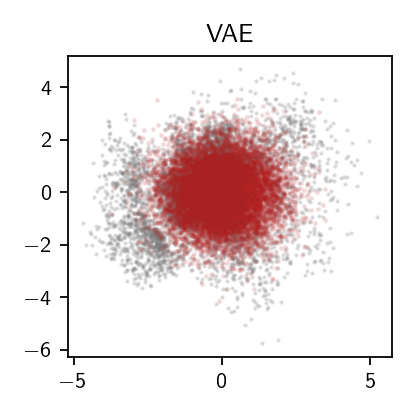}
    \end{subfigure}
    \hspace{-2ex}
    \begin{subfigure}[b]{0.25\textwidth}
        \includegraphics[width=\textwidth]{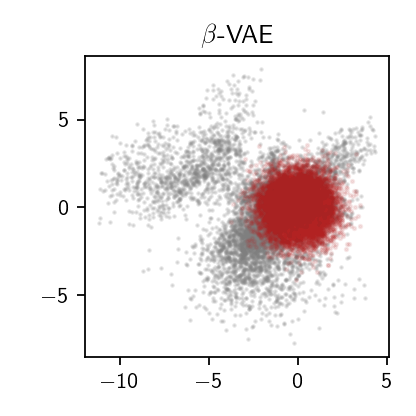}
    \end{subfigure}
    \hspace{-2ex}
    \begin{subfigure}[b]{0.25\textwidth}
        \includegraphics[width=\textwidth]{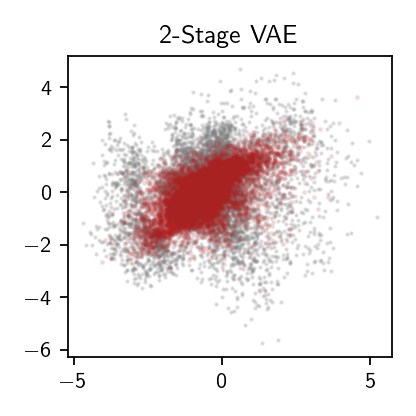}
    \end{subfigure}

    \begin{subfigure}[b]{0.25\textwidth}
        \includegraphics[width=\textwidth]{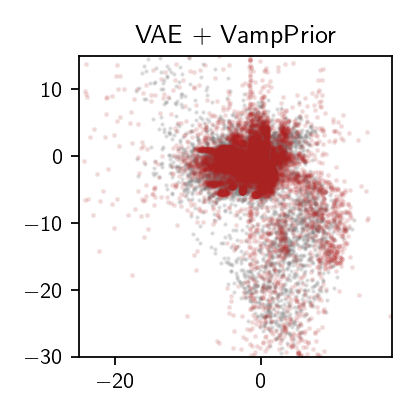}
    \end{subfigure}
    \hspace{-2ex}
    \begin{subfigure}[b]{0.25\textwidth}
        \includegraphics[width=\textwidth]{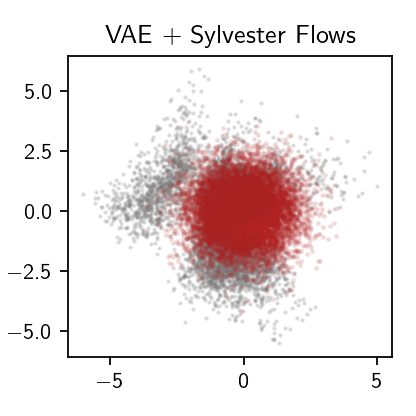}
    \end{subfigure}
    \hspace{-2ex}
    \begin{subfigure}[b]{0.25\textwidth}
        \includegraphics[width=\textwidth]{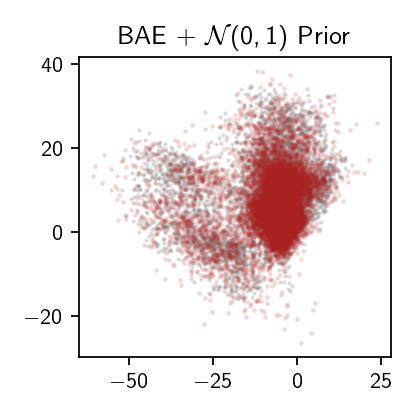}
    \end{subfigure}
    \hspace{-2ex}
    \begin{subfigure}[b]{0.25\textwidth}
        \includegraphics[width=\textwidth]{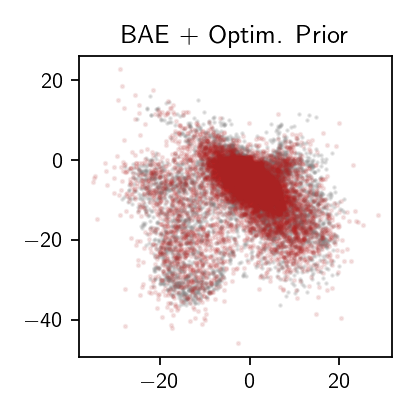}
    \end{subfigure}
    \caption{Diffrent priors and density estimations on the 2-dimensional latent space of \glspl{VAE} and \glspl{BAE}.
        All models are trained on $1000$ training samples from \mnist dataset.
        The gray points are test set samples while the red ones are samples from priors / density estimators.
        Here, we employ the isotropic Gaussian prior on the latent space of \gls{WAE}, \gls{VAE}, $\beta$-\gls{VAE} and \gls{VAE} with Sylveser Flows.
        The VampPrior is learned to explicitly model the aggregated posterior while 2-Stage \gls{VAE} uses another \gls{VAE} to estimate the density of the learned latent space.
        Meanwhile, for \glspl{BAE}, we use \glspl{DPMM} for ex-post density estimation.
    }
    \label{fig:2d_latent_space_5_prior}
\end{figure}

We also consider the $2$D latent space to visualize that ex-post density estimation with \gls{DPMM} helps to reduce the mismatch between the aggregated posterior and the prior.
As can be seen from \cref{fig:2d_latent_space_5_prior}, there are large mismatches between aggregated posterior of \glspl{VAE} and the $\cN(0,1)$ prior.
We can reduce this problem by using a more expressive prior like VampPrior, or performing ex-post density estimation with a second \gls{VAE}.
For \glspl{BAE}, it is clear that the flexible \gls{DPMM} estimator effectively fixes the mismatch and this results in better sample quality as reported in the main paper.

\clearpage
\section{Additional Results}

\subsection{Convergence of Wasserstein optimization}
\begin{figure}[htb]
    \centering
    \tiny
    \setlength{\figurewidth}{5.1cm}
    \setlength{\figureheight}{3.7cm}
    \begin{subfigure}[b]{0.34\textwidth}
       \centering
       \includegraphics{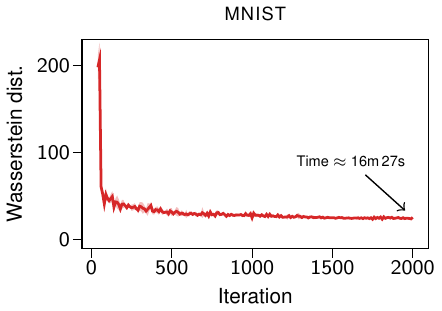}
    \end{subfigure}
    \begin{subfigure}[b]{0.32\textwidth}
       \centering
       \includegraphics{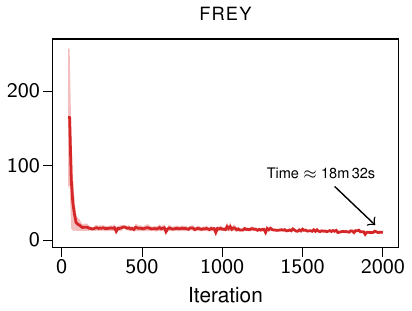}
    \end{subfigure}
    \begin{subfigure}[b]{0.32\textwidth}
     \centering
     \includegraphics{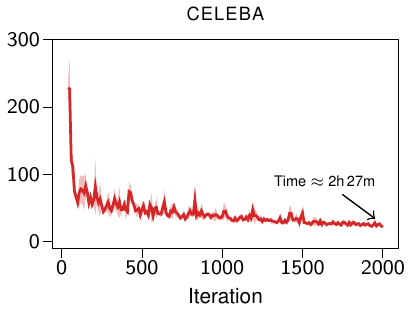}
  \end{subfigure}
    \caption{Convergence of Wasserstein optimization.
    The shaded areas represent the standard deviation computed over 4 random data splits.}
    \label{fig:wasserstein_convergence}
\end{figure}

\cref{fig:wasserstein_convergence} depicts the progressions of Wasserstein optimization in the \mnist, \freyyale and \celeba experiments.

\clearpage
\subsection{Tabulated results}
Detailed results on \mnist, \yale and \celeba datasets are reported from \cref{tab:ll_mnist} to \cref{tab:marginal_ll_celeba}.

\begin{table}[!hp]
    \centering
    \scalebox{.9}
    {\setlength{\tabcolsep}{5pt}
        \begin{sc}
            \small
            \rowcolors{4}{}{mylightgray}
            \begin{tabular}{rllll}
                \toprule
                                                           & \multicolumn{4}{c}{Log Likelihood $(\uparrow)$}                                                                                  \\
                \cmidrule(r){2-5}
                Training Size                                                     & \multicolumn{1}{c}{200} & \multicolumn{1}{c}{500}  & \multicolumn{1}{c}{1000} & \multicolumn{1}{c}{2000} \\
                \midrule
                WAE                                         &  1590.0 {\scriptsize (11.0)} &  1732.7 {\scriptsize (19.2)} &  1809.5 {\scriptsize (11.1)} &    1857.4 {\scriptsize (4.8)} \\
                $\bigstar$ WAE                             &  1675.2 {\scriptsize (10.6)} &  1779.6 {\scriptsize (10.2)} &   1839.3 {\scriptsize (6.1)} &    1871.1 {\scriptsize (3.1)} \\
                \midrule
                VAE                                         &   1635.1 {\scriptsize (8.0)} &   1744.6 {\scriptsize (4.5)} &   1805.5 {\scriptsize (4.8)} &    1847.1 {\scriptsize (3.7)} \\
                $\bigstar$ VAE                              &   1697.0 {\scriptsize (9.9)} &   1776.2 {\scriptsize (6.8)} &   1829.5 {\scriptsize (2.8)} &    1849.9 {\scriptsize (4.4)} \\
                \midrule
                $\beta$-VAE                                 &  1626.2 {\scriptsize (10.3)} &   1749.7 {\scriptsize (9.2)} &   1812.8 {\scriptsize (3.8)} &    1862.3 {\scriptsize (4.9)} \\
                $\bigstar$ $\beta$-VAE                      &   1698.2 {\scriptsize (8.0)} &   1780.2 {\scriptsize (9.3)} &   1841.2 {\scriptsize (3.4)} &    1871.9 {\scriptsize (4.4)} \\
                \midrule
                VAE + Sylveser Flows                        &   1635.4 {\scriptsize (6.1)} &   1743.5 {\scriptsize (1.5)} &   1799.1 {\scriptsize (5.5)} &   1836.3 {\scriptsize (7.2)} \\
                $\bigstar$ VAE + Sylveser Flows             &   1711.4 {\scriptsize (3.0)} &   1781.0 {\scriptsize (2.9)} &   1816.7 {\scriptsize (6.2)} &   1848.1 {\scriptsize (6.5)} \\
                \midrule
                VAE + VampPrior                             &  1543.0 {\scriptsize (12.6)} &  1669.9 {\scriptsize (22.0)} &   1756.8 {\scriptsize (2.6)} &    1818.6 {\scriptsize (3.6)} \\
                $\bigstar$ VAE + VampPrior                  &  1609.6 {\scriptsize (14.4)} &  1732.1 {\scriptsize (14.2)} &   1798.1 {\scriptsize (5.4)} &    1839.3 {\scriptsize (4.0)} \\
                \midrule
                BAE + $\mathcal{N}(0, 1)$ Prior             &  1609.0 {\scriptsize (10.6)} &   1761.0 {\scriptsize (9.1)} &  1837.6 {\scriptsize (18.4)} &   1827.9 {\scriptsize (5.7)} \\
                $\bigstar$ BAE + $\mathcal{N}(0, 1)$ Prior  &  1681.2 {\scriptsize (24.5)} &  1798.6 {\scriptsize (22.8)} &  1827.0 {\scriptsize (35.9)} &   1842.2 {\scriptsize (37.4)} \\
                \midrule
                BAE + Optim. Prior (\textbf{Ours})          &  \textbf{1743.5} {\scriptsize (12.0)} &   \textbf{1845.1} {\scriptsize (1.2)} &   \textbf{1879.1} {\scriptsize (6.3)} &   \textbf{1906.8} {\scriptsize (1.1)} \\
                \bottomrule
            \end{tabular}
        \end{sc}}
    \caption{Evaluation of all methods in terms of test log-likelihood (\emph{the higher, the better}) on \mnist.
        The parentheses are the standard deviations.
        \textcolor{black}{$\bigstar$} indicates that we use the union of the training data and the data used to optimize prior to train the model.}
    \label{tab:ll_mnist}
\end{table}

\begin{table}[!hp]
    \centering
    \scalebox{.9}
    {\setlength{\tabcolsep}{5pt}
        \begin{sc}
            \small
            \rowcolors{4}{}{mylightgray}
            \begin{tabular}{rllll}
                \toprule
                                                           & \multicolumn{4}{c}{Log Likelihood $(\uparrow)$}                                                                                                                \\
                \cmidrule(r){2-5}
                Training Size                              & \multicolumn{1}{c}{50}                          & \multicolumn{1}{c}{100}            & \multicolumn{1}{c}{200}            & \multicolumn{1}{c}{500}            \\
                \midrule
                \midrule
                WAE                                        & 689.7 {\scriptsize (10.4)}                      & 724.8 {\scriptsize (4.4)}          & 754.5 {\scriptsize (3.9)}          & 787.0 {\scriptsize (0.7)}          \\
                $\bigstar$ WAE                             & 718.4 {\scriptsize (0.9)}                       & 740.6 {\scriptsize (4.6)}          & 765.7 {\scriptsize (2.2)}          & 794.3 {\scriptsize (1.6)}          \\
                \midrule
                VAE                                        & 692.3 {\scriptsize (8.4)}                       & 723.5 {\scriptsize (2.8)}          & 738.4 {\scriptsize (3.2)}          & 774.1 {\scriptsize (1.3)}          \\
                $\bigstar$ VAE                             & 701.2 {\scriptsize (5.9)}                       & 728.2 {\scriptsize (3.5)}          & 749.4 {\scriptsize (2.0)}          & 774.8 {\scriptsize (2.1)}          \\
                \midrule
                $\beta$-VAE                                & 707.1 {\scriptsize (5.7)}                       & 733.8 {\scriptsize (8.5)}          & 761.1 {\scriptsize (3.4)}          & 791.8 {\scriptsize (0.7)}          \\
                $\bigstar$ $\beta$-VAE                     & 712.1 {\scriptsize (7.6)}                       & 737.8 {\scriptsize (4.7)}          & 763.4 {\scriptsize (1.3)}          & 790.8 {\scriptsize (1.5)}          \\
                \midrule
                VAE + Sylvester Flows                      & 705.4 {\scriptsize (4.8)}                       & 729.3 {\scriptsize (4.4)}          & 738.2 {\scriptsize (1.6)}          & 766.8 {\scriptsize (0.9)}          \\
                $\bigstar$ VAE + Sylvester Flows           & 682.1 {\scriptsize (11.7)}                      & 716.3 {\scriptsize (4.3)}          & 739.6 {\scriptsize (2.1)}          & 765.3 {\scriptsize (1.2)}          \\
                \midrule
                VAE + VampPrior                            & 690.0 {\scriptsize (6.9)}                       & 722.8 {\scriptsize (1.9)}          & 740.6 {\scriptsize (1.8)}          & 766.8 {\scriptsize (2.7)}          \\
                $\bigstar$ VAE + VampPrior                 & 691.7 {\scriptsize (6.1)}                       & 716.9 {\scriptsize (4.7)}          & 737.8 {\scriptsize (5.3)}          & 764.2 {\scriptsize (2.2)}          \\
                \midrule
                BAE + $\mathcal{N}(0, 1)$ Prior            & 426.1 {\scriptsize (27.6)}                      & 668.8 {\scriptsize (12.8)}         & 724.9 {\scriptsize (21.2)}         & 775.5 {\scriptsize (4.6)}          \\
                $\bigstar$ BAE + $\mathcal{N}(0, 1)$ Prior & 388.0 {\scriptsize (13.6)}                      & 570.4 {\scriptsize (9.1)}          & 688.2 {\scriptsize (5.1)}          & 752.5 {\scriptsize (1.0)}          \\
                \midrule
                BAE + Optim. Prior (\textbf{Ours})         & \textbf{730.3} {\scriptsize (3.0)}              & \textbf{754.3} {\scriptsize (3.1)} & \textbf{771.6} {\scriptsize (3.0)} & \textbf{793.5} {\scriptsize (2.0)} \\
                \bottomrule
            \end{tabular}
        \end{sc}}
    \caption{Evaluation of all methods in terms of test log-likelihood (\emph{the higher, the better}) on \yale.
        The same interpretation as \cref{tab:ll_mnist}.}
    \label{tab:ll_face}
\end{table}

\newpage
\begin{table}[!hp]
    \centering
    \scalebox{.9}
    {\setlength{\tabcolsep}{5pt}
        \begin{sc}
            \small
            \rowcolors{4}{}{mylightgray}
            \begin{tabular}{rllll}
                \toprule
                                                           & \multicolumn{4}{c}{Log Likelihood $(\uparrow)$}                                                                                                                     \\
                \cmidrule(r){2-5}
                Training Size                              & \multicolumn{1}{c}{500}                         & \multicolumn{1}{c}{1000}             & \multicolumn{1}{c}{2000}            & \multicolumn{1}{c}{4000}             \\
                \midrule
                \midrule
                WAE                                        & 5732.6 {\scriptsize (35.3)}                     & 6266.4 {\scriptsize (73.4)}          & 6703.6 {\scriptsize (24.9)}         & 6928.3 {\scriptsize (32.5)}          \\
                $\bigstar$ WAE                             & 6509.7 {\scriptsize (49.2)}                     & 6659.8 {\scriptsize (30.4)}          & 6864.0 {\scriptsize (23.7)}         & 7021.6 {\scriptsize (24.3)}          \\
                \midrule
                VAE                                        & 5914.2 {\scriptsize (78.3)}                     & 6406.4 {\scriptsize (39.6)}          & 6683.6 {\scriptsize (87.5)}         & 6976.4 {\scriptsize (11.9)}          \\
                $\bigstar$ VAE                             & 6460.1 {\scriptsize (33.7)}                     & 6694.1 {\scriptsize (63.1)}          & 6831.8 {\scriptsize (97.2)}         & 7039.5 {\scriptsize (36.5)}          \\
                \midrule
                $\beta$-VAE                                & 5710.2 {\scriptsize (49.0)}                     & 6192.5 {\scriptsize (91.9)}          & 6640.6 {\scriptsize (139.4)}        & 7000.9 {\scriptsize (7.9)}           \\
                $\bigstar$ $\beta$-VAE                     & 6445.3 {\scriptsize (94.0)}                     & 6654.6 {\scriptsize (44.5)}          & 6859.0 {\scriptsize (39.8)}         & 7007.7 {\scriptsize (86.3)}          \\
                \midrule
                VAE + Sylvester Flows                      & 5481.6 {\scriptsize (108.4)}                    & 5984.2 {\scriptsize (37.4)}          & 6415.5 {\scriptsize (33.5)}         & 6699.9 {\scriptsize (46.9)}          \\
                $\bigstar$ VAE + Sylvester Flows           & 6241.3 {\scriptsize (149.2)}                    & 6437.2 {\scriptsize (58.2)}          & 6519.9 {\scriptsize (88.5)}         & 6831.5 {\scriptsize (121.2)}         \\
                \midrule
                VAE + VampPrior                            & 5776.6 {\scriptsize (95.9)}                     & 6242.2 {\scriptsize (92.2)}          & 6691.5 {\scriptsize (24.4)}         & 6999.7 {\scriptsize (15.9)}          \\
                $\bigstar$ VAE + VampPrior                 & 6531.7 {\scriptsize (61.5)}                     & 6591.6 {\scriptsize (97.4)}          & 6868.3 {\scriptsize (27.8)}         & 6990.7 {\scriptsize (37.3)}          \\
                \midrule
                2-Stage VAE                                & 5914.2 {\scriptsize (78.3)}                     & 6406.4 {\scriptsize (39.6)}          & 6683.6 {\scriptsize (87.5)}         & 6976.4 {\scriptsize (11.9)}          \\
                $\bigstar$ 2-Stage VAE                     & 6460.1 {\scriptsize (33.7)}                     & 6694.1 {\scriptsize (63.1)}          & 6831.8 {\scriptsize (97.2)}         & 7039.5 {\scriptsize (36.5)}          \\
                \midrule
                BAE + $\mathcal{N}(0, 1)$ Prior            & 5581.9 {\scriptsize (70.8)}                     & 6273.3 {\scriptsize (54.2)}          & 6848.3 {\scriptsize (15.1)}         & 7154.5 {\scriptsize (15.6)}          \\
                $\bigstar$ BAE + $\mathcal{N}(0, 1)$ Prior & 6574.1 {\scriptsize (46.6)}                     & 6826.5 {\scriptsize (31.0)}          & 7038.3 {\scriptsize (17.8)}         & 7223.1 {\scriptsize (13.2)}          \\
                \midrule
                BAE + Optim. Prior (\textbf{Ours})         & \textbf{6781.3} {\scriptsize (32.4)}            & \textbf{7065.8} {\scriptsize (15.0)} & \textbf{7244.7} {\scriptsize (8.7)} & \textbf{7370.0} {\scriptsize (13.2)} \\
                \bottomrule
            \end{tabular}
        \end{sc}}
    \caption{Evaluation of all methods in terms of test log-likelihood (\emph{the higher, the better}) on \celeba.
        The same interpretation as \cref{tab:ll_mnist}.}
    \label{tab:ll_celeba}
\end{table}

\begin{table}[!hp]
    \centering
    \scalebox{.9}
    {\setlength{\tabcolsep}{5pt}
        \begin{sc}
            \small
            \rowcolors{4}{}{mylightgray}
            \begin{tabular}{rllll}
                \toprule
                                                           & \multicolumn{4}{c}{FID $(\downarrow)$}                                                                                                                   \\
                \cmidrule(r){2-5}
                Training Size                              & \multicolumn{1}{c}{500}                & \multicolumn{1}{c}{1000}            & \multicolumn{1}{c}{2000}            & \multicolumn{1}{c}{4000}            \\
                \midrule
                \midrule
                WAE                                        & 342.14 {\scriptsize (19.02)}           & 309.79 {\scriptsize (12.58)}        & 275.10 {\scriptsize (8.71)}         & 253.06 {\scriptsize (5.52)}         \\
                $\bigstar$ WAE                             & 294.26 {\scriptsize (8.41)}            & 276.24 {\scriptsize (10.49)}        & 261.64 {\scriptsize (6.08)}         & 246.92 {\scriptsize (3.28)}         \\
                \midrule
                VAE                                        & 271.70 {\scriptsize (5.12)}            & 240.69 {\scriptsize (3.44)}         & 230.61 {\scriptsize (7.05)}         & 209.08 {\scriptsize (6.28)}         \\
                $\bigstar$ VAE                             & 248.18 {\scriptsize (12.20)}           & 237.29 {\scriptsize (12.48)}        & 231.50 {\scriptsize (14.17)}        & 206.92 {\scriptsize (9.91)}         \\
                \midrule
                $\beta$-VAE                                & 323.00 {\scriptsize (10.88)}           & 295.54 {\scriptsize (12.45)}        & 276.71 {\scriptsize (15.61)}        & 250.61 {\scriptsize (5.30)}         \\
                $\bigstar$ $\beta$-VAE                     & 285.81 {\scriptsize (5.58)}            & 277.44 {\scriptsize (12.97)}        & 271.82 {\scriptsize (6.69)}         & 262.72 {\scriptsize (17.92)}        \\
                \midrule
                VAE + Sylvester Flows                      & 221.71 {\scriptsize (10.50)}           & 214.94 {\scriptsize (12.01)}        & 207.86 {\scriptsize (9.93)}         & 198.94 {\scriptsize (10.10)}        \\
                $\bigstar$ VAE + Sylvester Flows           & 210.24 {\scriptsize (3.48)}            & 215.00 {\scriptsize (5.79)}         & 204.42 {\scriptsize (11.86)}        & 179.26 {\scriptsize (49.53)}        \\
                \midrule
                VAE + VampPrior                            & 144.41 {\scriptsize (16.61)}           & 131.02 {\scriptsize (2.22)}         & 112.82 {\scriptsize (4.05)}         & 96.20 {\scriptsize (2.79)}          \\
                $\bigstar$ VAE + VampPrior                 & 120.02 {\scriptsize (8.62)}            & 120.23 {\scriptsize (7.16)}         & 102.67 {\scriptsize (7.61)}         & 95.95 {\scriptsize (4.86)}          \\
                \midrule
                2-Stage VAE                                & 78.23 {\scriptsize (2.56)}             & 69.37 {\scriptsize (2.39)}          & 67.69 {\scriptsize (1.55)}          & 74.47 {\scriptsize (4.52)}          \\
                $\bigstar$ 2-Stage VAE                     & 72.21 {\scriptsize (3.05)}             & 69.25 {\scriptsize (3.32)}          & 72.64 {\scriptsize (4.62)}          & 84.95 {\scriptsize (3.91)}          \\
                \midrule
                NS-GAN                                     & 252.33 {\scriptsize (27.03)}           & 171.18 {\scriptsize (15.51)}        & 205.05 {\scriptsize (97.46)}        & 128.29 {\scriptsize (3.81)}         \\
                $\bigstar$ NS-GAN                          & 151.28 {\scriptsize (2.27)}            & 150.74 {\scriptsize (4.39)}         & 137.64 {\scriptsize (4.14)}         & 139.43 {\scriptsize (8.77)}         \\
                \midrule
                $\bigstar$ DiffAugment-GAN                 & \textbf{66.09} {\scriptsize (0.27)}    & \textbf{58.76} {\scriptsize (0.17)} & \textbf{50.22} {\scriptsize (2.62)} & \textbf{45.14} {\scriptsize (0.13)} \\
                \midrule
                BAE + $\mathcal{N}(0, 1)$ Prior            & 89.36 {\scriptsize (4.56)}             & 81.31 {\scriptsize (2.50)}          & 72.50 {\scriptsize (1.37)}          & 71.85 {\scriptsize (0.17)}          \\
                $\bigstar$ BAE + $\mathcal{N}(0, 1)$ Prior & 86.03 {\scriptsize (3.53)}             & 75.86 {\scriptsize (0.45)}          & 71.21 {\scriptsize (1.41)}          & 70.72 {\scriptsize (0.39)}          \\
                \midrule
                BAE + Optim. Prior (\textbf{Ours})         & 68.59 {\scriptsize (3.08)}             & 66.11 {\scriptsize (0.96)}          & 68.34 {\scriptsize (0.86)}          & 67.18 {\scriptsize (0.80)}          \\
                \bottomrule
            \end{tabular}
        \end{sc}}
    \caption{Evaluation of all methods in terms of FID (\emph{the lower, the better}) on \celeba.
        The same interpretation as \cref{tab:ll_mnist}.}
    \label{tab:fid_celeba}
\end{table}

\newpage
\begin{table}[!hp]
    \centering
    \scalebox{.9}
    {\setlength{\tabcolsep}{5pt}
        \begin{sc}
            \small
            \rowcolors{4}{}{mylightgray}
            \begin{tabular}{rllll}
                \toprule
                                                 & \multicolumn{4}{c}{Log Marginal Likelihood $(\uparrow)$}                                                                                         \\
                \cmidrule(r){2-5}
                Training Size                    & \multicolumn{1}{c}{200}                                  & \multicolumn{1}{c}{500}     & \multicolumn{1}{c}{1000}   & \multicolumn{1}{c}{2000}   \\
                \midrule
                \midrule
                VAE                              & 1648.2 {\scriptsize (10.1)}                              & 1744.0 {\scriptsize (5.6)}  & 1795.1 {\scriptsize (2.6)} & 1829.7 {\scriptsize (2.5)} \\
                $\bigstar$ VAE                   & 1702.4 {\scriptsize (8.9)}                               & 1771.0 {\scriptsize (6.7)}  & 1816.2 {\scriptsize (4.6)} & 1832.2 {\scriptsize (4.8)} \\
                \midrule
                $\beta$-VAE                      & 1497.1 {\scriptsize (12.5)}                              & 1625.9 {\scriptsize (7.7)}  & 1687.3 {\scriptsize (3.8)} & 1734.4 {\scriptsize (4.2)} \\
                $\bigstar$ $\beta$-VAE           & 1570.1 {\scriptsize (7.8)}                               & 1655.7 {\scriptsize (7.9)}  & 1715.2 {\scriptsize (2.9)} & 1747.1 {\scriptsize (5.6)} \\
                \midrule
                VAE + Sylvester Flows            & 1627.0 {\scriptsize (6.9)}                               & 1709.8 {\scriptsize (1.8)}  & 1755.4 {\scriptsize (4.6)} & 1783.3 {\scriptsize (5.5)} \\
                $\bigstar$ VAE + Sylvester Flows & 1688.0 {\scriptsize (3.2)}                               & 1741.8 {\scriptsize (2.5)}  & 1771.4 {\scriptsize (3.2)} & 1794.8 {\scriptsize (5.0)} \\
                \midrule
                VAE + VampPrior                  & 1545.6 {\scriptsize (10.5)}                              & 1681.7 {\scriptsize (20.2)} & 1758.4 {\scriptsize (4.1)} & 1810.7 {\scriptsize (2.2)} \\
                $\bigstar$ VAE + VampPrior       & 1616.3 {\scriptsize (15.3)}                              & 1737.5 {\scriptsize (11.4)} & 1795.6 {\scriptsize (4.8)} & 1829.1 {\scriptsize (2.6)} \\
                \bottomrule
            \end{tabular}
        \end{sc}}
    \caption{Evaluation of all methods in terms of test log marginal likelihood of \gls{VAE} models (\emph{the higher, the better}) on \mnist.
        The same interpretation as \cref{tab:ll_mnist}.}
    \label{tab:marginal_ll_mnist}
\end{table}

\begin{table}[!hp]
    \centering
    \scalebox{.9}
    {\setlength{\tabcolsep}{5pt}
        \begin{sc}
            \small
            \rowcolors{4}{}{mylightgray}
            \begin{tabular}{rllll}
                \toprule
                                                 & \multicolumn{4}{c}{Log Marginal Likelihood $(\uparrow)$}                                                                                     \\
                \cmidrule(r){2-5}
                Training Size                    & \multicolumn{1}{c}{50}                                   & \multicolumn{1}{c}{100}   & \multicolumn{1}{c}{200}   & \multicolumn{1}{c}{500}   \\
                \midrule
                \midrule
                VAE                              & 693.8 {\scriptsize (7.8)}                                & 720.8 {\scriptsize (3.4)} & 734.5 {\scriptsize (2.8)} & 767.2 {\scriptsize (0.6)} \\
                $\bigstar$ VAE                   & 704.2 {\scriptsize (5.6)}                                & 723.7 {\scriptsize (3.1)} & 742.4 {\scriptsize (1.9)} & 765.1 {\scriptsize (1.2)} \\
                \midrule
                $\beta$-VAE                      & 628.1 {\scriptsize (2.7)}                                & 655.2 {\scriptsize (9.9)} & 683.0 {\scriptsize (3.9)} & 712.5 {\scriptsize (1.6)} \\
                $\bigstar$ $\beta$-VAE           & 658.5 {\scriptsize (13.4)}                               & 683.9 {\scriptsize (5.4)} & 707.2 {\scriptsize (2.7)} & 731.5 {\scriptsize (2.3)} \\
                \midrule
                VAE + Sylvester Flows            & 668.6 {\scriptsize (5.2)}                                & 686.5 {\scriptsize (3.4)} & 695.1 {\scriptsize (1.5)} & 718.0 {\scriptsize (0.8)} \\
                $\bigstar$ VAE + Sylvester Flows & 655.6 {\scriptsize (4.9)}                                & 677.2 {\scriptsize (3.8)} & 695.7 {\scriptsize (0.7)} & 717.2 {\scriptsize (0.7)} \\
                \midrule
                VAE + VampPrior                  & 672.7 {\scriptsize (7.9)}                                & 697.4 {\scriptsize (6.8)} & 733.4 {\scriptsize (3.2)} & 759.0 {\scriptsize (1.2)} \\
                $\bigstar$ VAE + VampPrior       & 703.9 {\scriptsize (4.2)}                                & 721.5 {\scriptsize (4.0)} & 736.8 {\scriptsize (4.1)} & 760.0 {\scriptsize (2.2)} \\
                \bottomrule
            \end{tabular}
        \end{sc}}
    \caption{Evaluation of all methods in terms of test log marginal likelihood (\emph{the higher, the better}) of \gls{VAE} models on \yale.
        The same interpretation as \cref{tab:ll_mnist}.}
    \label{tab:marginal_ll_face}
\end{table}

\begin{table}[!hp]
    \centering
    \scalebox{.9}
    {\setlength{\tabcolsep}{5pt}
        \begin{sc}
            \small
            \rowcolors{4}{}{mylightgray}
            \begin{tabular}{rllll}
                \toprule
                                                 & \multicolumn{4}{c}{Log Marginal Likelihood $(\uparrow)$}                                                                                             \\
                \cmidrule(r){2-5}
                Training Size                    & \multicolumn{1}{c}{500}                                  & \multicolumn{1}{c}{1000}    & \multicolumn{1}{c}{2000}     & \multicolumn{1}{c}{4000}     \\
                \midrule
                \midrule
                VAE                              & 5973.4 {\scriptsize (66.7)}                              & 6416.6 {\scriptsize (36.6)} & 6673.7 {\scriptsize (82.6)}  & 6943.4 {\scriptsize (8.7)}   \\
                $\bigstar$ VAE                   & 6470.0 {\scriptsize (30.7)}                              & 6676.7 {\scriptsize (57.4)} & 6807.6 {\scriptsize (89.3)}  & 7001.2 {\scriptsize (37.5)}  \\
                \midrule
                $\beta$-VAE                      & 5496.0 {\scriptsize (52.9)}                              & 6007.8 {\scriptsize (89.8)} & 6457.8 {\scriptsize (147.2)} & 6820.1 {\scriptsize (7.4)}   \\
                $\bigstar$ $\beta$-VAE           & 6294.4 {\scriptsize (98.1)}                              & 6472.2 {\scriptsize (46.1)} & 6680.5 {\scriptsize (42.8)}  & 6844.6 {\scriptsize (93.9)}  \\
                \midrule
                VAE + Sylvester Flows            & 5545.8 {\scriptsize (97.8)}                              & 5988.2 {\scriptsize (40.9)} & 6387.9 {\scriptsize (37.8)}  & 6649.9 {\scriptsize (47.1)}  \\
                $\bigstar$ VAE + Sylvester Flows & 6226.9 {\scriptsize (140.3)}                             & 6406.2 {\scriptsize (53.7)} & 6485.3 {\scriptsize (85.4)}  & 6787.9 {\scriptsize (126.9)} \\
                \midrule
                VAE + VampPrior                  & 5842.2 {\scriptsize (82.8)}                              & 6273.8 {\scriptsize (86.3)} & 6682.6 {\scriptsize (16.8)}  & 6984.6 {\scriptsize (7.4)}   \\
                $\bigstar$ VAE + VampPrior       & 6538.3 {\scriptsize (62.4)}                              & 6595.9 {\scriptsize (93.8)} & 6852.3 {\scriptsize (18.6)}  & 6966.1 {\scriptsize (28.0)}  \\
                \bottomrule
            \end{tabular}
        \end{sc}}
    \caption{Evaluation of all methods in terms of test log marginal likelihood (\emph{the higher, the better}) of \gls{VAE} models on \celeba.
        The same interpretation as \cref{tab:ll_mnist}.}
    \label{tab:marginal_ll_celeba}
\end{table}

\newpage
\subsection{More qualitative results}
\begin{figure}[H]

   \centering
   \begin{subfigure}[b]{0.32\textwidth}
      \includegraphics[width=\textwidth]{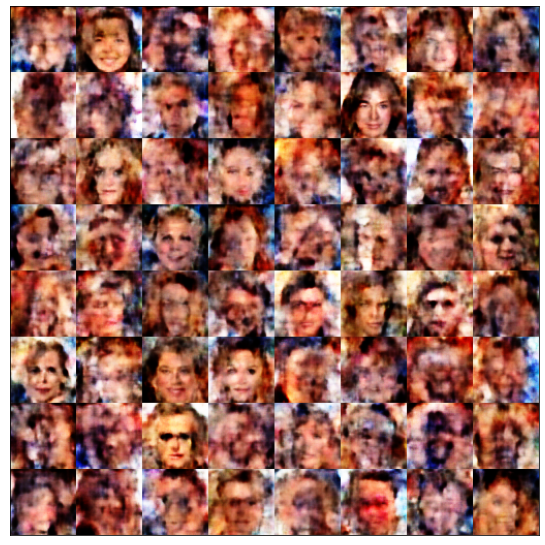}
      \vspace{-2.8ex}
      \caption{WAE}
  \end{subfigure}
   \begin{subfigure}[b]{0.32\textwidth}
       \includegraphics[width=\textwidth]{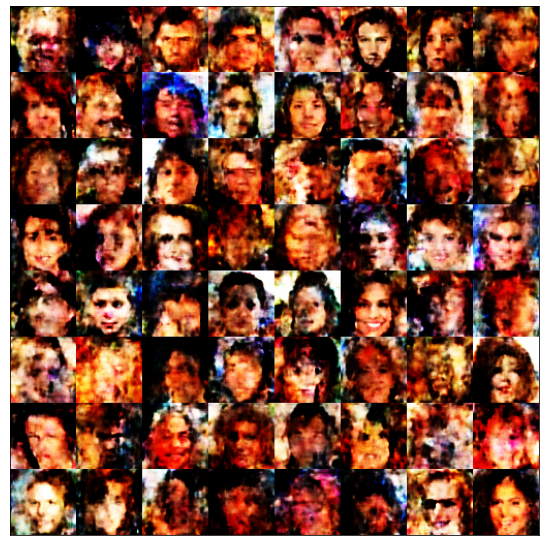}
       \vspace{-2.8ex}
       \caption{VAE}
   \end{subfigure}
   \begin{subfigure}[b]{0.32\textwidth}
       \includegraphics[width=\textwidth]{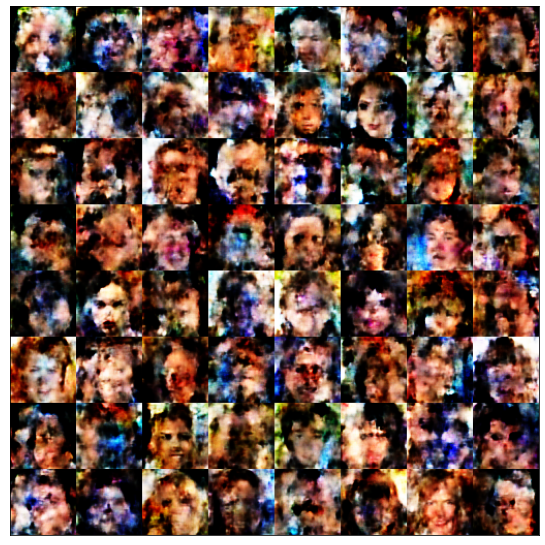}
       \vspace{-2.8ex}
       \caption{$\beta$-VAE}
   \end{subfigure}

   \begin{subfigure}[b]{0.32\textwidth}
      \includegraphics[width=\textwidth]{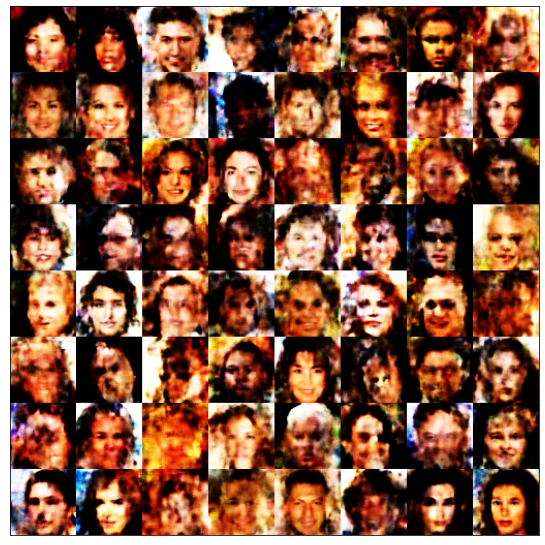}
      \vspace{-2.8ex}
      \caption{VAE + Sylvester Flows}
  \end{subfigure}
   \begin{subfigure}[b]{0.32\textwidth}
       \includegraphics[width=\textwidth]{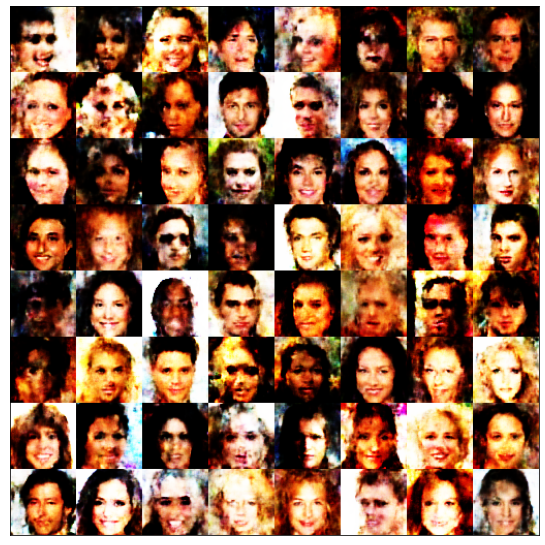}
       \vspace{-2.8ex}
       \caption{VAE + VampPrior}
   \end{subfigure}
   \begin{subfigure}[b]{0.32\textwidth}
       \includegraphics[width=\textwidth]{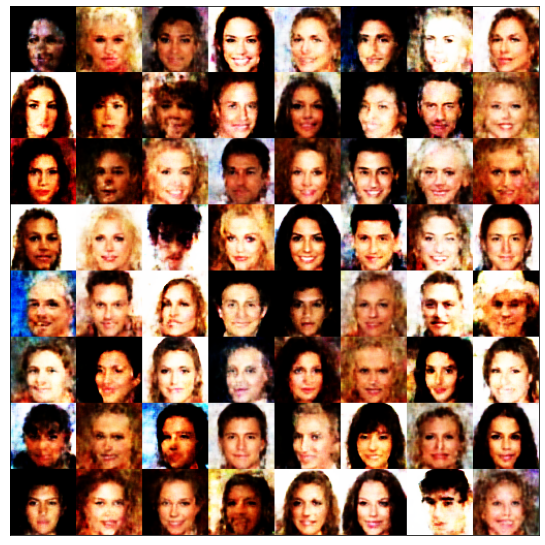}
       \vspace{-2.8ex}
      \caption{2Stage-VAE}
   \end{subfigure}

   \begin{subfigure}[b]{0.32\textwidth}
      \includegraphics[width=\textwidth]{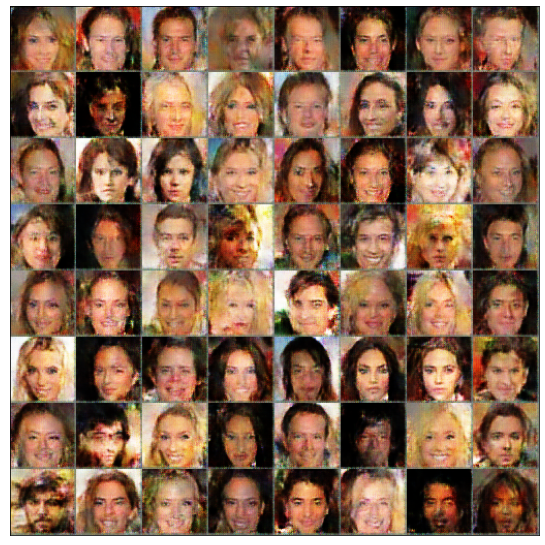}
      \vspace{-2.8ex}
      \caption{NS-GAN}
  \end{subfigure}
   \begin{subfigure}[b]{0.316\textwidth}
      \includegraphics[width=\textwidth]{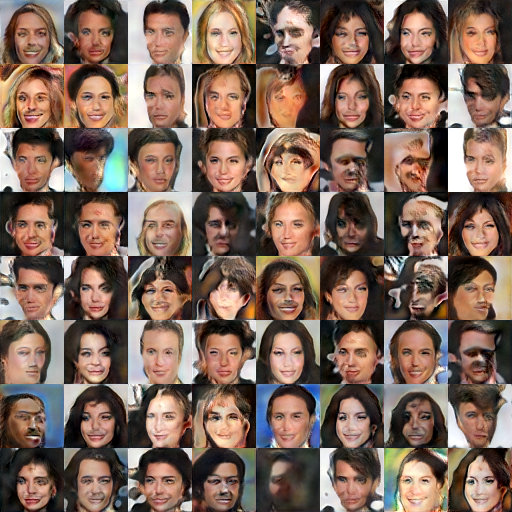}
       \vspace{-2.8ex}
       \caption{StyleGAN2 + DiffAugment}
   \end{subfigure}
   \begin{subfigure}[b]{0.32\textwidth}
      \includegraphics[width=\textwidth]{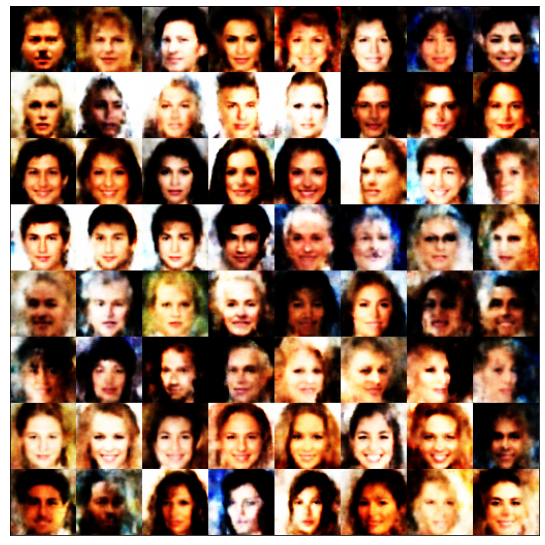}
      \vspace{-2.8ex}
      \caption{BAE + $\mathcal{N}(0,1)$ Prior}
  \end{subfigure}
   
   \begin{subfigure}[b]{0.32\textwidth}
       \includegraphics[width=\textwidth]{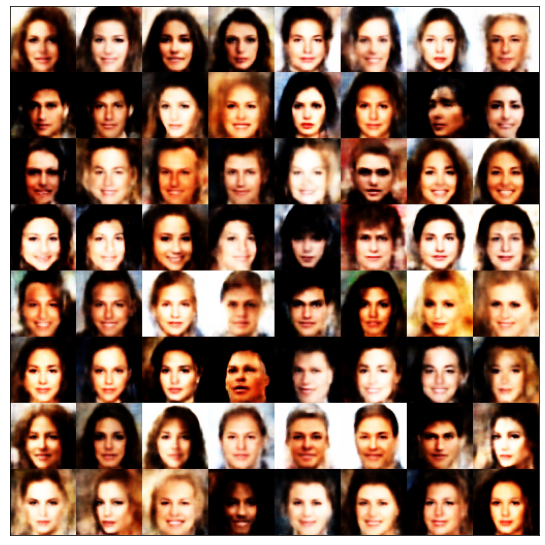}
       \vspace{-2.8ex}
       \caption{BAE + Optim. Prior (\textbf{Ours})}
   \end{subfigure}
   \begin{subfigure}[b]{0.32\textwidth}
      \hfill
  \end{subfigure}
   \begin{subfigure}[b]{0.32\textwidth}
      \hfill
   \end{subfigure}

   \caption{Qualitative evaluation for sample quality for autoencoders and GANs on \celeba.
      Here, we use 500 samples for training/inference.
   }
   \label{fig:celeba_generaed_imgs}
\end{figure}

\begin{table}[!hp]
   \centering
   \scalebox{.9}
   {\setlength{\tabcolsep}{5pt}
      \begin{sc}
         \small
         \begin{tabular}{rc}
            \toprule
             & CelebA - Reconstructions \\
             \midrule
             \raisebox{6pt}{Ground Truth} & \includegraphics[clip,width=0.7\linewidth]{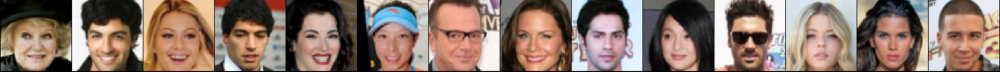} \\[-2.15pt] 
             
             \raisebox{6pt}{$\bigstar$ WAE} & \includegraphics[clip,width=0.7\linewidth]{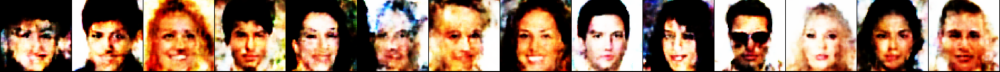} \\[-2.15pt] 

             \raisebox{6pt}{$\bigstar$ VAE} & \includegraphics[clip,width=0.7\linewidth]{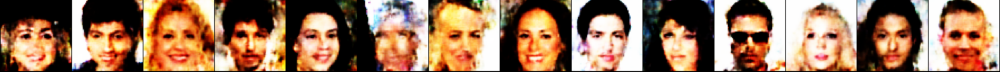} \\[-2.15pt] 
            
             \raisebox{6pt}{$\bigstar$ $\beta$-VAE} & \includegraphics[clip,width=0.7\linewidth]{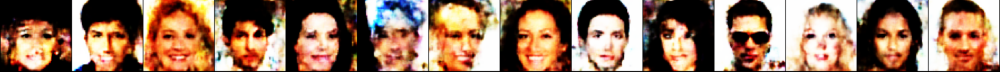} \\[-2.15pt] 

             \raisebox{6pt}{$\bigstar$ VAE + Sylvester Flows} & \includegraphics[clip,width=0.7\linewidth]{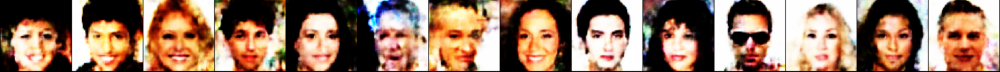} \\[-2.15pt] 

             \raisebox{6pt}{$\bigstar$ VAE + VampPrior} & \includegraphics[clip,width=0.7\linewidth]{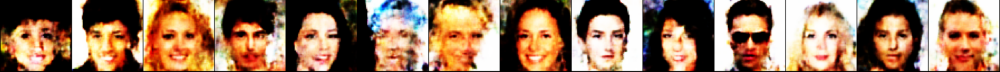} \\[-2.15pt] 

             \raisebox{6pt}{$\bigstar$ 2-Stage VAE} & \includegraphics[clip,width=0.7\linewidth]{figures/celeba_conv/res_vae_combined_appendix.png} \\[-2.15pt] 

             \raisebox{6pt}{$\bigstar$ BAE + $\cN(0, 1)$ Prior} & \includegraphics[clip,width=0.7\linewidth]{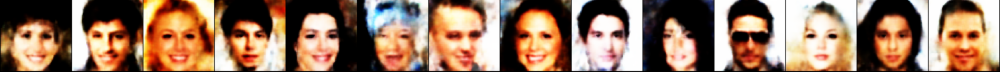} \\[-2.15pt] 

             \raisebox{6pt}{BAE + Optim. Prior (\textbf{Ours})} & \includegraphics[clip,width=0.7\linewidth]{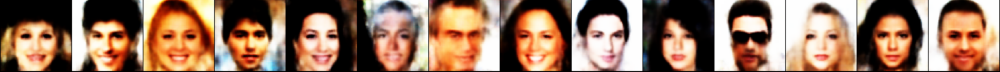} \\[-2.15pt] 

            \bottomrule
         \end{tabular}
      \end{sc}}
   \caption{Qualitative evaluation for reconstructed samples on \celeba.
      \textcolor{black}{$\bigstar$} indicates that we use the union of the training data and the data used to optimize prior to train the model.
      Here, the training size is $1000$.}
   \label{tab:recons_celeba_appendix}
\end{table}

\begin{table}[!hp]
   \centering
   \scalebox{.9}
   {\setlength{\tabcolsep}{5pt}
      \begin{sc}
         \small
         \begin{tabular}{rc}
            \toprule
             & MNIST - Reconstructions \\
             \midrule
             \raisebox{6pt}{Ground Truth} & \includegraphics[clip,width=0.7\linewidth]{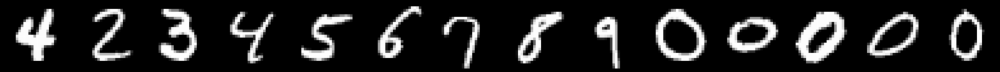} \\[-2.15pt] 
             
             \raisebox{6pt}{WAE} & \includegraphics[clip,width=0.7\linewidth]{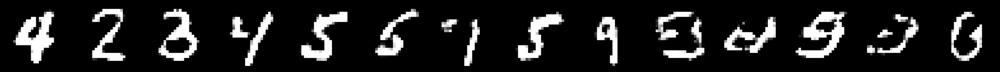} \\[-2.15pt] 
             \raisebox{6pt}{$\bigstar$ WAE} & \includegraphics[clip,width=0.7\linewidth]{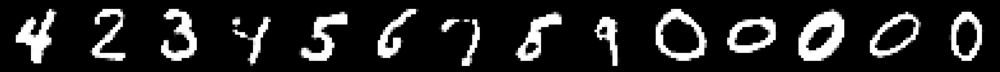} \\[-2.15pt] 

             \raisebox{6pt}{VAE} & \includegraphics[clip,width=0.7\linewidth]{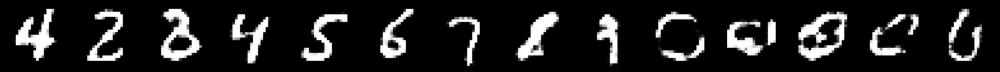} \\[-2.15pt] 
             \raisebox{6pt}{$\bigstar$ VAE} & \includegraphics[clip,width=0.7\linewidth]{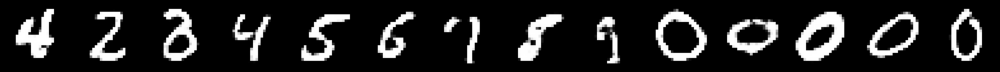} \\[-2.15pt] 
            
             \raisebox{6pt}{$\beta$-VAE} & \includegraphics[clip,width=0.7\linewidth]{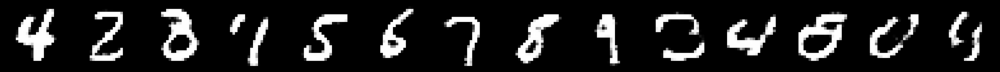} \\[-2.15pt] 
             \raisebox{6pt}{$\bigstar$ $\beta$-VAE} & \includegraphics[clip,width=0.7\linewidth]{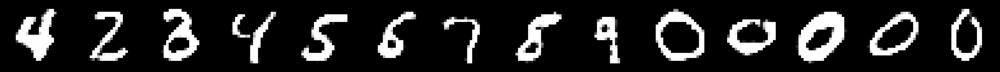} \\[-2.15pt] 

             \raisebox{6pt}{VAE + Sylvester Flows} & \includegraphics[clip,width=0.7\linewidth]{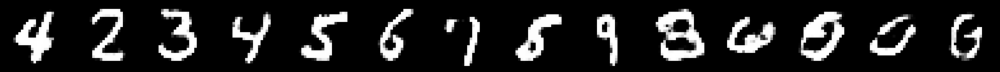} \\[-2.15pt] 
             \raisebox{6pt}{$\bigstar$ VAE + Sylvester Flows} & \includegraphics[clip,width=0.7\linewidth]{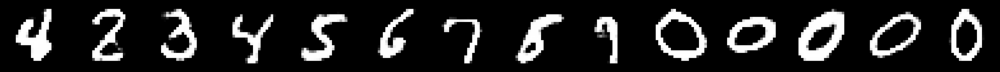} \\[-2.15pt] 

             \raisebox{6pt}{VAE + VampPrior} & \includegraphics[clip,width=0.7\linewidth]{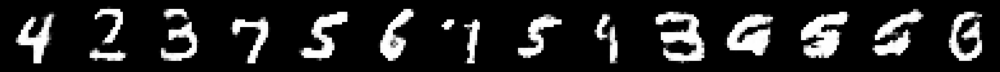} \\[-2.15pt] 
             \raisebox{6pt}{$\bigstar$ VAE + VampPrior} & \includegraphics[clip,width=0.7\linewidth]{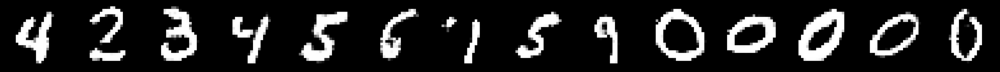} \\[-2.15pt] 

             \raisebox{6pt}{2-Stage VAE} & \includegraphics[clip,width=0.7\linewidth]{figures/mnist_conv/res_vae_appendix.png} \\[-2.15pt] 
             \raisebox{6pt}{$\bigstar$ 2-Stage VAE} & \includegraphics[clip,width=0.7\linewidth]{figures/mnist_conv/res_vae_combined_appendix.png} \\[-2.15pt] 

             \raisebox{6pt}{BAE + $\cN(0, 1)$ Prior} & \includegraphics[clip,width=0.7\linewidth]{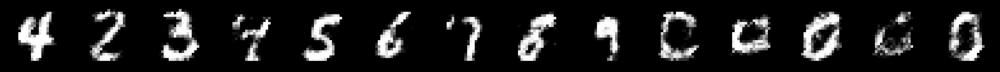} \\[-2.15pt] 
             \raisebox{6pt}{$\bigstar$ BAE + $\cN(0, 1)$ Prior} & \includegraphics[clip,width=0.7\linewidth]{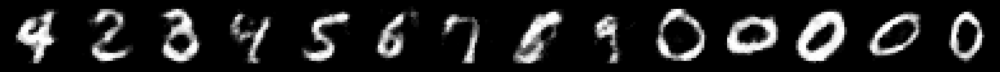} \\[-2.15pt] 

             \raisebox{6pt}{BAE + Optim. Prior (\textbf{Ours})} & \includegraphics[clip,width=0.7\linewidth]{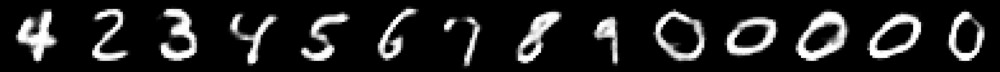} \\[-2.15pt] 

            \bottomrule
         \end{tabular}
      \end{sc}}
   \caption{Qualitative evaluation for reconstructed samples on \mnist.
      \textcolor{black}{$\bigstar$} indicates that we use the union of the training data and the data used to optimize prior to train the model.
      Here, the training size is $200$.}
   \label{tab:recons_mnist_appendix}
\end{table}

\begin{table}[!hp]
   \centering
   \scalebox{.9}
   {\setlength{\tabcolsep}{5pt}
      \begin{sc}
         \small
         \begin{tabular}{rc}
            \toprule
             & MNIST - Generated Samples \\
             \midrule
             
             \raisebox{6pt}{WAE} & \includegraphics[clip,width=0.7\linewidth]{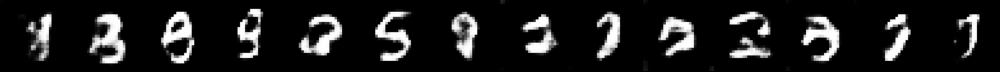} \\[-2.15pt] 
             \raisebox{6pt}{$\bigstar$ WAE} & \includegraphics[clip,width=0.7\linewidth]{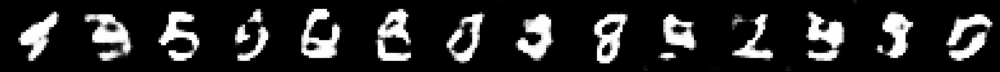} \\[-2.15pt] 

             \raisebox{6pt}{VAE} & \includegraphics[clip,width=0.7\linewidth]{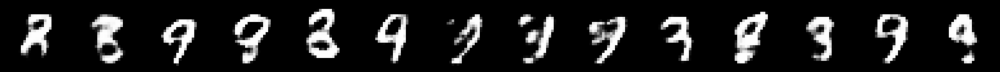} \\[-2.15pt] 
             \raisebox{6pt}{$\bigstar$ VAE} & \includegraphics[clip,width=0.7\linewidth]{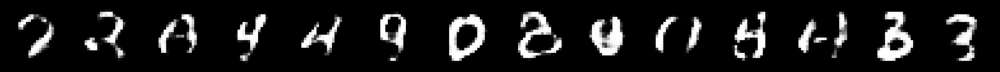} \\[-2.15pt] 
            
             \raisebox{6pt}{$\beta$-VAE} & \includegraphics[clip,width=0.7\linewidth]{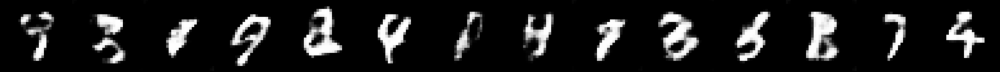} \\[-2.15pt] 
             \raisebox{6pt}{$\bigstar$ $\beta$-VAE} & \includegraphics[clip,width=0.7\linewidth]{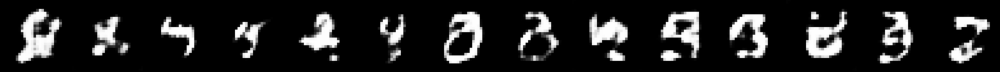} \\[-2.15pt] 

             \raisebox{6pt}{VAE + Sylvester Flows} & \includegraphics[clip,width=0.7\linewidth]{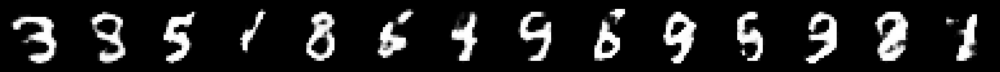} \\[-2.15pt] 
             \raisebox{6pt}{$\bigstar$ VAE + Sylvester Flows} & \includegraphics[clip,width=0.7\linewidth]{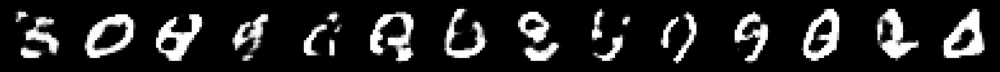} \\[-2.15pt] 

             \raisebox{6pt}{VAE + VampPrior} & \includegraphics[clip,width=0.7\linewidth]{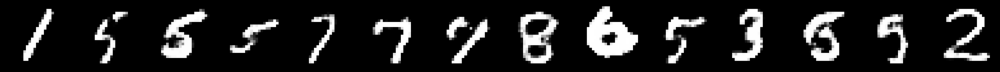} \\[-2.15pt] 
             \raisebox{6pt}{$\bigstar$ VAE + VampPrior} & \includegraphics[clip,width=0.7\linewidth]{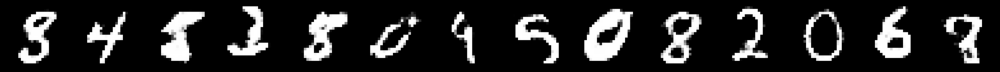} \\[-2.15pt] 

             \raisebox{6pt}{2-Stage VAE} & \includegraphics[clip,width=0.7\linewidth]{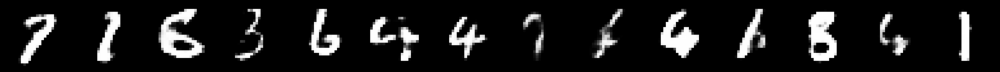} \\[-2.15pt] 
             \raisebox{6pt}{$\bigstar$ 2-Stage VAE} & \includegraphics[clip,width=0.7\linewidth]{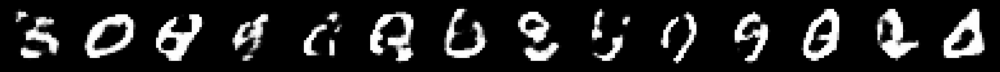} \\[-2.15pt] 

             \raisebox{6pt}{BAE + $\cN(0, 1)$ Prior} & \includegraphics[clip,width=0.7\linewidth]{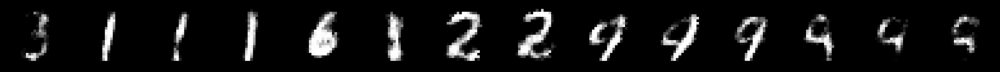} \\[-2.15pt] 
             \raisebox{6pt}{$\bigstar$ BAE + $\cN(0, 1)$ Prior} & \includegraphics[clip,width=0.7\linewidth]{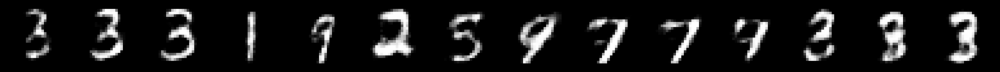} \\[-2.15pt] 

             \raisebox{6pt}{BAE + Optim. Prior (\textbf{Ours})} & \includegraphics[clip,width=0.7\linewidth]{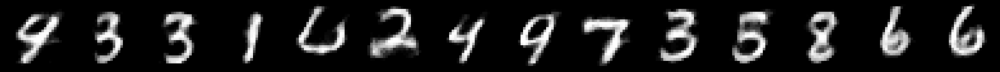} \\[-2.15pt] 

            \bottomrule
         \end{tabular}
      \end{sc}}
   \caption{Qualitative evaluation for generated samples on \mnist.
      \textcolor{black}{$\bigstar$} indicates that we use the union of the training data and the data used to optimize prior to train the model.
      Here, the training size is $200$.}
   \label{tab:gen_mnist_appendix}
\end{table}

\begin{table}[!hp]
   \centering
   \scalebox{.9}
   {\setlength{\tabcolsep}{5pt}
      \begin{sc}
         \small
         \begin{tabular}{rc}
            \toprule
             & Yale - Reconstructions \\
             \midrule
             \raisebox{6pt}{Ground Truth} & \includegraphics[clip,width=0.7\linewidth]{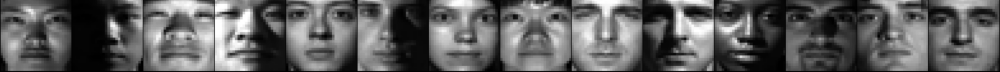} \\[-2.15pt] 
             
             \raisebox{6pt}{WAE} & \includegraphics[clip,width=0.7\linewidth]{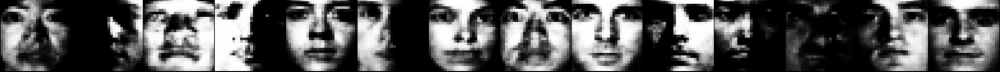} \\[-2.15pt] 
             \raisebox{6pt}{$\bigstar$ WAE} & \includegraphics[clip,width=0.7\linewidth]{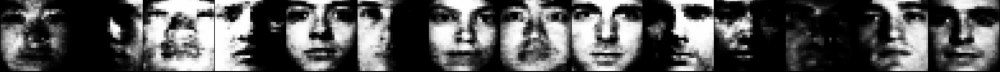} \\[-2.15pt] 

             \raisebox{6pt}{VAE} & \includegraphics[clip,width=0.7\linewidth]{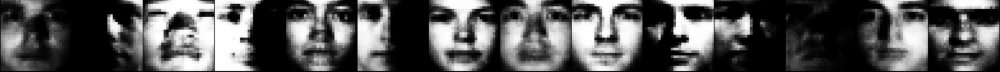} \\[-2.15pt] 
             \raisebox{6pt}{$\bigstar$ VAE} & \includegraphics[clip,width=0.7\linewidth]{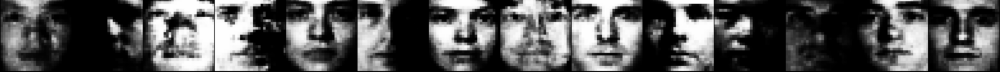} \\[-2.15pt] 
            
             \raisebox{6pt}{$\beta$-VAE} & \includegraphics[clip,width=0.7\linewidth]{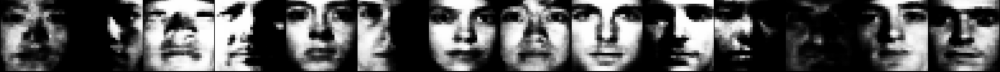} \\[-2.15pt] 
             \raisebox{6pt}{$\bigstar$ $\beta$-VAE} & \includegraphics[clip,width=0.7\linewidth]{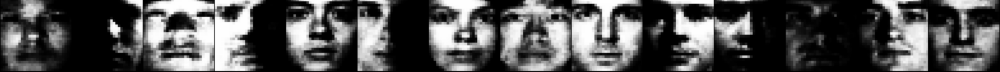} \\[-2.15pt] 

             \raisebox{6pt}{VAE + Sylvester Flows} & \includegraphics[clip,width=0.7\linewidth]{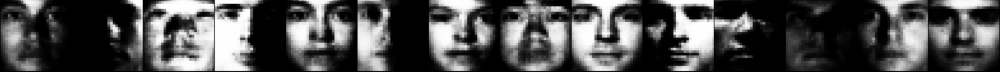} \\[-2.15pt] 
             \raisebox{6pt}{$\bigstar$ VAE + Sylvester Flows} & \includegraphics[clip,width=0.7\linewidth]{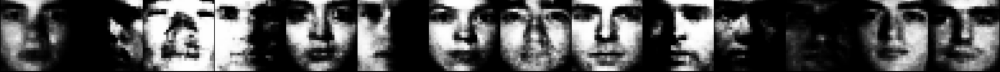} \\[-2.15pt] 

             \raisebox{6pt}{VAE + VampPrior} & \includegraphics[clip,width=0.7\linewidth]{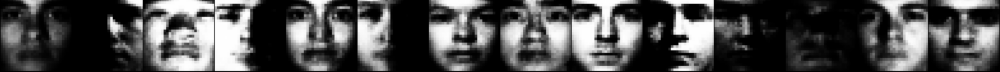} \\[-2.15pt] 
             \raisebox{6pt}{$\bigstar$ VAE + VampPrior} & \includegraphics[clip,width=0.7\linewidth]{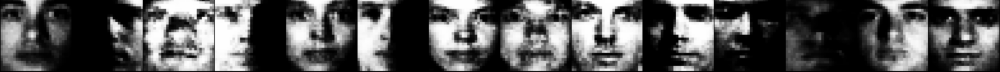} \\[-2.15pt] 

             \raisebox{6pt}{2-Stage VAE} & \includegraphics[clip,width=0.7\linewidth]{figures/face_conv/res_vae_appendix.png} \\[-2.15pt] 
             \raisebox{6pt}{$\bigstar$ 2-Stage VAE} & \includegraphics[clip,width=0.7\linewidth]{figures/face_conv/res_vae_combined_appendix.png} \\[-2.15pt] 

             \raisebox{6pt}{BAE + $\cN(0, 1)$ Prior} & \includegraphics[clip,width=0.7\linewidth]{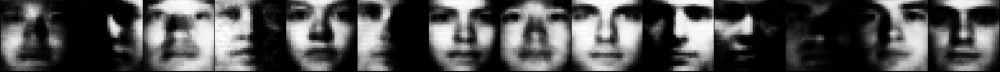} \\[-2.15pt] 
             \raisebox{6pt}{$\bigstar$ BAE + $\cN(0, 1)$ Prior} & \includegraphics[clip,width=0.7\linewidth]{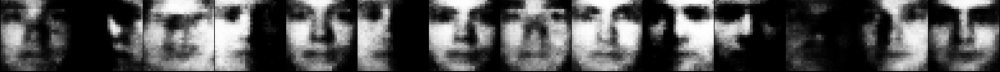} \\[-2.15pt] 

             \raisebox{6pt}{BAE + Optim. Prior (\textbf{Ours})} & \includegraphics[clip,width=0.7\linewidth]{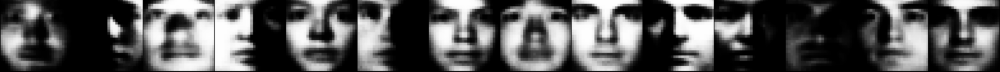} \\[-2.15pt] 

            \bottomrule
         \end{tabular}
      \end{sc}}
   \caption{Qualitative evaluation for reconstructed samples on \yale.
      \textcolor{black}{$\bigstar$} indicates that we use the union of the training data and the data used to optimize prior to train the model.
      Here, the training size is $500$.}
   \label{tab:recons_fale_appendix}
\end{table}

\begin{table}[!hp]
   \centering
   \scalebox{.9}
   {\setlength{\tabcolsep}{5pt}
      \begin{sc}
         \small
         \begin{tabular}{rc}
            \toprule
             & Yale - Generated Samples \\
             \midrule
             
             \raisebox{6pt}{WAE} & \includegraphics[clip,width=0.7\linewidth]{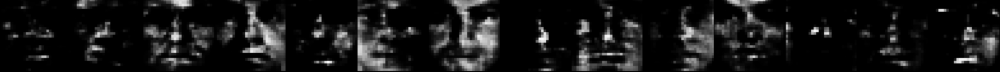} \\[-2.15pt] 
             \raisebox{6pt}{$\bigstar$ WAE} & \includegraphics[clip,width=0.7\linewidth]{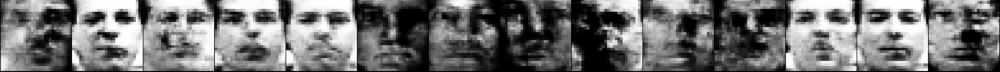} \\[-2.15pt] 

             \raisebox{6pt}{VAE} & \includegraphics[clip,width=0.7\linewidth]{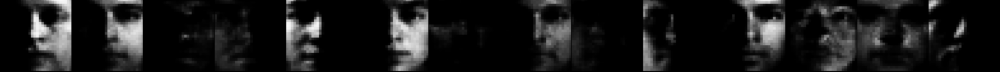} \\[-2.15pt] 
             \raisebox{6pt}{$\bigstar$ VAE} & \includegraphics[clip,width=0.7\linewidth]{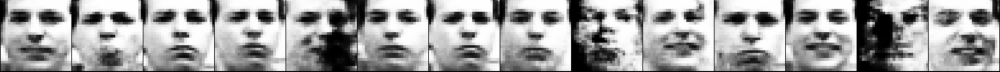} \\[-2.15pt] 
            
             \raisebox{6pt}{$\beta$-VAE} & \includegraphics[clip,width=0.7\linewidth]{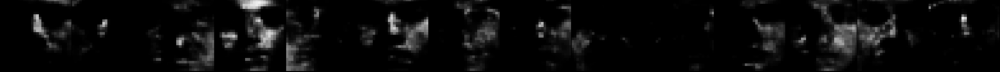} \\[-2.15pt] 
             \raisebox{6pt}{$\bigstar$ $\beta$-VAE} & \includegraphics[clip,width=0.7\linewidth]{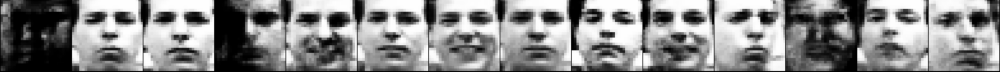} \\[-2.15pt] 

             \raisebox{6pt}{VAE + Sylvester Flows} & \includegraphics[clip,width=0.7\linewidth]{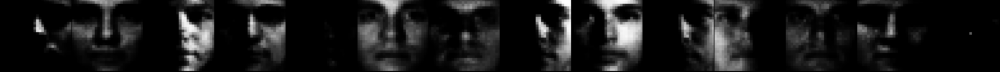} \\[-2.15pt] 
             \raisebox{6pt}{$\bigstar$ VAE + Sylvester Flows} & \includegraphics[clip,width=0.7\linewidth]{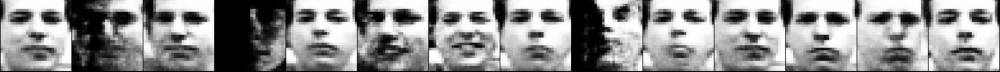} \\[-2.15pt] 

             \raisebox{6pt}{VAE + VampPrior} & \includegraphics[clip,width=0.7\linewidth]{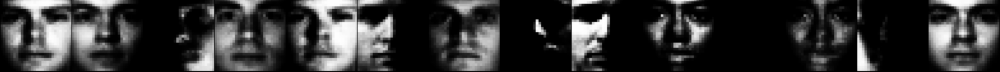} \\[-2.15pt] 
             \raisebox{6pt}{$\bigstar$ VAE + VampPrior} & \includegraphics[clip,width=0.7\linewidth]{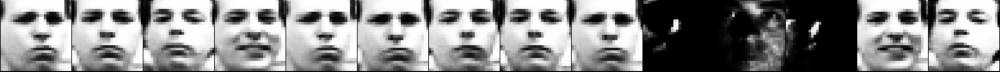} \\[-2.15pt] 

             \raisebox{6pt}{2-Stage VAE} & \includegraphics[clip,width=0.7\linewidth]{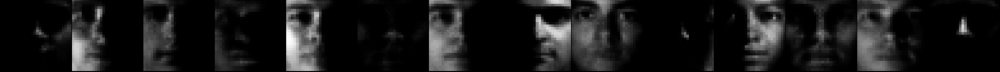} \\[-2.15pt] 
             \raisebox{6pt}{$\bigstar$ 2-Stage VAE} & \includegraphics[clip,width=0.7\linewidth]{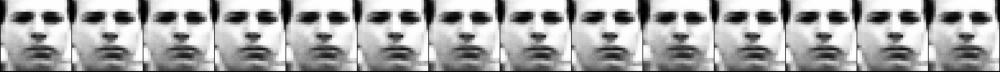} \\[-2.15pt] 

             \raisebox{6pt}{BAE + $\cN(0, 1)$ Prior} & \includegraphics[clip,width=0.7\linewidth]{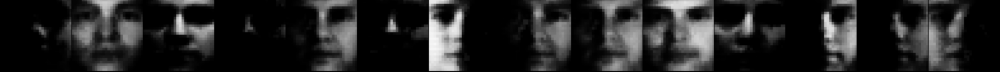} \\[-2.15pt] 
             \raisebox{6pt}{$\bigstar$ BAE + $\cN(0, 1)$ Prior} & \includegraphics[clip,width=0.7\linewidth]{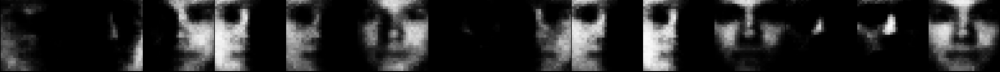} \\[-2.15pt] 

             \raisebox{6pt}{BAE + Optim. Prior (\textbf{Ours})} & \includegraphics[clip,width=0.7\linewidth]{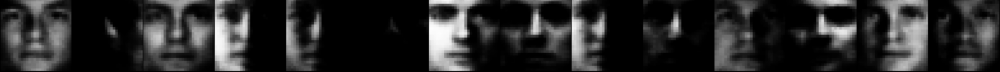} \\[-2.15pt] 

            \bottomrule
         \end{tabular}
      \end{sc}}
   \caption{Qualitative evaluation for generated samples on \yale.
      \textcolor{black}{$\bigstar$} indicates that we use the union of the training data and the data used to optimize prior to train the model.
      Here, the training size is $500$.}
   \label{tab:gen_fale_appendix}
\end{table}

\end{document}